\documentclass{article}

\usepackage{microtype}
\usepackage{graphicx}
\usepackage{subfigure}
\usepackage{booktabs} 

\usepackage{hyperref}

\usepackage[accepted]{icml2019}

\usepackage[lower, theorems, autonum, eqref]{changdefs}
\usepackage{wrapfig}
\usepackage{multirow, multicol}

\newcommand{\hKp}{{\hK'}}
\newcommand{\hKpp}{{\hK''}}

\newcommand{\vgf}{v^{\textnormal{\tiny GF}}}
\newcommand{\vsvgd}{v^{\textnormal{\tiny SVGD}}}

\newcommand{\hvsvgd}{{\hvv}^{\textnormal{\tiny SVGD}}}
\newcommand{\hvgfsf}{{\hvv}^{\textnormal{\tiny GFSF}}}
\newcommand{\ugf}{u^{\textnormal{\tiny GF}}}

\newcommand{\ublob}{u^{\textnormal{\tiny Blob}}}
\newcommand{\ugfsd}{u^{\textnormal{\tiny GFSD}}}
\newcommand{\ugfsf}{u^{\textnormal{\tiny GFSF}}}
\newcommand{\hugfsf}{{\huu}^{\textnormal{\tiny GFSF}}}
\newcommand{\CC}{C_c^{\infty}}
\newcommand{\clCC}{\clC_c^{\infty}}
\newcommand{\supp}{\mathrm{supp}}

\usepackage{color}

\icmltitlerunning{Understanding and Accelerating Particle-Based Variational Inference}

\begin{document}

\twocolumn[
\icmltitle{Understanding and Accelerating Particle-Based Variational Inference}



\icmlsetsymbol{equal}{*}

\begin{icmlauthorlist}
  \icmlauthor{Chang Liu}{thu}
  \icmlauthor{Jingwei Zhuo}{thu}
  \icmlauthor{Pengyu Cheng}{dku}
  \icmlauthor{Ruiyi Zhang}{dku}
  \icmlauthor{Jun Zhu}{thu}
  \icmlauthor{Lawrence Carin}{dku}
\end{icmlauthorlist}

\icmlaffiliation{thu}{Dept. of Comp. Sci. \& Tech., Institute for AI, BNRist Center, Tsinghua-Fuzhou Inst. for Data Tech., THBI Lab, Tsinghua University, Beijing, 100084, China}
\icmlaffiliation{dku}{Dept. of Elec. \& Comp. Engineering, Duke University, NC, USA}

\icmlcorrespondingauthor{Jun Zhu}{dcszj@tsinghua.edu.cn}
\icmlcorrespondingauthor{Lawrence Carin}{lcarin@duke.edu}

\icmlkeywords{Wasserstein space, gradient flow, Bayesian inference, particle-based variational inference, Nesterov's acceleration, Riemannian optimization}

\vskip 0.3in
]



\printAffiliationsAndNotice{}  

\begin{abstract}
  Particle-based variational inference methods (ParVIs) have gained attention in the Bayesian inference literature, for their capacity to yield flexible and accurate approximations.
  We explore ParVIs from the perspective of Wasserstein gradient flows, and make both theoretical and practical contributions.
  We unify various finite-particle approximations that existing ParVIs use, and recognize that the approximation is essentially a compulsory smoothing treatment, in either of two equivalent forms.
  This novel understanding reveals the assumptions and relations of existing ParVIs, and also inspires new ParVIs. 
  We propose an acceleration framework and a principled bandwidth-selection method for general ParVIs; these are based on the developed theory and leverage the geometry of the Wasserstein space.
  Experimental results show the improved convergence by the acceleration framework and enhanced sample accuracy by the bandwidth-selection method.
\end{abstract}

\vspace{-5pt}
\section{Introduction}
\vspace{-3pt}
Bayesian inference provides powerful tools for modeling and reasoning with uncertainty.
In Bayesian learning one seeks access to the posterior distribution $p$ on the support space $\clX$ (\ie, the latent space) given data.
As $p$ is intractable in general, various approximations have been developed.
Variational inference methods (VIs) seek to approximate $p$ within a certain distribution family by minimizing typically the Kullback-Leibler (KL) divergence to $p$.
The approximating distribution is commonly chosen to be a member of a parametric family~\citep{wainwright2008graphical, hoffman2013stochastic}, 
but often it poses a restrictive assumption and the closeness to $p$ suffers.
Markov chain Monte Carlo (MCMC) methods~\citep{geman1987stochastic, neal2011mcmc, welling2011bayesian, ding2014bayesian} aim to directly draw samples from $p$.
Although asymptotically accurate, they often converge slowly in practice due to undesirable autocorrelation between samples.
A relatively large sample size is needed~\cite{liu2016stein} for a good result, increasing the cost for downstream tasks.

Recently, particle-based variational inference methods (ParVIs) 
have been proposed.
They use a set of samples, or particles, to represent the approximating distribution (like MCMC) and deterministically update particles by minimizing the KL-divergence to $p$ (like VIs). 
ParVIs have greater non-parametric flexibility than classical VIs, and are also more particle-efficient than MCMCs, 
since they make full use of a finite number of particles by taking particle interaction into account. 
The availability of optimization-based update rules also makes them converge faster.
Stein Variational Gradient Descent (SVGD)~\cite{liu2016stein} is a representative method of this type;
it updates particles by leveraging a proper vector field that minimizes the KL-divergence.
Its unique benefits make SVGD popular, with many variants~\cite{liu2018riemannian, zhuo2018message, chen2018stein, futami2018frank} and applications~\cite{feng2017learning, pu2017vae, liu2017steinpolicy, haarnoja2017reinforcement, zhang2018policy, zhang2019scalable}.

SVGD was later understood as simulating the steepest descending curves, or \emph{gradient flows}, of the KL-divergence on a certain kernel-related distribution space $\clP_{\clH}(\clX)$ \cite{liu2017steinflow}. 
Inspired by this, more ParVIs have been developed by simulating the gradient flow on the Wasserstein space $\clP_2(\clX)$ \cite{ambrosio2008gradient, villani2008optimal} with a finite set of particles.
The particle optimization method (PO) \cite{chen2017particle} and $w$-SGLD method \cite{chen2018unified} explore the minimizing movement scheme (MMS) (\citet{jordan1998variational}; \citet{ambrosio2008gradient}, Def.~2.0.6) of the gradient flow and make approximations for tractability with a finite set of particles.
The Blob method (originally called $w$-SGLD-B) \cite{chen2018unified} uses the vector field form of the gradient flow and approximates the update direction with a finite set of particles.
Empirical comparisons of these ParVIs have been conducted, but theoretical understanding on their finite-particle approximations for gradient flow simulation remains unknown, particularly for the assumption and relation of these approximations.
We also note that, from the optimization point of view, all existing ParVIs simulate the gradient flow, but no ParVI yet exploits the geometry of $\clP_2(\clX)$ and uses the more appealing accelerated first-order methods on the manifold $\clP_2(\clX)$.
Moreover, the smoothing kernel bandwidth of ParVIs is found crucial for performance~\cite{zhuo2018message} and a more principled bandwidth selection method is needed, relative to the current heuristic median method~\cite{liu2016stein}.

In this work, we examine the $\clP_2(\clX)$ gradient flow perspective to address these problems and demands, and contribute to the ParVI field a unified theory on the finite-particle approximations, and two practical techniques: an acceleration framework and a principled bandwidth selection method.
The theory discovers that various finite-particle approximations of ParVIs are essentially a smoothing operation, in the form of either smoothing the density or smoothing functions.
We reveal the two forms of smoothing, 
and draw a connection among ParVIs by discovering their equivalence.
Furthermore, we recognize that ParVIs actually and necessarily make an assumption on the approximation family. 
The theory also establishes a principle for developing ParVIs, and we use this principle to conceive two novel models. 
The acceleration framework follows the Riemannian version~\cite{liu2017accelerated, zhang2018estimate} of Nesterov's acceleration method~\cite{nesterov1983method}, which enjoys a proved convergence improvement over direct gradient flow simulation.
In developing the framework, we make novel use of 
the Riemannian structure of $\clP_2(\clX)$, particularly the inverse exponential map and parallel transport. 
We emphasize that the direct application of Nesterov's acceleration on every particle in $\clX$ is unsound theoretically, since the KL-divergence is not minimized on $\clX$ but on $\clP_2(\clX)$, and each single particle is not optimizing a function.
For the bandwidth method, we elaborate on the goal of smoothing and develop a principled approach for setting the bandwidth parameter.
Experimental results show the improved sample quality by the principled bandwidth method over the median method~\cite{liu2016stein}, and improved convergence of the acceleration framework on both supervised and unsupervised tasks.

\textbf{Related work}~~
On understanding SVGD, \citet{liu2017steinflow} first views it as a gradient flow on $\clP_{\clH}(\clX)$.
\citet{chen2018unified} then find it implausible to exactly formulate SVGD as a gradient flow on $\clP_2(\clX)$. 
In this work, we find SVGD approximates the $\clP_2(\clX)$ gradient flow by smoothing functions, achieving an understanding and improvement of all ParVIs within a universal perspective. 
The $\clP_{\clH}(\clX)$ viewpoint is difficult to apply in general and it lacks appealing properties.

On accelerating ParVIs, the particle optimization method (PO) \cite{chen2017particle} resembles the Polyak's momentum~\cite{polyak1964some} version of SVGD.
However, it was not derived originally for acceleration and is thus not theoretically sound for such; we observe that PO is less stable than our acceleration framework, similar to the discussion by \citet{sutskever2013importance}.
More recently, \citet{taghvaei2018accelerated} also considered accelerating the $\clP_2(\clX)$ gradient flow.
They use the variational formulation of optimization methods \cite{wibisono2016variational} and define components in the formulation for $\clP_2(\clX)$, while we leverage the geometry of $\clP_2(\clX)$ and apply Riemannian acceleration methods.
Algorithmically, \citet{taghvaei2018accelerated} use a set of momentums and we use auxiliary particles.
However, both approaches face the problem of a finite-particle approximation, solved here systematically using our theory.
Their implementation is recognized, in our theory, as smoothing the density.
Moreover, our novel perspective on $\clP_2(\clX)$ and corresponding techniques provide a general tool that enables other Riemannian optimization techniques for ParVIs.

On incorporating Riemannian structure with ParVIs, \citet{liu2018riemannian} consider the case for which the support space $\clX$ is a Riemannian manifold, either specified by task or taken as the manifold in information geometry \cite{amari2016information}. 
We summarize that they utilize the geometry of the Riemannian support space $\clX$, while we leverage deeper knowledge on the Riemannian structure of $\clP_2(\clX)$ itself.
Algorithmically, our acceleration is model-agnostic and computationally cheaper.
The work of \citet{detommaso2018stein} considers second-order information of the KL-divergence on $\clP_{\clH}(\clX)$. 
Our acceleration remains first-order, and we consider the Wasserstein space $\clP_2(\clX)$ so that with our theory the acceleration is applicable for all ParVIs.


\vspace{-5pt}
\section{Preliminaries}
\vspace{-5pt}
We first introduce the Wasserstein space $\clP_2(\clX)$ and the gradient flow on it, and review related ParVIs.
We only consider Euclidean support $\clX = \bbR^D$ to reduce unnecessary sophistication, and highlight our key contributions.

We denote $\clCC$ as the set of compactly supported $\bbR^D$-valued smooth functions on $\clX$, and $\CC$ for scalar-valued functions.
Denote $\clL^2_{q}$ as the Hilbert space of $\bbR^D$-valued functions $\{u: \bbR^D \to \bbR^D \mid \int \lrVert{u(x)}_2^2 \dd q < \infty\}$ with inner product $\lrangle{ u, v }_{\clL^2_q} := \int u(x)\cdot v(x) \dd q$, and $L^2_q$ for scalar-valued functions.
The Lebesgue measure is taken if $q$ is not specified.
We define the push-forward of a distribution $q$ under a measurable transformation $\clT: \clX \to \clX$ as the distribution of the $\clT$-transformed random variable of $q$, denoted as $\clT_{\#} q$.

\vspace{-5pt}
\subsection{The Riemannian Structure of the Wasserstein Space $\clP_2(\clX)$}
\vspace{-5pt}
\label{sec:rm}

\begin{figure}[t]\vspace{-3pt}
  \centering
  \includegraphics[width=.45\textwidth]{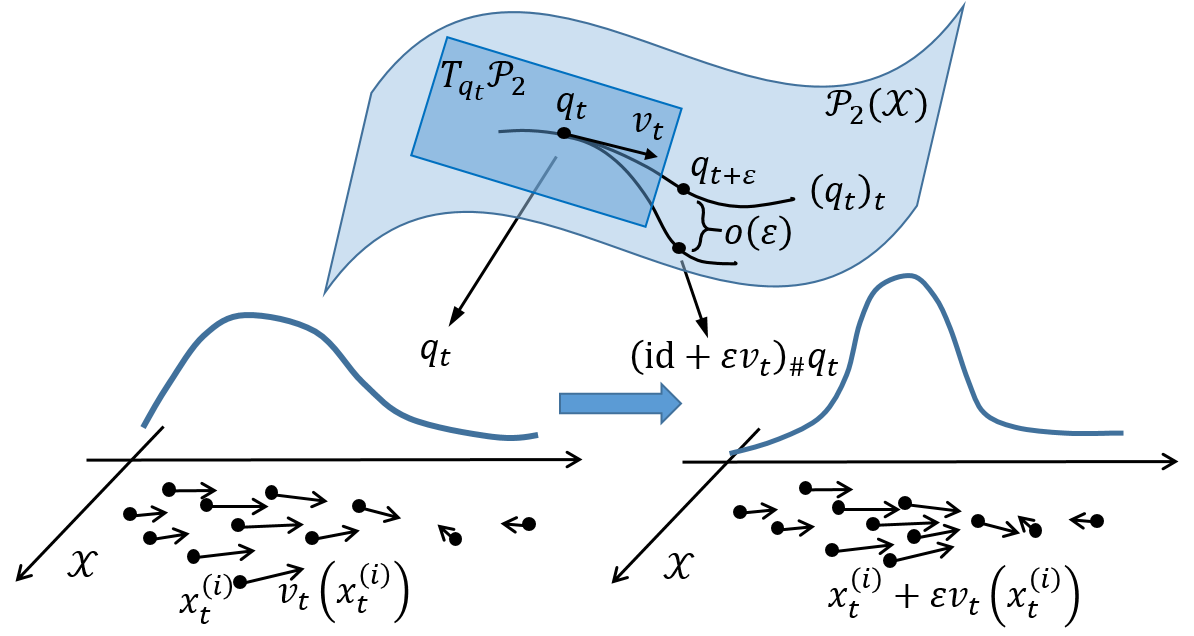}\vspace{-10pt}
  \caption{Illustration of the Wasserstein space $\clP_2(\clX)$ and related concepts.}
  \vspace{-18pt}
  \label{fig:wspace}
\end{figure}

Figure~\ref{fig:wspace} illustrates the related concepts discussed here.
Consider distributions on a support space $\clX$ with distance $d(\cdot,\cdot)$, and denote $\clP(\clX)$ as the set of all such distributions.
The Wasserstein space is the metric space
$\clP_2(\clX) \!:=\! \{q\in\clP(\clX): \! \exists x_0\in\clX \st \bbE_{q}[d(x_0,x)^2] < +\infty\}$
equipped with the well-known Wasserstein distance $W_2$ (\citet{villani2008optimal}, Def.~6.4).
Its Riemannian structure is then discovered \citep{otto2001geometry, benamou2000computational}, enabling explicit expression of quantities of interest, like gradient. 
The first step is the recognition of tangent vectors and tangent spaces on it.
For any smooth curve $(q_t)_t$ on $\clP_2(\clX)$, there exists an a.e.-unique time-dependent vector field $v_t(x)$ on $\clX$ such that for a.e. $t\in\bbR$, $\partial_t q_t + \nabla\cdot(v_t q_t) = 0$ and $v_t\in \overline{\{\nabla\varphi: \varphi \in \CC\}}^{\clL^2_{q_t}}$
where the overline means closure (\citet{villani2008optimal}, Thm.~13.8; \citet{ambrosio2008gradient}, Thm.~8.3.1, Prop.~8.4.5).
The unique existence of such a $v_t$ allows us to recognize $v_t$ as the tangent vector of the curve at $q_t$, and the mentioned closure as the tangent space at $q_t$, denoted by $T_{q_t} \clP_2$ (\citet{ambrosio2008gradient}, Def.~8.4.1).
The inherited inner product in $T_{q_t} \clP_2$ from $\clL^2_{q_t}$ defines a Riemannian structure on $\clP_2(\clX)$, and it is consistent with the Wasserstein distance $W_2$ due to the Benamou-Brenier formula~\cite{benamou2000computational}.
Restricted to a parametric family as a submanifold of $\clP_2(\clX)$, this structure gives a metric in the parameter space that differs from the Fisher-Rao metric \cite{amari2016information}, and it has been used in classical VIs \cite{chen2018natural}.
Finally, the vector field representation provides a convenient means of simulating the distribution curve $(q_t)_t$: it is known that $(\id + \varepsilon v_t)_{\#} q_t$ is a first-order approximation of $q_{t+\varepsilon}$ in terms of $W_2$ (\citet{ambrosio2008gradient}, Prop.~8.4.6).
This means that given a set of samples $\{x^{(i)}\}_i$ of $q_t$, $\{x^{(i)} + \varepsilon v_t(x^{(i)})\}_i$ is approximately a set of samples of $q_{t+\varepsilon}$ for small $\varepsilon$.


\vspace{-5pt}
\subsection{Gradient Flows on $\clP_2(\clX)$}
\vspace{-5pt}
\label{sec:gf}

Gradient flow of a function $F$ is roughly the family of \emph{steepest descending curves} $\{(q_t)_t\}$ for $F$.
It has various technical definitions on metric spaces (\citet{ambrosio2008gradient}, Def.~11.1.1; \citet{villani2008optimal}, Def.~23.7), \eg, as the limit of the minimizing movement scheme (MMS; \citet{ambrosio2008gradient}, Def.~2.0.6):
{\setlength\abovedisplayskip{0pt}
\setlength\belowdisplayskip{5pt}
\begin{align}
  q_{t+\varepsilon} = \argmin_{q\in \clP_2(\clX)} F(q) + \frac{1}{2\varepsilon}W_2^2(q, q_t),
  \label{eqn:mms}
\end{align}
}and these definitions all coincide when a Riemannian structure is endowed (\citet{villani2008optimal}, Prop.~23.1, Rem.~23.4; \citet{ambrosio2008gradient}, Thm.~11.1.6; \citet{erbar2010heat}, Lem.~2.7).
In this case the gradient flow $(q_t)_t$ has its tangent vector at any $t$ being the gradient of $F$ at $q_t$, defined as:
{\setlength\abovedisplayskip{4pt}
\setlength\belowdisplayskip{5pt}
\begin{align}
  \!\! \grad F(q_t) \!:=\! \maxargmax_{v:\; \lrVert{v}_{T_{q_t} \! \clP_2} = 1} \frac{\ud}{\ud \varepsilon} F \Big(\! (\id \!+\! \varepsilon v)_{\#} q_t \!\Big) \! \Big|_{\varepsilon=0}, \!
  \label{eqn:grad}
\end{align}
}where ``$\maxargmax$'' denotes the scalar multiplication of the maximum and the maximizer.
For Bayesian inference tasks, given an absolutely continuous target distribution $p$, 
we aim to minimize the KL-divergence (a.k.a relative entropy) $\KL_p(q) := \int_{\clX} \log(q/p) \dd q$.
Its gradient flow $(q_t)_t$ has its tangent vector at any $t$ being:
{\setlength\abovedisplayskip{2pt}
\setlength\belowdisplayskip{4pt}
\begin{align}
  \vgf_t := -\grad \KL_p(q_t) = \nabla\log p - \nabla\log q_t,
  \label{eqn:vgf}
\end{align}
}whenever $q_t$ is absolutely continuous (\citet{villani2008optimal}, Thm.~23.18; \citet{ambrosio2008gradient}, Example~11.1.2).
When $\KL_p$ is geodesically $\mu$-convex on $\clP_2(\clX)$,\footnote{{\it E.g.}, $p$ is $\mu$-log-concave on $\clX$ (\citet{villani2008optimal}, Thm.~17.15).}
the gradient flow $(q_t)_t$ enjoys exponential convergence: $W_2(q_t, p) \le e^{-\mu t} W_2(q_0, p)$ (\citet{villani2008optimal}, Thm.~23.25, Thm.~24.7; \citet{ambrosio2008gradient}, Thm.~11.1.4), as expected.

\begin{remark}
  \label{rem:langevin}
  The Langevin dynamics $\ud x = \nabla \log p(x) \dd t + \sqrt{2} \dd B_t(x)$ ($B_t$ is the Brownian motion) is also known to produce the gradient flow of $\KL_p$ on $\clP_2(\clX)$ (\eg, \citet{jordan1998variational} from the MMS perspective).
  It produces the same distribution curve $(q_t)_t$ as the deterministic dynamics $\ud x = \vgf_t(x) \dd t$.
\end{remark}
\vspace{-4pt}


\vspace{-5pt}
\subsection{Particle-Based Variational Inference Methods}
\vspace{-5pt}
\label{sec:parvis}

Stein Variational Gradient Descent (SVGD)~\cite{liu2016stein} uses a vector field $v$ to update particles: $x^{(i)}_{k+1} = x^{(i)}_k + \varepsilon v(x^{(i)}_k)$, and $v$ is selected to maximize the decreasing rate: $-\frac{\ud}{\ud \varepsilon} \! \KL_p \big(\! (\id + \varepsilon v)_{\#} q \big) \big|_{\varepsilon=0}$,
where $q$ is the distribution that $\{x^{(i)}\}_i$ obeys.
When $v$ is optimized over the vector-valued reproducing kernel Hilbert space (RKHS) $\clH^D$ of a kernel $K$ (\citet{steinwart2008support}, Def.~4.18), the solution can be derived in closed-form:
{\setlength\abovedisplayskip{3pt}
\setlength\belowdisplayskip{3pt}
\begin{align}
  \vsvgd(\cdot) := \bbE_{q(x)}[K(x,\cdot)\nabla\log p(x) + \nabla_{x} K(x,\cdot)].
  \label{eqn:svgdvec}
\end{align}
}Noting that the optimization problem fits the form of \eqref{eqn:grad}, \citet{liu2017steinflow} interprets SVGD as the gradient flow on $\clP_{\clH}$, a distribution manifold that takes $\clH^D$ as its tangent space.
Equation~\origeqref{eqn:svgdvec} can be estimated by a finite set of particles, equivalently taking $q(x)$ as the empirical distribution $\hqq(x) := \frac{1}{N} \sum_{i=1}^N \delta_{x^{(i)}}(x)$.

Other methods have been developed to simulate the $\clP_2(\clX)$ gradient flow.
The Blob method~\cite{chen2018unified} estimates \eqref{eqn:vgf} with a finite set of particles.
It reformulates the intractable part $\ugf := -\nabla\log q$ from the perspective of variation:
  $\ugf = \nabla \big( -\frac{\delta}{\delta q} \bbE_q[\log q] \big)$, 
and then partly smooths the density $q$ by convolving with a kernel $K$:
{\setlength\abovedisplayskip{3pt}
\setlength\belowdisplayskip{6pt}
\begin{align}
  \ublob = \nabla \big( -\frac{\delta}{\delta q} \bbE_q[\log(q*K)] \big)
  \label{eqn:blobvec}
\end{align}
}$= -\nabla\log \tqq - \nabla \big( (q/\tqq) * K \big)$, where $\tqq := q*K$ and ``$*$'' denotes convolution.
This form enables the usage of $\hqq$.

The particle optimization method (PO)~\cite{chen2017particle} simulates the $\clP_2(\clX)$ gradient flow by MMS~\origeqref{eqn:mms}, where the Wasserstein distance $W_2$ is estimated by solving a dual optimal transport problem that optimizes over quadratic functions.
The resulting update rule $x^{(i)}_k = $
{\setlength\abovedisplayskip{5pt}
\setlength\belowdisplayskip{5pt}
\begin{align}
  x^{(i)}_{k-1} + \varepsilon (\vsvgd(x^{(i)}_{k-1}) + \clN(0, \sigma^2 I)) + \mu (x^{(i)}_{k-1} - x^{(i)}_{k-2})
\end{align}
}(with parameters $\varepsilon, \sigma, \mu$) comes in the form of the Polyak's momentum~\cite{polyak1964some} version of SVGD.
The $w$-SGLD method~\cite{chen2018unified} estimates $W_2$ by solving the primal problem with entropy regularization.
The algorithm is similar to PO.

\vspace{-6pt}
\section{ParVIs as Approximations to $\clP_2(\clX)$ Gradient Flow}
\vspace{-3pt}
\label{sec:uni}

This part of the paper constitutes our main theoretical contributions.
We find that ParVIs approximate the $\clP_2(\clX)$ gradient flow by smoothing, either smoothing the density or smoothing functions.
We make recognition of existing ParVIs, analyze equivalence and the necessity of smoothing, and develop two novel ParVIs based on the theory.


\vspace{-6pt}
\subsection{SVGD Approximates $\clP_2(\clX)$ Gradient Flow}
\vspace{-5pt}
\label{sec:gf-svgd}

Currently SVGD is only known to simulate the gradient flow on the distribution space $\clP_{\clH}(\clX)$.
We first interpret SVGD as a simulation of the gradient flow on the Wasserstein space $\clP_2(\clX)$ with a finite set of particles, so that all existing ParVIs can be analyzed from a common perspective.
Noting that $\vgf$ is an element of the Hilbert space $\clL^2_q$, we can identify it by:
{\setlength\abovedisplayskip{5pt}
\setlength\belowdisplayskip{3pt}
\begin{align}
  \vgf = \maxargmax_{v\in \clL^2_q, \lrVert{v}_{\clL^2_q}=1} \lrangle{ \vgf, v }_{\clL^2_q}.
  \label{eqn:svgdopt}
\end{align}
}We then find that by changing the optimization domain from $\clL^2_q$ to the vector-valued RKHS $\clH^D$ of a kernel $K$, the problem can be solved in closed-form and the solution coincides with $\vsvgd$.
This connects the two notions:
\begin{theorem}[$\vsvgd$ approximates $\vgf$]
  \label{thm:svgd}
  The SVGD vector field $\vsvgd$ defined in \eqref{eqn:svgdvec} approximates the vector field $\vgf$ of the gradient flow on $\clP_2(\clX)$, in the following sense:
  {\setlength\abovedisplayskip{5pt}
  \setlength\belowdisplayskip{-2pt}
  \begin{align}
	\vsvgd = \maxargmax_{v\in\clH^D, \lrVert{v}_{\clH^D}=1} \lrangle{ \vgf, v }_{\clL^2_q}.
  \end{align}}
  \vspace{-6pt}
\end{theorem}
The proof is provided in Appendix~A.1.
We will see that $\clH^D$ is roughly a subspace of $\clL^2_q$, so $\vsvgd$ can be seen as the projection of $\vgf$ on $\clH^D$.

The current $\clP_{\clH}(\clX)$ gradient flow interpretation of SVGD \cite{liu2017steinflow, chen2018unified} is not fully satisfying.
We point out that $\clP_{\clH}$ is not yet a well defined Riemannian manifold.
It is characterized by taking $\clH^D$ as its tangent space, but it is only known that the tangent space of a manifold is determined by the manifold's topology~\cite{do1992riemannian}, and it is unknown if there uniquely exists a manifold with a specified tangent space.
Particularly, the tangent vector of any smooth curve should uniquely exist in the tangent space.
Wasserstein space $\clP_2(\clX)$ satisfies this (\citet{villani2008optimal}, Thm.~13.8; \citet{ambrosio2008gradient}, Thm.~8.3.1, Prop.~8.4.5), but it remains unknown for $\clP_{\clH}$.
The manifold $\clP_{\clH}$ also lacks appealing properties like an explicit expression of the distance~\cite{chen2018unified}.
SVGD has also been formulated as a Vlasov process~\cite{liu2017steinflow, chen2018unified}; 
this shows that SVGD keeps $p$ invariant, but does not provide much knowledge on its convergence behavior.

\vspace{-5pt}
\subsection{ParVIs Approximate $\clP_2(\clX)$ Gradient Flow by Smoothing}
\vspace{-5pt}
\label{sec:gf-summary}

With the above knowledge, all ParVIs approximate the $\clP_2(\clX)$ gradient flow.
We then find that the approximation is made by a compulsory smoothing treatment, either smoothing the density or smoothing functions.

\textbf{Smoothing the Density}~~
We note that the Blob method approximates the $\clP_2(\clX)$ gradient flow by replacing $q$ with a smoothed density $\tqq := \hqq * K$ in the variational formulation of the gradient flow.
In the $w$-SGLD method~\cite{chen2018unified}, an entropy regularization is introduced to the primal optimal transport problem.
This term avoids the density solution being highly concentrated, and thus effectively poses a smoothing requirement on densities.

\textbf{Smoothing Functions}~~
We note in Theorem~\ref{thm:svgd} that SVGD approximates the $\clP_2(\clX)$ gradient flow by replacing the function family $\clL^2_q$ with $\clH^D$ in an optimization formulation of the gradient flow.
We then reveal the 
fact that a function in $\clH^D$ is roughly a kernel smoothed function in $\clL^2_q$, as stated formally in the following theorem:
\begin{theorem}[$\clH^D$ smooths $\clL^2_q$]
  \label{thm:smfun}
  For $\clX = \bbR^D$, a Gaussian kernel $K$ on $\clX$ and an absolutely continuous $q$, the vector-valued RKHS $\clH^D$ of $K$ is isometrically isomorphic to the closure $\clG := \overline{\{\phi*K: \phi\in\clCC\}}^{\clL^2_q}$.
  \vspace{-1pt}
\end{theorem}
The proof is presented in Appendix~A.2.
Noting that $\clCC$ is roughly $\clL^2_q$ in the sense $\overline{\clCC}^{\clL^2_q} \!=\! \clL^2_q$ (\citet{kovavcik1991spaces}, Thm.~2.11), the closure $\clG$ is roughly the set of kernel smoothed functions in $\clL^2_q$, and it is roughly $\clH^D$.

As mentioned in Section~\ref{sec:parvis}, the particle optimization method (PO)~\cite{chen2017particle} restricts the optimization domain to be quadratic functions when solving the dual optimal transport problem.
Since quadratic functions have restricted sharpness (no change in second-order derivatives), this treatment effectively smooths the function family.

\textbf{Equivalence}~~
The above analysis draws more importance when we note the equivalence between the two smoothing approaches. 
Recall that the objective in the optimization problem that SVGD uses, \eqref{eqn:svgdopt}, is $\lrangle{ \vgf, v }_{\clL^2_q} = \bbE_q[\vgf \cdot v]$.
We generalize this objective in the form $\bbE_q[L(v)]$ with a linear map $L: \clL^2_q \to L^2_q$.
Due to the interchangeability of the integral and the linearity of $L$, we have the relation:
{\setlength\abovedisplayskip{2pt}
\setlength\belowdisplayskip{2pt}
\begin{align}
  \bbE_{\tqq}[L(v)] = \bbE_{q*K}[L(v)] = \bbE_{q}[L(v)*K] = \bbE_{q}[L(v*K)],
\end{align}
}which indicates that smoothing the density $q*K$ is equivalent to smoothing functions $v*K$.
This equivalence bridges the two types of ParVIs, making the analysis and techniques (\eg, our acceleration and bandwidth methods in Secs.~\ref{sec:wnag} and~\ref{sec:bandw}, respectively) for one type also applicable to the other.

\textbf{Necessity and ParVI Assumption}~~
We stress that this smoothing endeavor is essentially required by a well-defined gradient flow of the KL-divergence $\KL_p(q)$.
It returns infinity when $q$ is not absolutely continuous, as is the case $q = \hqq$, thus the gradient flow cannot be reasonably defined. 
So we recognize that ParVIs have to make the assumption that $q$ is smooth, and this can be implemented by either smoothing the density or smoothing functions.

This claim seems straightforward for smoothing density methods, but is perhaps obscure for smoothing function methods.
We now make a direct analysis for SVGD, that neither smoothing the density (take $q=\hqq$) nor smoothing functions (optimize over $\clL^2_p$ \footnote{
  Why not $\clL^2_q$: $v \in \clL^2_p$ is required by the condition of Stein's identity~\cite{liu2017steinflow}, on which SVGD is based.
  Also, $\hqq$ is not absolutely continuous so $\clL^2_{\hqq}$ is not a proper Hilbert space of functions.
}) leads to an unreasonable result.
\begin{theorem}[Necessity of smoothing for SVGD]
  \label{thm:rawsvgd}
  For $q=\hqq$ and $v\in \clL^2_p$, problem~\origeqref{eqn:svgdopt} has no optimal solution.
  In fact the supremum of the objective is infinite, indicating that a maximizing sequence of $v$ tends to be ill-posed.
  \vspace{-5pt}
\end{theorem}
The proof is given in Appendix~A.3.
SVGD claims no assumption on the form of the approximating density $q$ as long as its samples are known, but we find it actually transmits the restriction on $q$ to the functions $v$.
The choice for $v$ in $\clH^D$ is not just for a tractable solution, but more importantly, for guaranteeing a valid vector field. 
We see that there is no free lunch in making the smoothing assumption.
ParVIs have to assume a smoothed density or functions.

\vspace{-5pt}
\subsection{New ParVIs with Smoothing}
\vspace{-5pt}
\label{sec:gf-general}

The theoretical understanding of smoothing for the finite-particle approximation constructs a principle for developing new ParVIs.
We conceive two new instances based on the smoothing-density and smoothing-function formulations.

\textbf{GFSD}~~
We directly approximate $q$ with smoothed density $\tqq := \hqq*K$ and adopt the vector field form of gradient flow (\eqref{eqn:vgf}): 
$ \ugfsd := -\nabla\log \tqq$.
We call the corresponding method Gradient Flow with Smoothed Density (GFSD).

\textbf{GFSF}~~
We discover another novel optimization formulation to identify $\ugf$, which could build a new ParVI by smoothing functions.
We reform the intractable component $\ugf := -\nabla\log q$, as $q\ugf + \nabla q = 0$, and treat it as an equality that holds in the weak sense.
This means $\bbE_q[\phi\cdot u - \nabla \cdot \phi] = 0, \forall \phi\in \clCC$,\footnote{
  We also consider scalar-valued functions $\varphi\in\CC$ smoothed in $\clH$, which gives the same result, as shown in Appendix~B.2.
} or equivalently,
{\setlength\abovedisplayskip{0pt}
\setlength\belowdisplayskip{-1pt}
\begin{align}
  \ugf = \argmin_{u\in\clL^2}
	\max_{\substack{\phi\in\clCC, \\ \lrVert{\phi}_{\clL^2_q}=1}} \big(\bbE_q[\phi\cdot u - \nabla\cdot\phi]\big)^2.
\end{align}
}We take $q=\hqq$ and smooth functions $\phi \in \clCC$ with kernel $K$, which is equivalent to taking $\phi$ from the vector-valued RKHS $\clH^D$ according to Theorem~\ref{thm:smfun}:
{\setlength\abovedisplayskip{3pt}
\setlength\belowdisplayskip{1pt}
\begin{align}
  \ugfsf := \argmin_{u\in\clL^2}
	\max_{\substack{\phi\in\clH^D, \\ \lrVert{\phi}_{\clH^D}=1}} \big(\bbE_{\hqq}[\phi\cdot u - \nabla\cdot\phi]\big)^2.
  \label{eqn:gfsf}
\end{align}
}The closed-form solution is $\hugfsf = \hKp \hK^{-1}$ in matrix form, where $\hugfsf_{:,i} := \ugfsf(x^{(i)})$, $\hK_{ij} := K(x^{(i)}, x^{(j)})$, and $\hKp_{:,i} := \sum_j \nabla_{x^{(j)}} K(x^{(j)}, x^{(i)})$ (see Appendix~B.1).
We call this method the Gradient Flow with Smoothed test Functions (GFSF).
Note that the above objective fits the form $\bbE_q[L(\phi)]$ with $L$ linear, indicating the equivalence to smoothing the density, as discussed in Section~\ref{sec:gf-summary}.
An interesting relation between the matrix-form expression of GFSF and SVGD is that $\hvgfsf = \hgg + \hKp \hK^{-1}$ while $\hvsvgd = \hgg \hK + \hKp$, where $\hgg_{:,i} := \nabla\log p(x^{(i)})$.
We also note that the GFSF estimate of $-\nabla\log q$ coincides with the method of \citet{li2018gradient}, which is derived by Stein's identity and approximating the $\ell^2$ space with RKHS $\clH$.

Due to Remark~\ref{rem:langevin}, all these ParVIs aim to simulate the same path on $\clP_2(\clX)$ as the Langevin dynamics (LD) (\citet{roberts2002langevin}).
They directly utilize the particle interaction via the smoothing kernel, so every particle is aware of others and they could be more particle-efficient \cite{liu2016stein} than the vanilla LD simulation. 
To scale to large datasets, LD has employed stochastic gradient in simulation~\cite{welling2011bayesian}, which is appropriate~\cite{chen2015convergence}.
Due to the connection to LD, ParVIs can also adopt stochastic gradient for scalability.
Finally, our theory could utilize more techniques for developing ParVIs, \eg, implicit distribution gradient estimation \cite{shi2018spectral}.

\vspace{-7pt}
\section{Accelerated First-Order Methods on $\clP_2(\clX)$}
\vspace{-5pt}
\label{sec:wnag}

We have developed a unified understanding on ParVIs under the $\clP_2(\clX)$ gradient flow perspective, which corresponds to the gradient descent method on $\clP_2(\clX)$.
It is well-known that the Nesterov's acceleration method~\cite{nesterov1983method} can give a faster convergence rate, and its Riemannian variants have been developed recently, such as Riemannian Accelerated Gradient (RAG)~\cite{liu2017accelerated} and Riemannian Nesterov's method (RNes)~\cite{zhang2018estimate}.
We aim to employ 
ParVIs with these methods.
However, this requires more knowledge on the geometry of $\clP_2(\clX)$.

\vspace{-5pt}
\subsection{Leveraging the Riemannian Structure of $\clP_2(\clX)$}
\vspace{-5pt}
\label{sec:exc}

\begin{figure}[t]\vspace{-2pt}
  \centering
  \includegraphics[width=.2\textwidth]{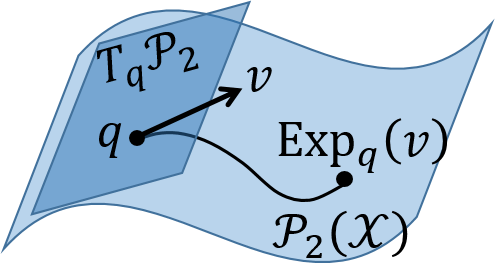}
  \includegraphics[width=.21\textwidth]{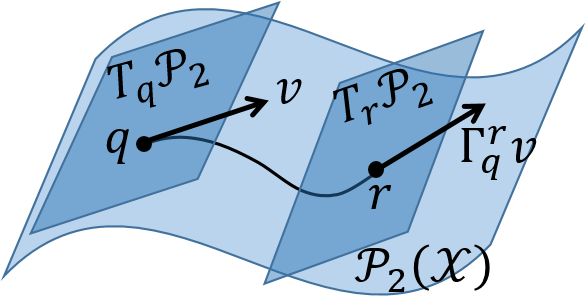}\vspace{-5pt}
  \caption{Illustration of the concepts exponential map (left) and parallel transport (right)}
  \label{fig:exppara}
  \vspace{-10pt}
\end{figure}

RAG and RNes require the \emph{exponential map} (and its inverse) and \emph{parallel transport} 
on $\clP_2(\clX)$.
As depicted in Fig.~\ref{fig:exppara}, the exponential map $\Exp_{q}: T_{q}\clP_2(\clX) \to \clP_2(\clX)$ is the displacement from position $q$ to a new position along the geodesic (a ``straight line'' on a manifold) with a given direction, and the parallel transport $\Gamma_q^r: T_q\clP_2(\clX) \to T_r\clP_2(\clX)$ is the transformation of a tangent vector at $q$ to another one at $r$ in a certain sense of parallel along the geodesic from $q$ to $r$.
We now examine these components and find practical estimation with a finite set of particles.

\textbf{Exponential Map on $\clP_2(\clX)$}~~
For an absolutely continuous measure $q$, $\Exp_q(v) = (\id + v)_{\#}q$ (\citet{villani2008optimal}, Coro.~7.22; \citet{ambrosio2008gradient}, Prop.~8.4.6; \citet{erbar2010heat}, Prop.~2.1). 
Finite-particle estimation is just an application of the map $x \mapsto x + v(x)$ on each particle.

\textbf{Inverse of the Exponential Map}~~
With details in Appendix~A.4, we find that the inverse exponential map $\Exp^{-1}_q (r)$ for $q, r \in \clP_2(\clX)$ can be expressed by the optimal transport map from $q$ to $r$.
We can approximate the map by a discrete one between particles $\{x^{(i)}\}_{i=1}^N$ of $q$ and $\{y^{(i)}\}_{i=1}^N$ of $r$.
However, this is still a costly task \cite{pele2009fast}.
Faster approaches like the Sinkhorn method~\cite{cuturi2013sinkhorn, xie2018fast} still require empirically $O(N^2)$ time, and it appears unacceptably unstable in our experiments.
We consider an approximation when $\{x^{(i)}\}_i$ and $\{y^{(i)}\}_i$ are pairwise close: $d(x^{(i)}, y^{(i)}) \ll \min \big\{ \min_{j \neq i} d(x^{(i)}, x^{(j)}), \min_{j \neq i} d(y^{(i)}, y^{(j)}) \big\}$.
In this case we have the result (deduction in Appendix~A.4):
\begin{proposition}[Inverse exponential map]
  \label{prop:invexp}
  For pairwise close samples $\{x^{(i)}\}_{i=1}^N$ of $q$ and $\{y^{(i)}\}_{i=1}^N$ of $r$, we have $\big( \Exp^{-1}_q (r) \big) (x^{(i)}) \approx y^{(i)} - x^{(i)}$.
  \vspace{-5pt}
\end{proposition}

\textbf{Parallel Transport on $\clP_2(\clX)$}~~
There has been formal research on the parallel transport on $\clP_2(\clX)$~\cite{lott2008some, lott2017intrinsic}, but the result is expensive to estimate with a finite set of particles. 
Here we utilize Schild's ladder method~\cite{ehlers1972geometry, kheyfets2000schild}, a first-order approximation of parallel transport that only requires the exponential map.
We derive the following estimation with a finite set of particles (deduction in Appendix~A.5):
\begin{proposition}[Parallel transport]
  \label{prop:para}
  For pairwise close samples $\{x^{(i)}\}_{i=1}^N$ of $q$ and $\{y^{(i)}\}_{i=1}^N$ of $r$, we have $\big( \Gamma_q^r (v) \big) (y^{(i)}) \approx v(x^{(i)})$, $\forall v\in T_q \clP_2(\clX)$.
\end{proposition}

Both results may not seem surprising.
This is because the geometry of $\clP_2(\clX)$ is determined by that of $\clX$.
We consider Euclidean $\clX$, so $\clP_2(\clX)$ also appears flat.
Extension to non-flat Riemannian $\clX$ can be done with the same procedure.

\vspace{-5pt}
\subsection{Acceleration Framework for ParVIs}
\vspace{-5pt}
\label{sec:acc}

\begin{algorithm}[t]
  \caption{The acceleration framework with Wasserstein Accelerated Gradient (WAG) and Wasserstein Nesterov's method (WNes)}
  \label{alg:wacc}
  \begin{algorithmic}[1]
	\STATE \mbox{WAG: select acceleration factor $\alpha>3$;}\\
		   \mbox{WNes: select or calculate $c_1, c_2 \in \bbR^+$ (Appendix~C.2);}
	\STATE Initialize $\{x_0^{(i)}\}_{i=1}^N$ distinctly; let $y_0^{(i)} = x_0^{(i)}$;
	\FOR{$k=1,2,\cdots,k_{\mathrm{max}},$}
	  \FOR{$i=1,\cdots,N$,}
		\STATE \mbox{Find $v(y^{(i)}_{k-1})$ by SVGD/Blob/GFSD/GFSF;}
		\STATE \mbox{$x^{(i)}_{k} = y^{(i)}_{k-1} + \varepsilon v(y^{(i)}_{k-1})$;}
		\STATE $y^{(i)}_k = x^{(i)}_{k} + $
		  {\setlength\abovedisplayskip{2pt}
		  \setlength\belowdisplayskip{2pt}
		  \begin{align}
			\begin{cases}
			  \!\text{WAG:}\; \frac{k-1}{k}(y^{(i)}_{k-1} \!-\! x^{(i)}_{k-1}\!) \!+\! \frac{k+\alpha-2}{k}\varepsilon v(y^{(i)}_{k-1});\\
			  \!\text{WNes:}\;\; c_1 (c_2 - 1) (x^{(i)}_k - x^{(i)}_{k-1});
			\end{cases}
		  \end{align}}
	  \ENDFOR
	\ENDFOR
	\STATE Return $\{x^{(i)}_{k_{\mathrm{max}}}\}_{i=1}^N$.
  \end{algorithmic}
\end{algorithm}

Now we apply RAG and RNes to the Wasserstein space $\clP_2(\clX)$ and construct an accelerated sequence $\{q_k\}_k$ minimizing $\KL_p$.
Both methods introduce an auxiliary variable $r_k \in \clP_2(\clX)$, on which the gradient is evaluated: $v_{k} := -\grad \KL(r_{k})$.
RAG \cite{liu2017accelerated} needs to solve a nonlinear equation in each step.
We simplify it with moderate approximations to give an explicit update rule:
$q_{k} \!=\! \Exp_{r_{k-1}} \! (\varepsilon v_{k-1}), r_{k} \!=\! \Exp_{q_{k}} \! \left[ -\Gamma_{r_{k-1}}^{q_{k}} \! \left( \frac{k-1}{k} \Exp^{-1}_{r_{k-1}} \! (q_{k-1}) - \frac{k+\alpha-2}{k} \varepsilon v_{k-1} \right) \right]$ (details in Appendix~C.1).
RNes \cite{zhang2018estimate} involves an additional variable.
We collapse the variable and reformulate RNes as: $q_{k} \!=\! \Exp_{r_{k-1}} \! (\varepsilon v_{k-1}), r_{k} \!=\! \Exp_{q_{k}} \! \big\{ c_1 \Exp^{-1}_{q_{k}} \! \big[ \Exp_{r_{k-1}} \! \big( (1 \!-\! c_2) \Exp^{-1}_{r_{k-1}} \! (q_{k-1}) \!+\! c_2 \Exp^{-1}_{r_{k-1}} \! (q_{k}) \big) \big] \big\}$ (details in Appendix~C.2).
To implement both methods with a finite set of particles, we leverage the geometric calculations in the previous subsection and estimate $v_k$ with ParVIs.
The resulting algorithms are called Wasserstein Accelerated Gradient (WAG) and Wasserstein Nesterov's method (WNes), and are presented in Alg.~\ref{alg:wacc} (deduction details in Appendix~C.3).
They form an acceleration framework for ParVIs.
In the deduction, the pairwise-close condition is satisfied,
so the usage of Propositions~\ref{prop:invexp} and~\ref{prop:para} is appropriate.

In theory, the acceleration framework inherits the proved improvement on the convergence rate from RAG and RNes, and it can be applied to all ParVIs, since our theory has recognized the equivalence of ParVIs.
In practice, the framework imposes a computational cost linear in the particle size $N$, which is not a significant overhead.
Moreover, we emphasize that we cannot directly apply the vanilla Nesterov's acceleration method~\cite{nesterov1983method} in $\clX$ on every particle, since a single particle is not optimizing a certain function.
Finally, the developed knowledge on the geometry of $\clP_2(\clM)$ (Propositions~\ref{prop:invexp},~\ref{prop:para}) also makes it possible to apply other optimization techniques on Riemannian manifolds to benefit ParVIs, \eg, Riemannian BFGS~\cite{gabay1982minimizing, qi2010riemannian, yuan2016riemannian} and Riemannian stochastic variance reduction gradient~\cite{zhang2016riemannian}.

\vspace{-5pt}
\section{Bandwidth Selection via the Heat Equation}
\vspace{-5pt}
\label{sec:bandw}

Our theory points out that all ParVIs need smoothing, and this can be done with a kernel.
Thus it is an essential problem to choose the bandwidth of the kernel.
SVGD uses the median method~\cite{liu2016stein} based on a heuristic for numerical stability.
We work towards a more principled method.
We first analyze the goal of smoothing, then build a practical algorithm.

As noted in Remark~\ref{rem:langevin}, the deterministic dynamics $\ud x = \vgf(x) \dd t$ and the Langevin dynamics produce the same rule of density evolution.
In particular, the deterministic dynamics $\ud x = -\nabla \log q_t(x) \dd t$ and the Brownian motion $\ud x = \sqrt{2} \dd B_t(x)$ produce the same evolution rule specified by the heat equation (HE): $\partial_t q_t(x) = \Delta q_t(x)$.
So a good smoothing kernel should let the evolving density under the approximated dynamics match the rule of HE.
This is the principle for selecting a kernel bandwidth.

We implement this principle for GFSD, which estimates $q_t$ by the kernel smoothed density $\tqq(x) = \tqq(x; \{x^{(i)}\}_i) = \frac{1}{N} \sum_{i=1}^N K(x, x^{(i)})$.
The approximate dynamics $\ud x = -\nabla \log \tqq(x) \dd t$ moves particles $\{x^{(i)}\}_i$ of $q_t$ to $\{x^{(i)} - \varepsilon \nabla \log \tqq(x^{(i)}) \}_i$, which approximates $q_{t+\varepsilon}$.
On the other hand, according to the evolution rule of the HE, the new density $q_{t+\varepsilon}$ should be approximated by $q_t + \varepsilon \partial_t q_t \approx \tqq + \varepsilon \Delta \tqq$.
The two approximations should match, which means $\tqq \big(x; \{x^{(i)} - \varepsilon \nabla \log \tqq(x^{(i)}) \}_i \big)$ should be close to $\tqq + \varepsilon \Delta \tqq$.
Expanding up to first order in $\varepsilon$, this requirement translates to imposing that the function $\lambda(x) := \Delta\tqq(x; \{x^{(i)}\}_i) + \sum_j \nabla_{x^{(j)}} \tqq(x; \{x^{(i)}\}_i) \cdot \nabla\log\tqq(x^{(j)}; \{x^{(i)}\}_i)$ should be close to zero.
To achieve this practically, we propose to minimize $\frac{N}{h^{D+2}} \bbE_{q(x)} [\lambda(x)^2] \approx \frac{1}{h^{D+2}} \sum_k \lambda(x^{(k)})^2$ w.r.t. the bandwidth $h$.
We introduce the factor $\frac{1}{h^{D+2}}$ (note that $x^2/h$ is dimensionless) to make the final objective dimensionless. 

We call this the HE method.
Although the derivation is based on GFSD, the resulting algorithm can be applied to all kernel-based ParVIs, like SVGD, Blob and GFSF, due to the equivalence of smoothing the density and functions from our theory.
See Appendix~D for further details.

\vspace{-8pt}
\section{Experiments}
\vspace{-5pt}

Detailed experimental settings and parameters are provided in Appendix~E, and codes are available at \url{https://github.com/chang-ml-thu/AWGF}.
Corresponding to WAG and WNes, we call the vanilla gradient flow simulation of ParVIs as Wasserstein Gradient Descent (WGD).

\vspace{-8pt}
\subsection{Toy Experiments}
\vspace{-5pt}

\begin{wrapfigure}{r}{.26\textwidth}
  \vspace{-24pt}
  \centering
  \includegraphics[width=.13\textwidth]{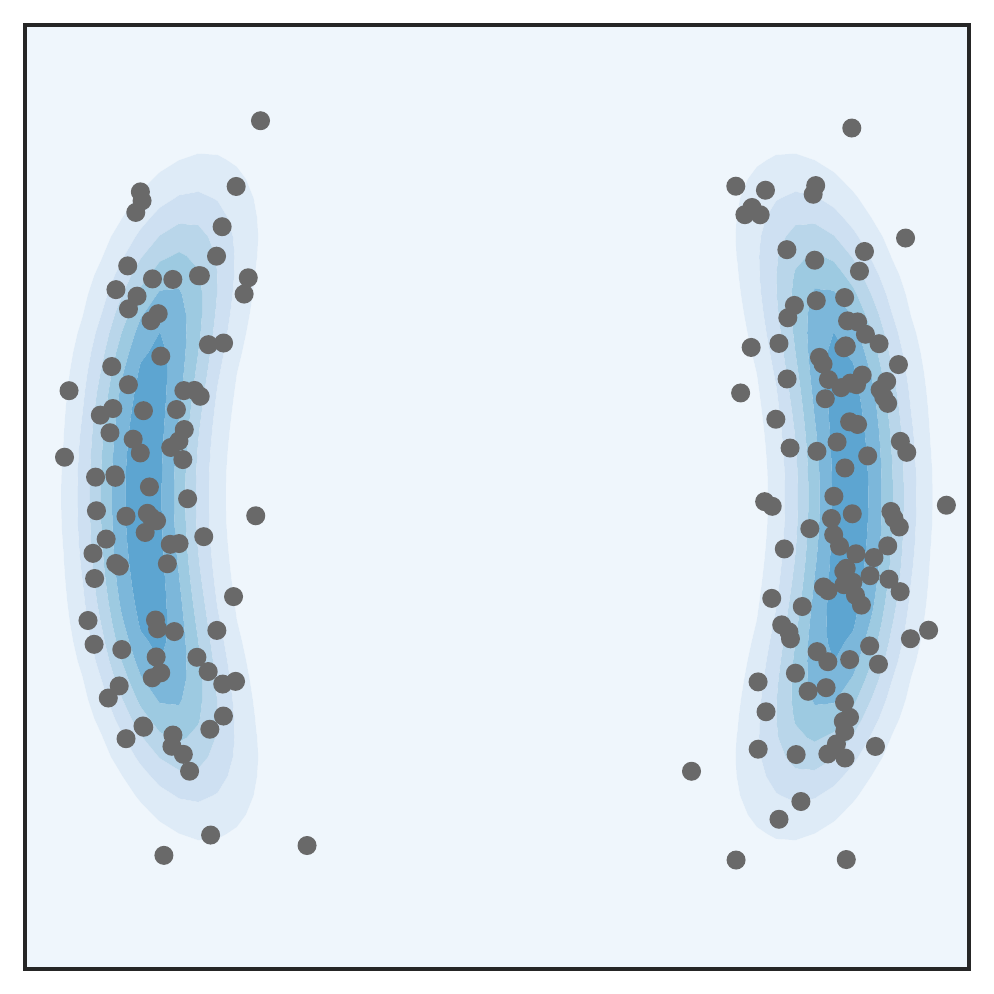}\hspace{-4pt}
  \includegraphics[width=.13\textwidth]{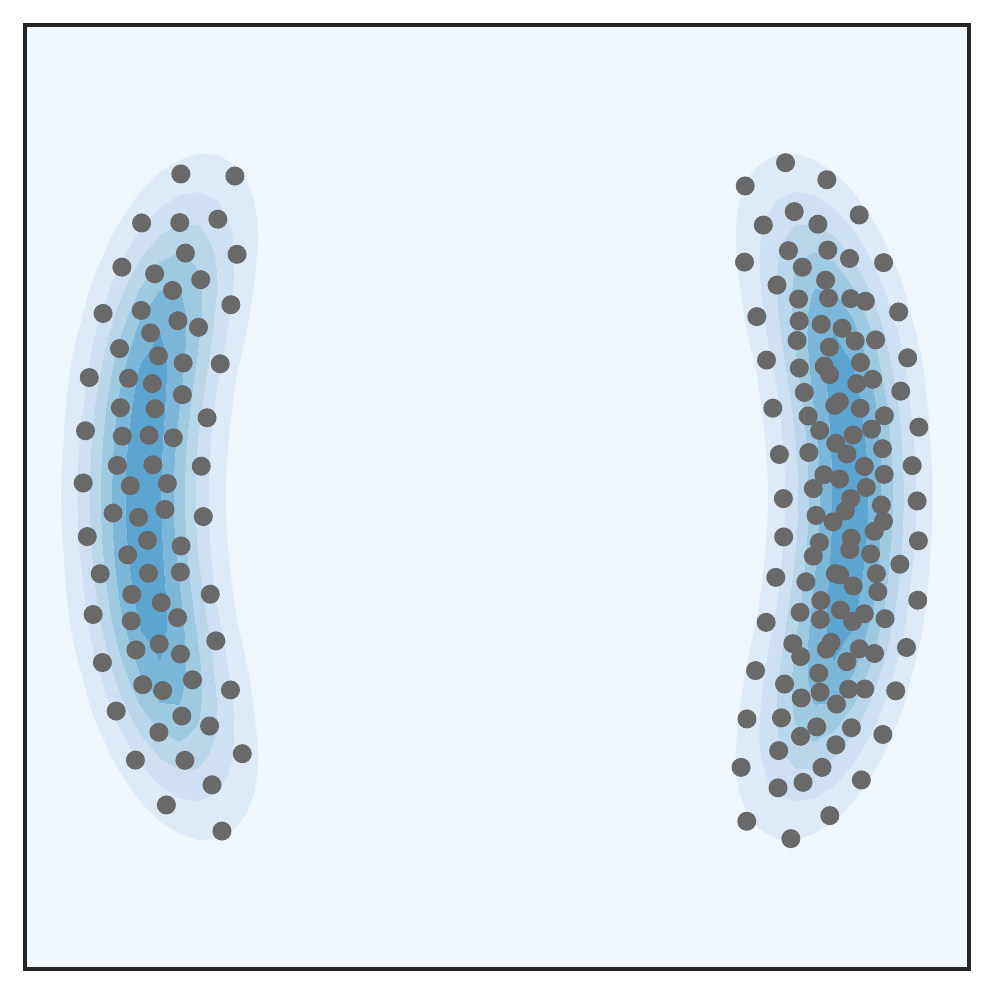}
  \\[-2.5pt]
  \includegraphics[width=.13\textwidth]{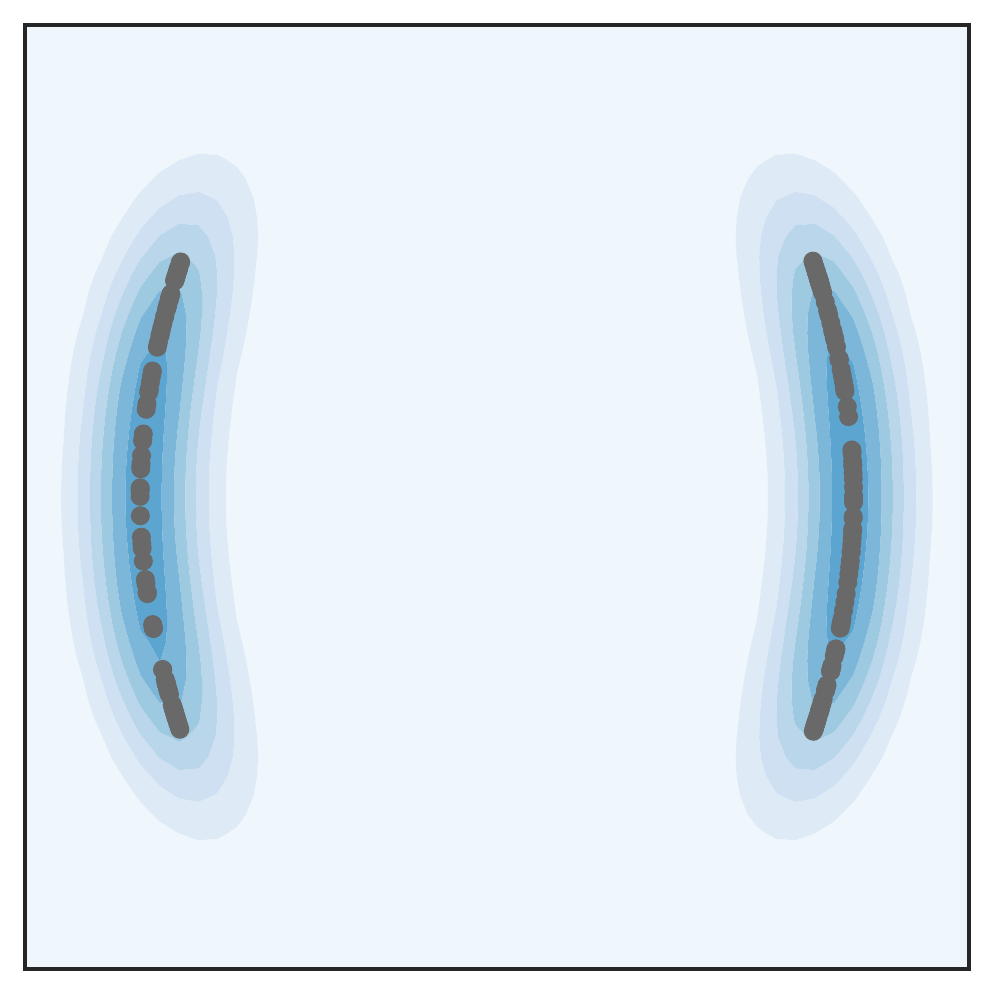}\hspace{-4pt}
  \includegraphics[width=.13\textwidth]{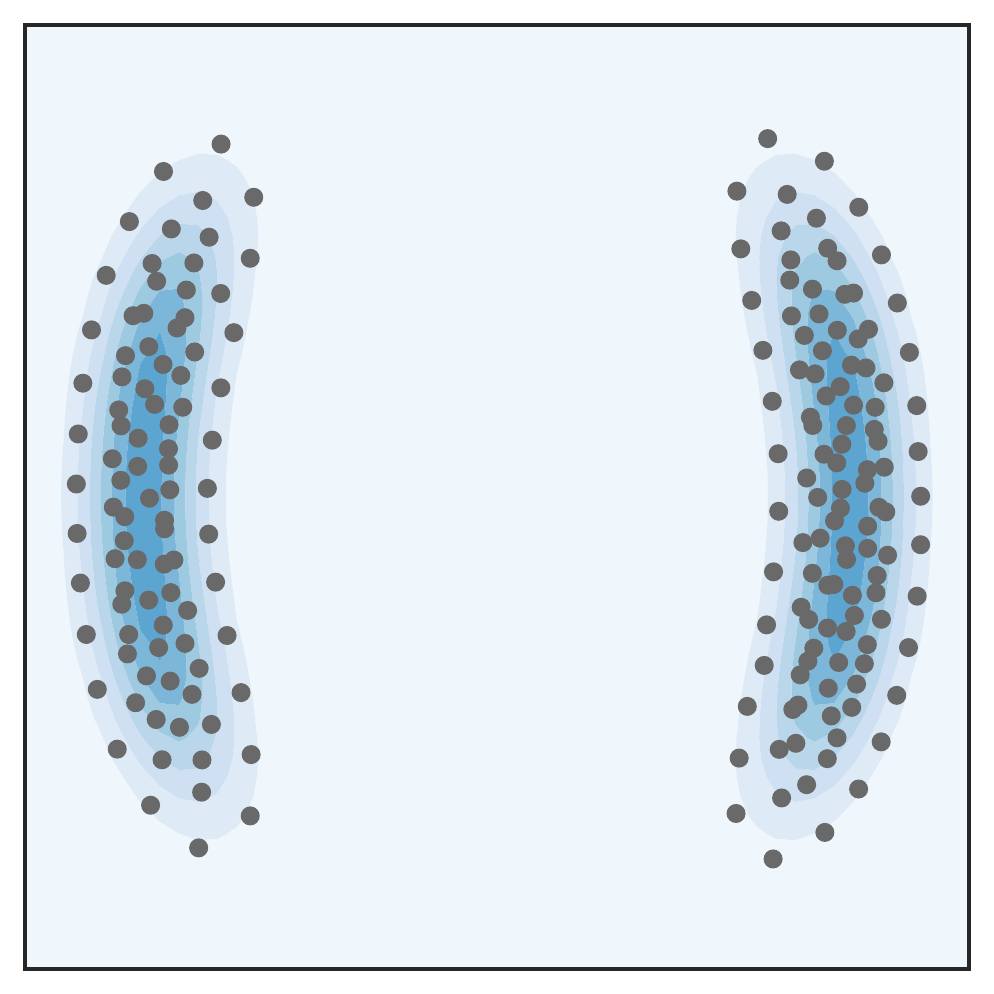}
  \\[-2.5pt]
  \includegraphics[width=.13\textwidth]{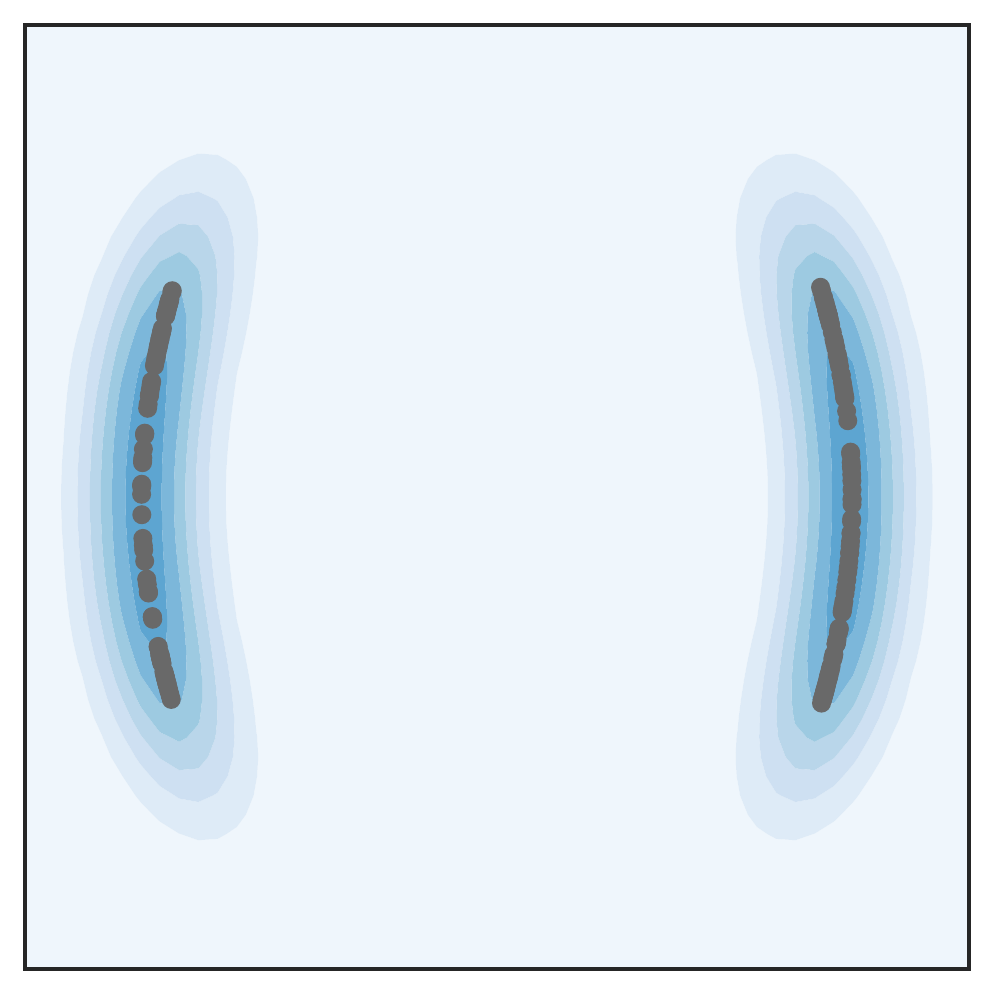}\hspace{-4pt}
  \includegraphics[width=.13\textwidth]{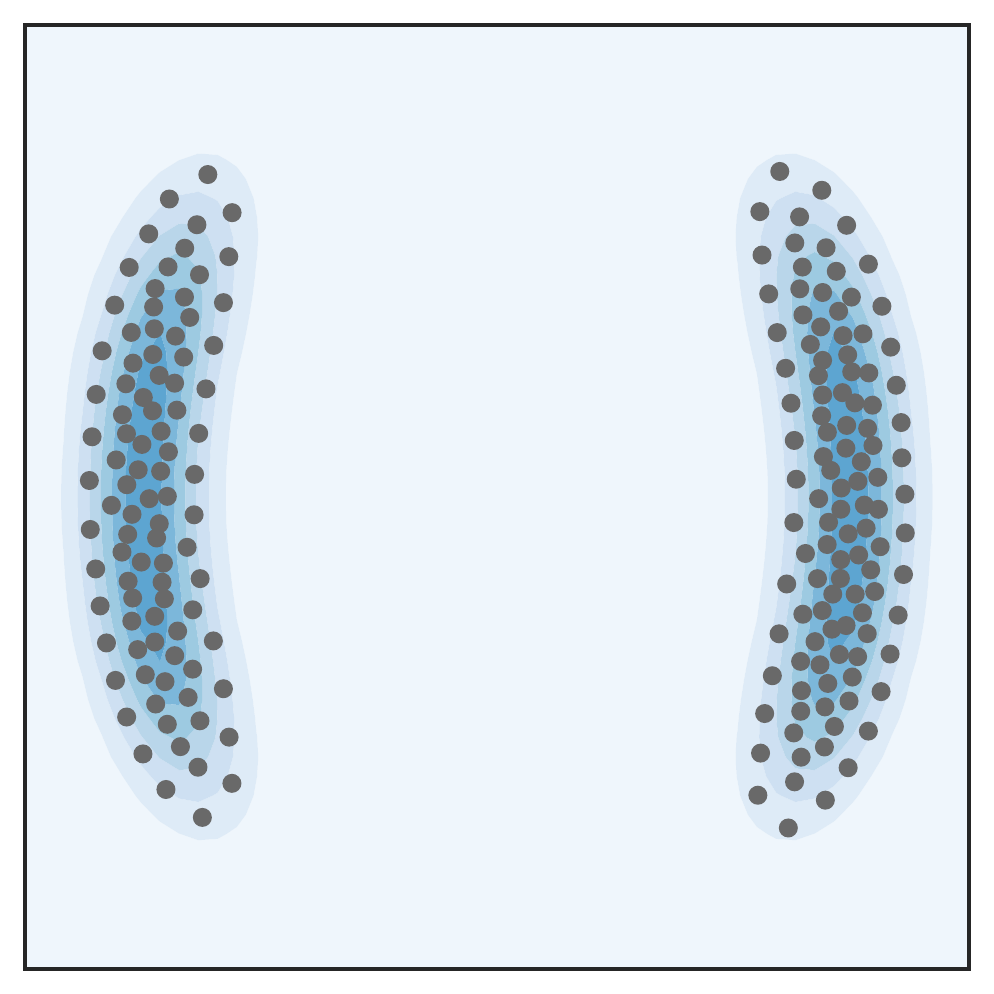}
  \\[-2.7pt]
  \includegraphics[width=.13\textwidth]{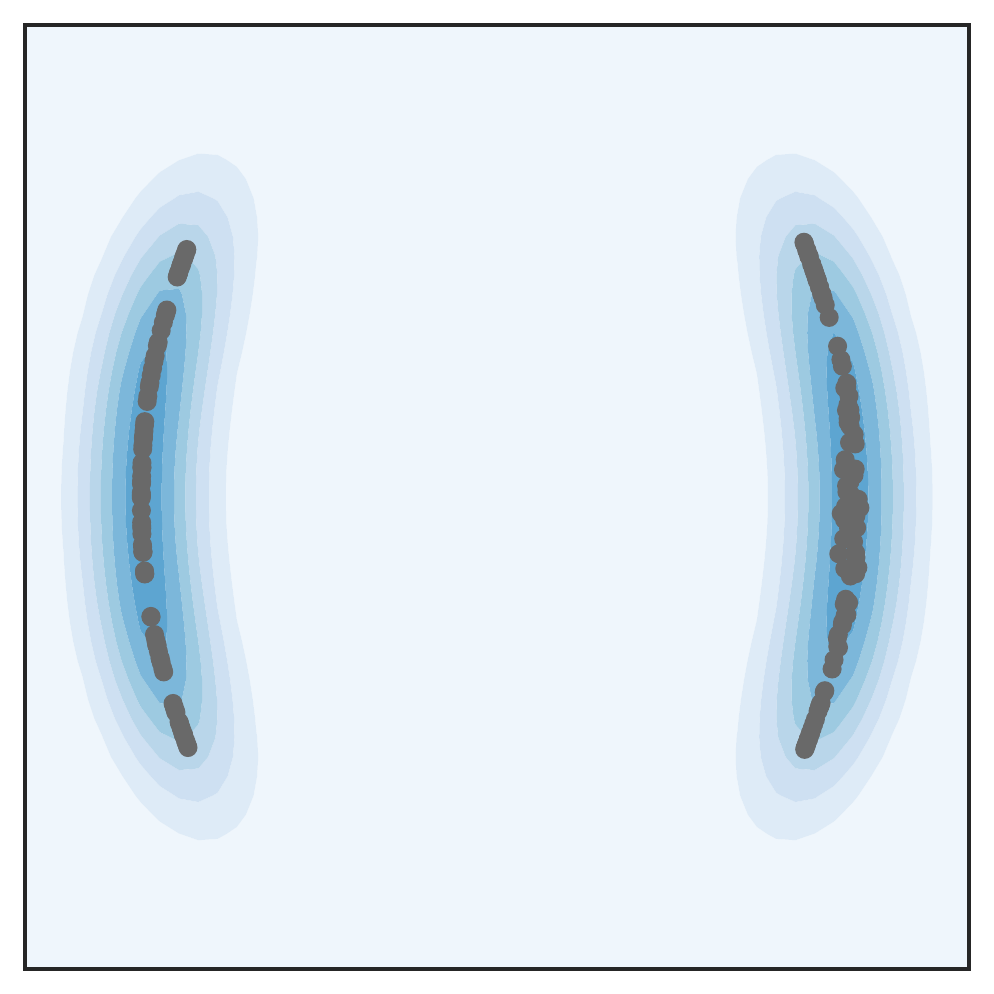}\hspace{-4pt}
  \includegraphics[width=.13\textwidth]{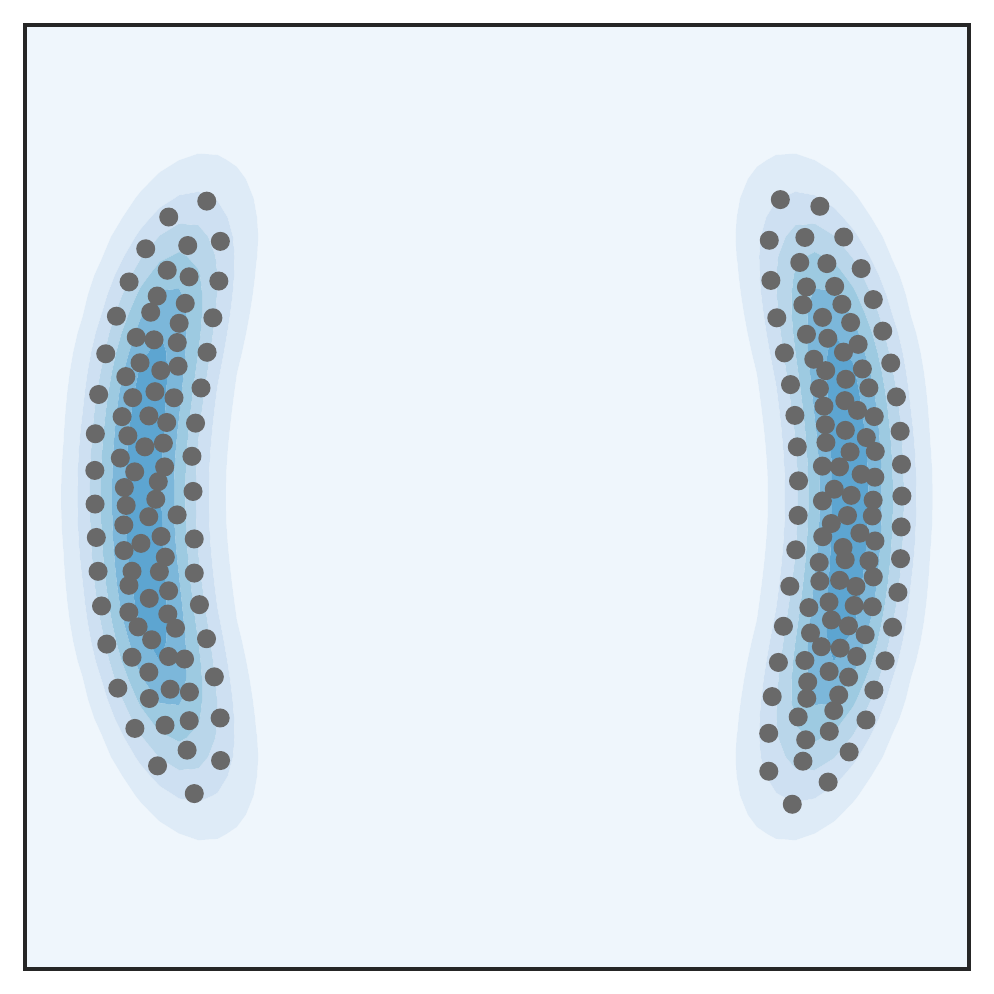}
  \\[-2.7pt]
  \caption{Comparison of HE (right column) with the median method (left column) for bandwidth selection. 
	Rows correspond to SVGD, Blob, GFSD and GFSF, respectively.
	\vspace{-10pt}
  }
  \label{fig:toy}
  \vspace{-10pt}
\end{wrapfigure}
We first investigate the benefit of the HE method for selecting the bandwidth, with comparison to the median method.
Figure~\ref{fig:toy} shows 200 particles produced by four ParVIs using both the HE and median methods. 
We find that the median method makes the particles collapse to the modes, since the numerical heuristic cannot guarantee the effect of smoothing.
The HE method achieves an attractive effect: the particles align neatly and distribute almost uniformly along the contour, 
building a more representative approximation.
SVGD gives diverse particles also with the median method, which may be due to the averaging of gradients for each particle. 

\vspace{-5pt}
\subsection{Bayesian Logistic Regression (BLR)}
\vspace{-5pt}

We show the accelerated convergence of the proposed WAG and WNes methods (Alg.~\ref{alg:wacc}) for various ParVIs on BLR.
Although PO is not developed for acceleration, we treat it as an empirical acceleration.
We follow the same settings as \citet{liu2016stein} and \citet{chen2018unified}, except that results are averaged over 10 random trials.
Results are evaluated by test accuracy (Fig.~\ref{fig:blr-acc}) and log-likelihood (Fig.~\ref{fig:blr-llh} in Appendix~F.1). 
For all four ParVIs, WAG and WNes notably improve the convergence over WGD and PO.
Moreover, WNes gets better results than WAG, especially at the early stage, and it is also more stable w.r.t. hyperparameters.
The performance of the PO method is roughly the same as WGD, matching the observation by \citet{chen2017particle}.
We also note that the four ParVIs have a similar performance, which is natural since they approximate the same gradient flow with equivalent smoothing treatments.

\begin{figure}[!tb]\vspace{-5pt}
  \centering
  \hspace{-5pt}
  \subfigure[SVGD]{
	\includegraphics[scale=.23]{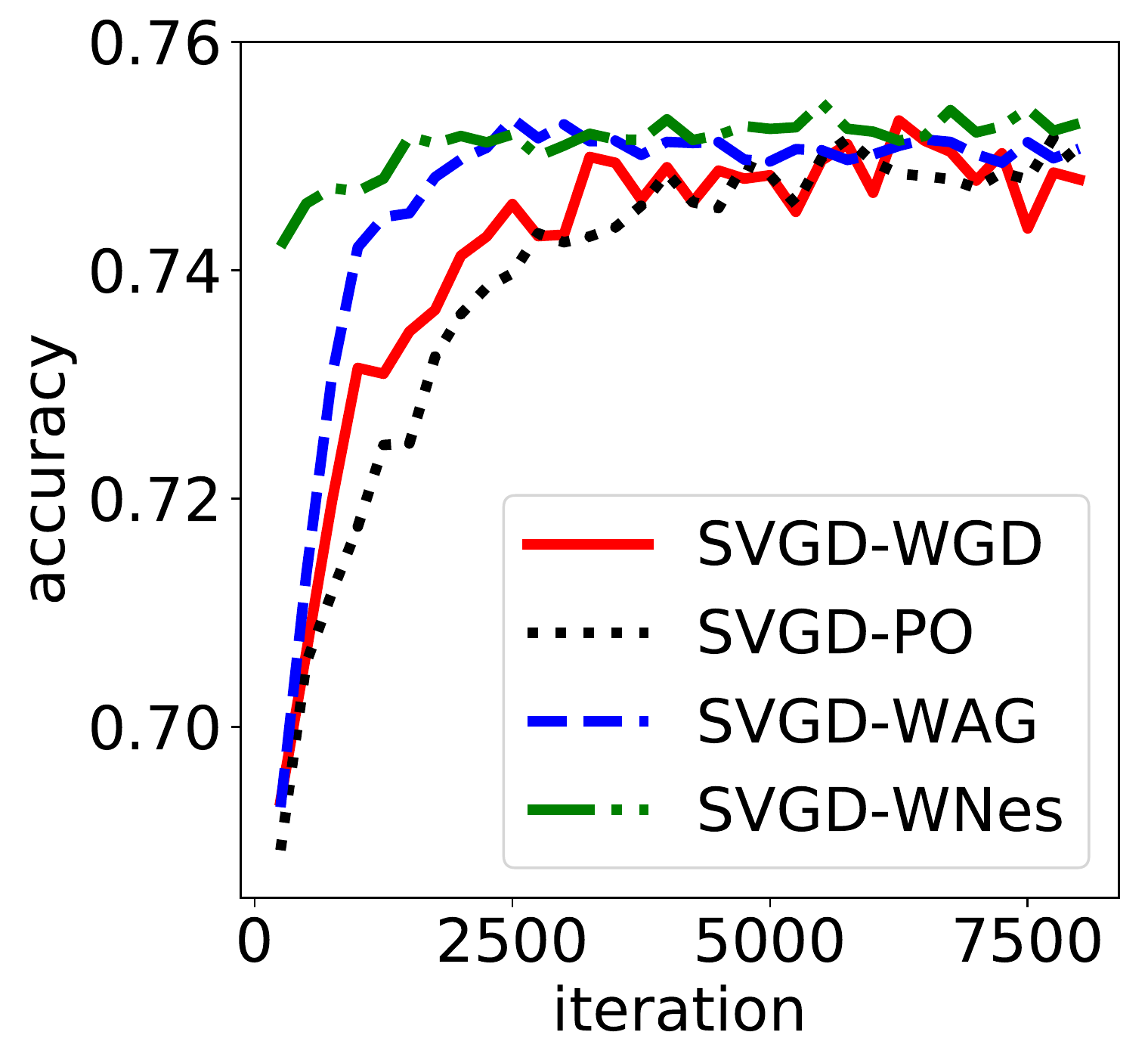}\vspace{-8pt}
	\label{fig:blrsvgd}
  }
  \hspace{-5pt}
  \subfigure[Blob]{
	\includegraphics[scale=.23]{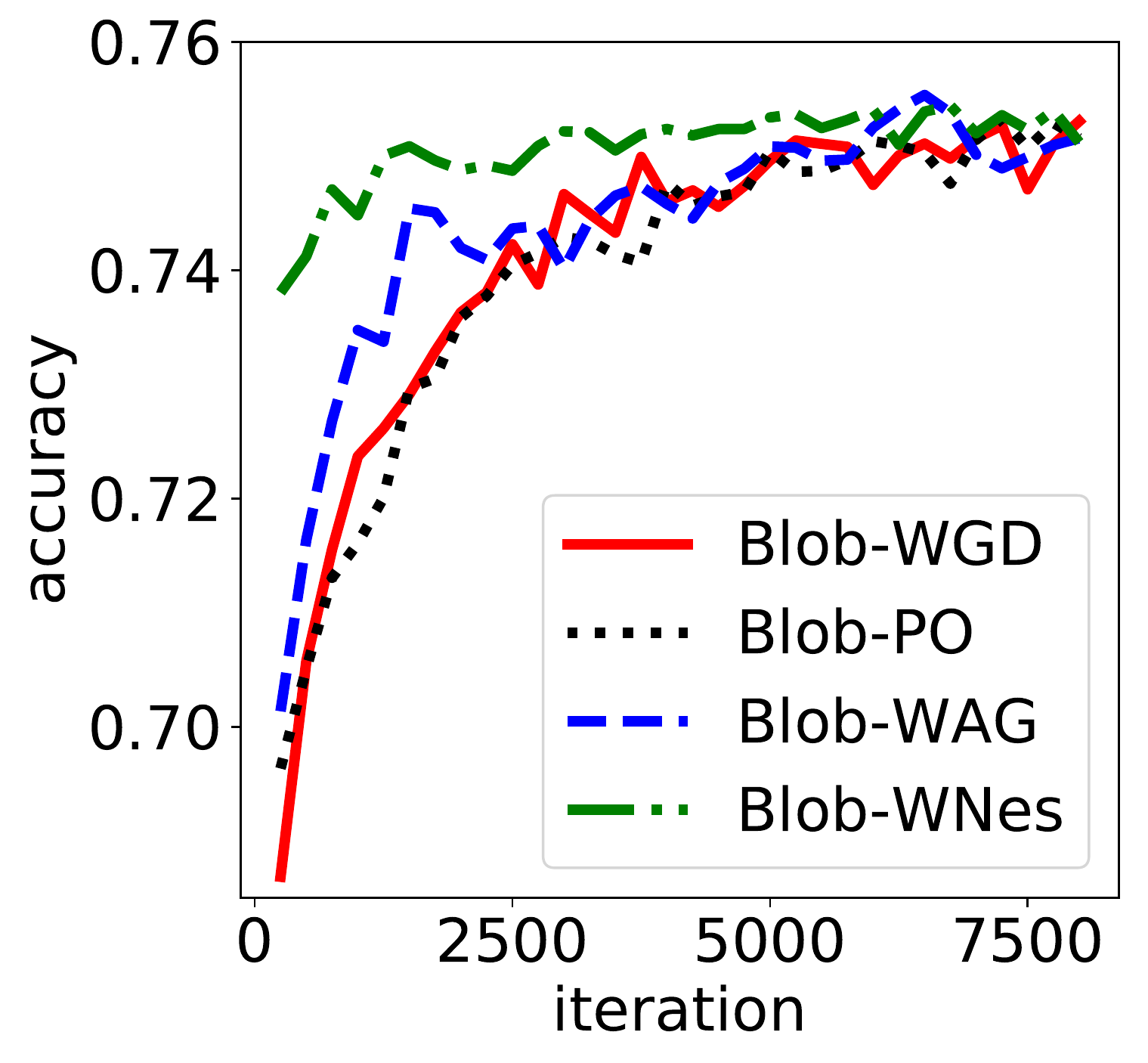}\vspace{-8pt}
    \label{fig:blrblob}
  }
  \\[-8pt]
  \hspace{-5pt}
  \subfigure[GFSD]{
	\includegraphics[scale=.23]{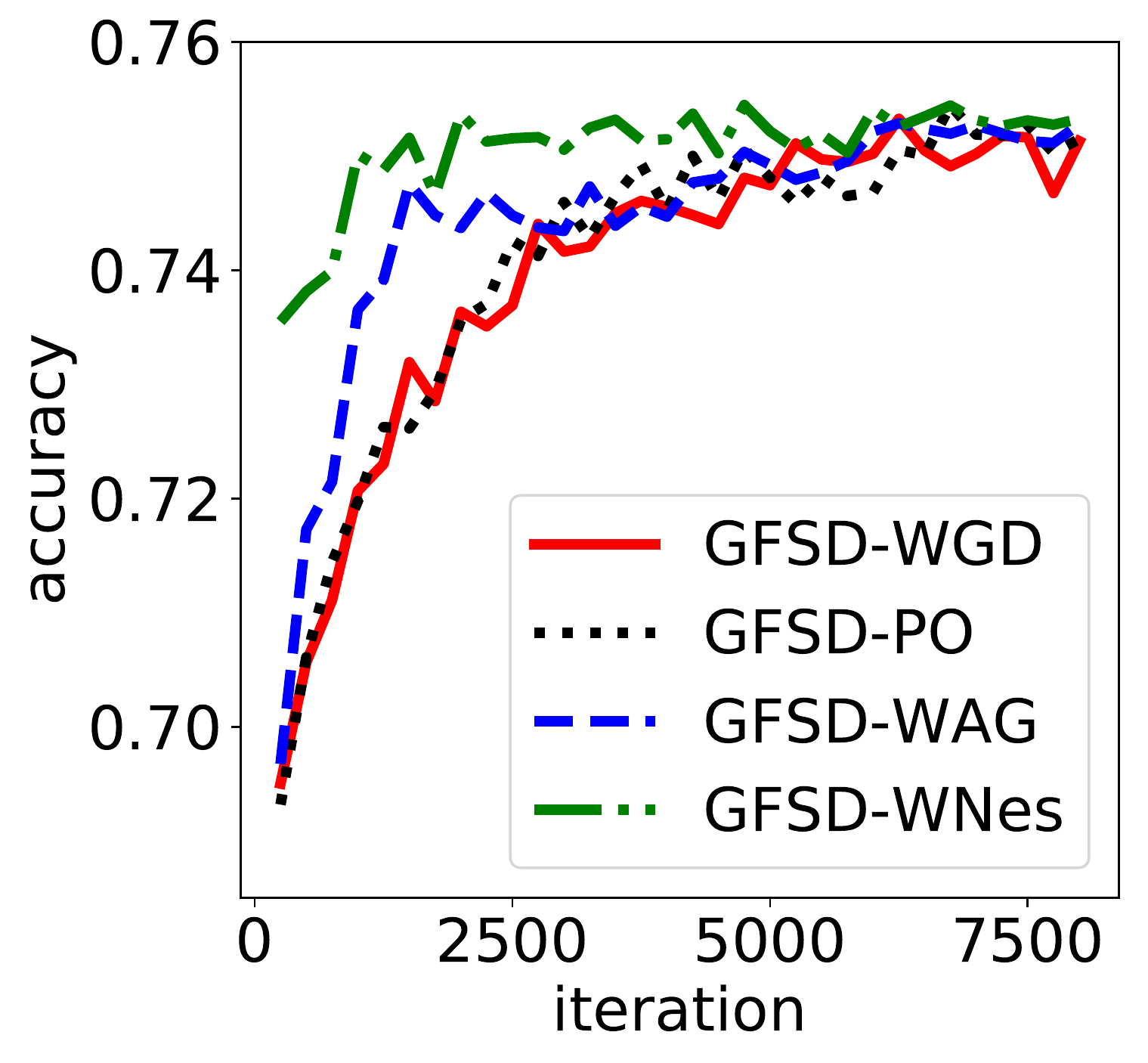}\vspace{-8pt}
	\label{fig:blrgfsd}
  }
  \hspace{-5pt}
  \subfigure[GFSF]{
	\includegraphics[scale=.23]{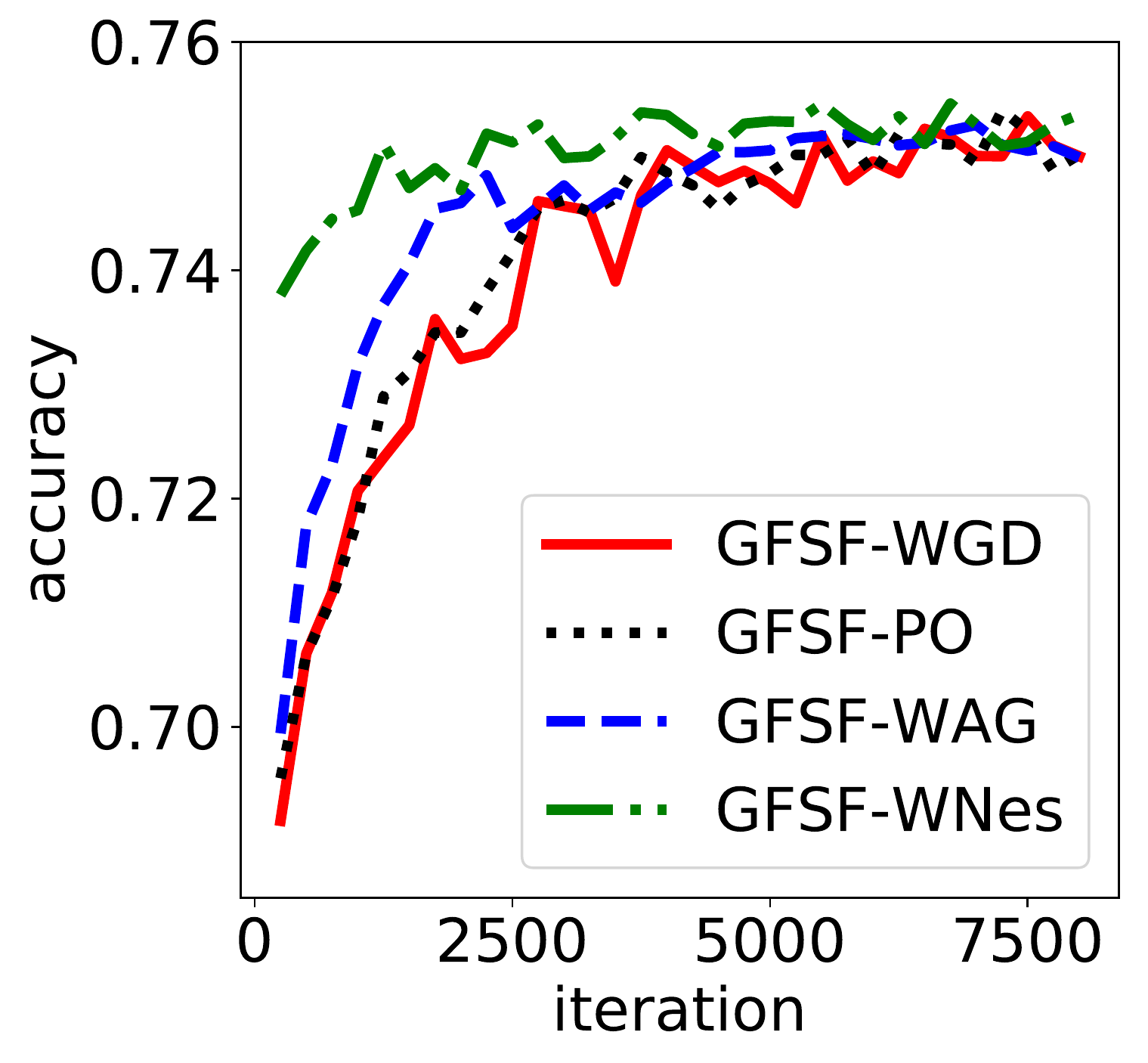}\vspace{-8pt}
	\label{fig:blrgfsf}
  }
  \vspace{-8pt}
  \caption{Acceleration effect of WAG and WNes on BLR on the Covertype dataset. Curves are averaged over 10 runs.} 
  \label{fig:blr-acc}
  \vspace{-14pt}
\end{figure}

\vspace{-5pt}
\subsection{Bayesian Neural Networks (BNNs)}
\vspace{-5pt}

\begin{table*}[h] 
  \vspace{-4pt}
  \caption{Results on BNN on the Kin8nm dataset (one of the UCI datasets \cite{asuncion2007uci}). Results averaged over 20 runs.} 
  \label{tab:bnnkin}
  \centering
  \footnotesize
  \begin{tabular}{ccccccccc}
	\toprule
	\multirow{2}{*}{Method} & \multicolumn{4}{c}{Avg. Test RMSE ($\e{-2}$)} & \multicolumn{4}{c}{Avg. Test LL} \\
	\cmidrule(lr){2-5} \cmidrule(lr){6-9}
	& SVGD & Blob & GFSD & GFSF & SVGD & Blob & GFSD & GFSF \\
	\midrule
	WGD			& 8.4$\pm$0.2 & 8.2$\pm$0.2 & 8.0$\pm$0.3 & 8.3$\pm$0.2 & 1.042$\pm$0.016 & 1.079$\pm$0.021 & 1.087$\pm$0.029 & 1.044$\pm$0.016 \\
	PO			& 7.8$\pm$0.2 & 8.1$\pm$0.2 & 8.1$\pm$0.2 & 8.0$\pm$0.2 & 1.114$\pm$0.022 & 1.070$\pm$0.020 & 1.067$\pm$0.017 & 1.073$\pm$0.016 \\
	WAG			& 7.0$\pm$0.2 & {\bf 7.0$\pm$0.2} & 7.1$\pm$0.1 & 7.0$\pm$0.1 & 1.167$\pm$0.015 & {\bf 1.169$\pm$0.015} & 1.167$\pm$0.017 & 1.190$\pm$0.014 \\
	WNes		& {\bf 6.9$\pm$0.1} & 7.0$\pm$0.2 & {\bf 6.9$\pm$0.1} & {\bf 6.8$\pm$0.1} & {\bf 1.171$\pm$0.014} & 1.168$\pm$0.014 & {\bf 1.173$\pm$0.016} & {\bf 1.193$\pm$0.014} \\
	\bottomrule
  \end{tabular}
  \vspace{-8pt}
\end{table*}

We test all methods on BNNs for a fixed number of iterations, following the settings of \citet{liu2016stein}, and present results in Table~\ref{tab:bnnkin}.
We observe that WAG and WNes acceleration methods outperform the WGD and PO for all the four ParVIs. 
The PO method also improves the performance, but not to the extent of WAG.

\vspace{-5pt}
\subsection{Latent Dirichlet Allocation (LDA)}
\vspace{-5pt}

\begin{figure}[tb]\vspace{-5pt}
  \centering
  \hspace{-5pt}
  \subfigure[SVGD]{
	\includegraphics[scale=.23]{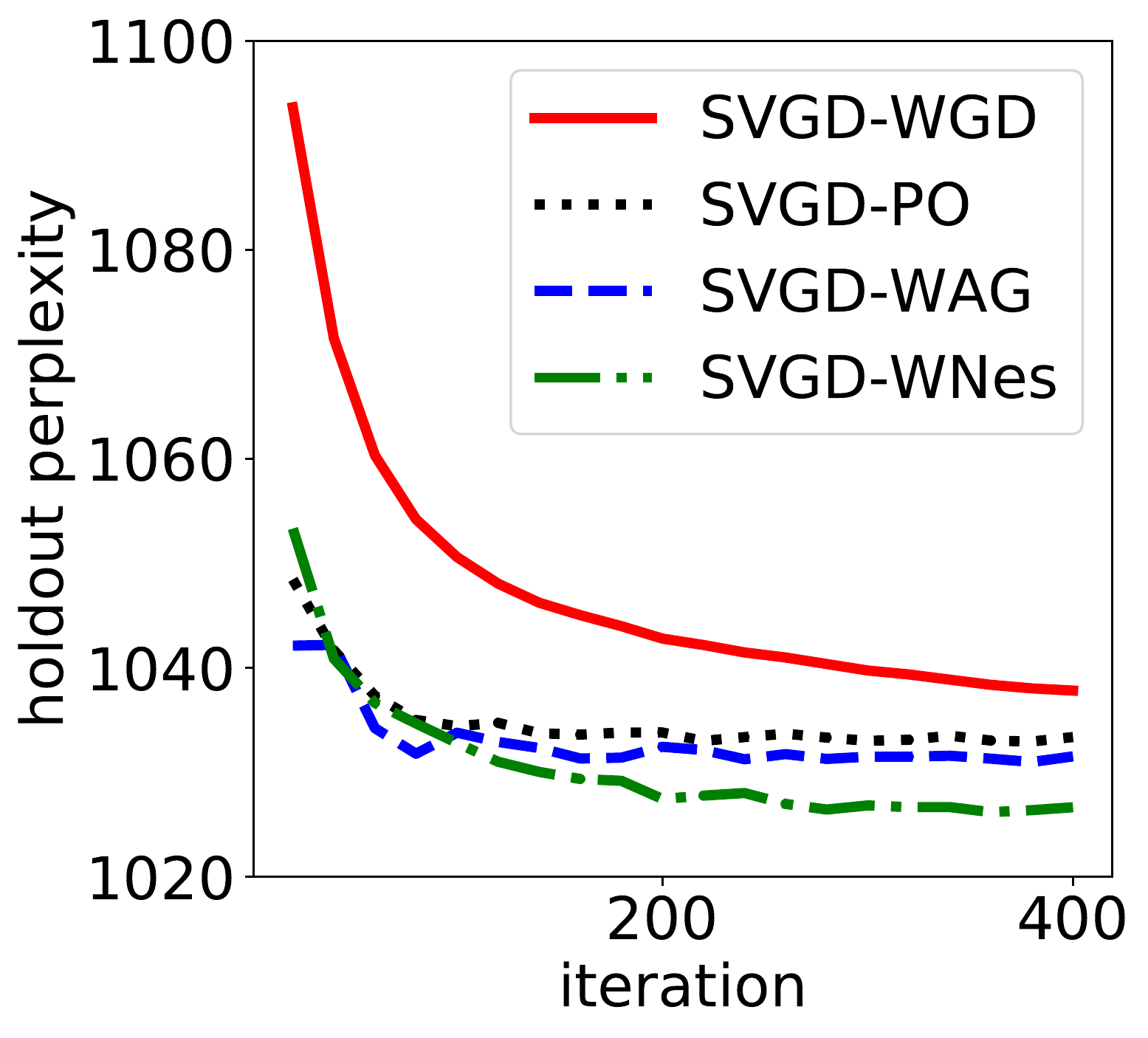}\vspace{-4pt}
	\label{fig:ldasvgd}
  }
  \hspace{-5pt}
  \subfigure[Blob]{
	\includegraphics[scale=.23]{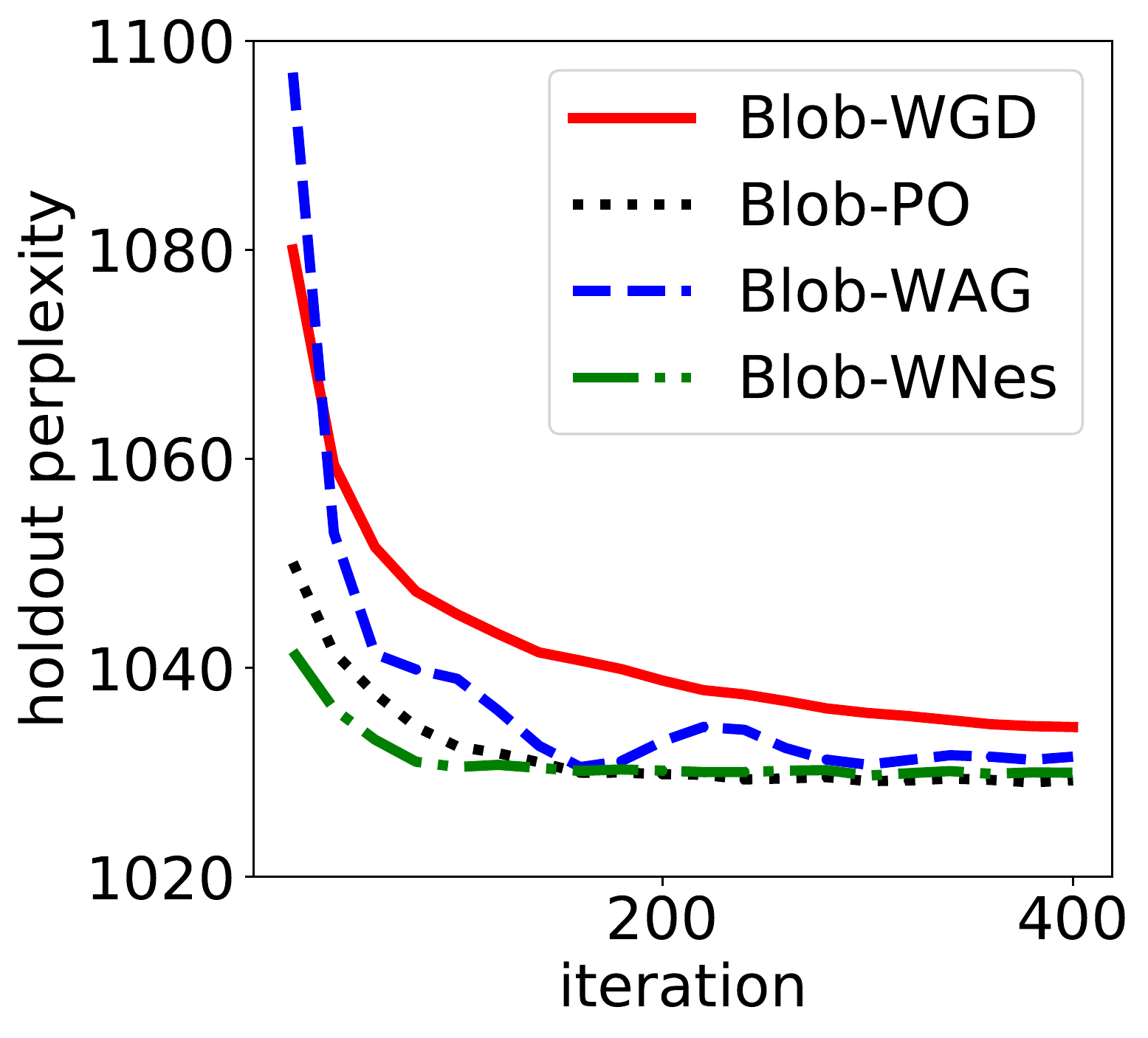}\vspace{-4pt}
    \label{fig:ldablob}
  }
  \\[-8pt]
  \hspace{-5pt}
  \subfigure[GFSD]{
    \includegraphics[scale=.23]{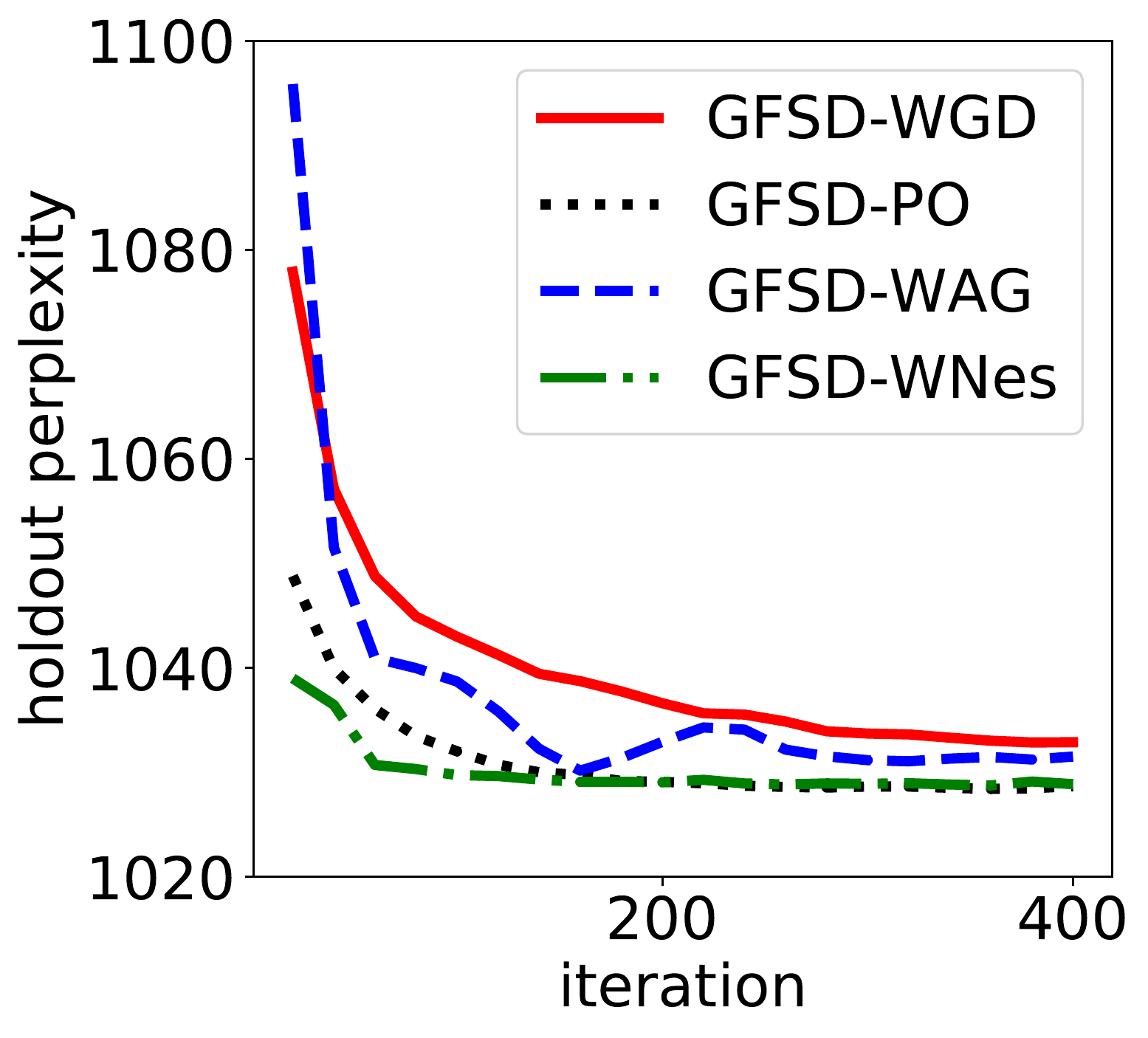}\vspace{-4pt}
    \label{fig:ldagfsd}
  }
  \hspace{-5pt}
  \subfigure[GFSF]{
    \includegraphics[scale=.23]{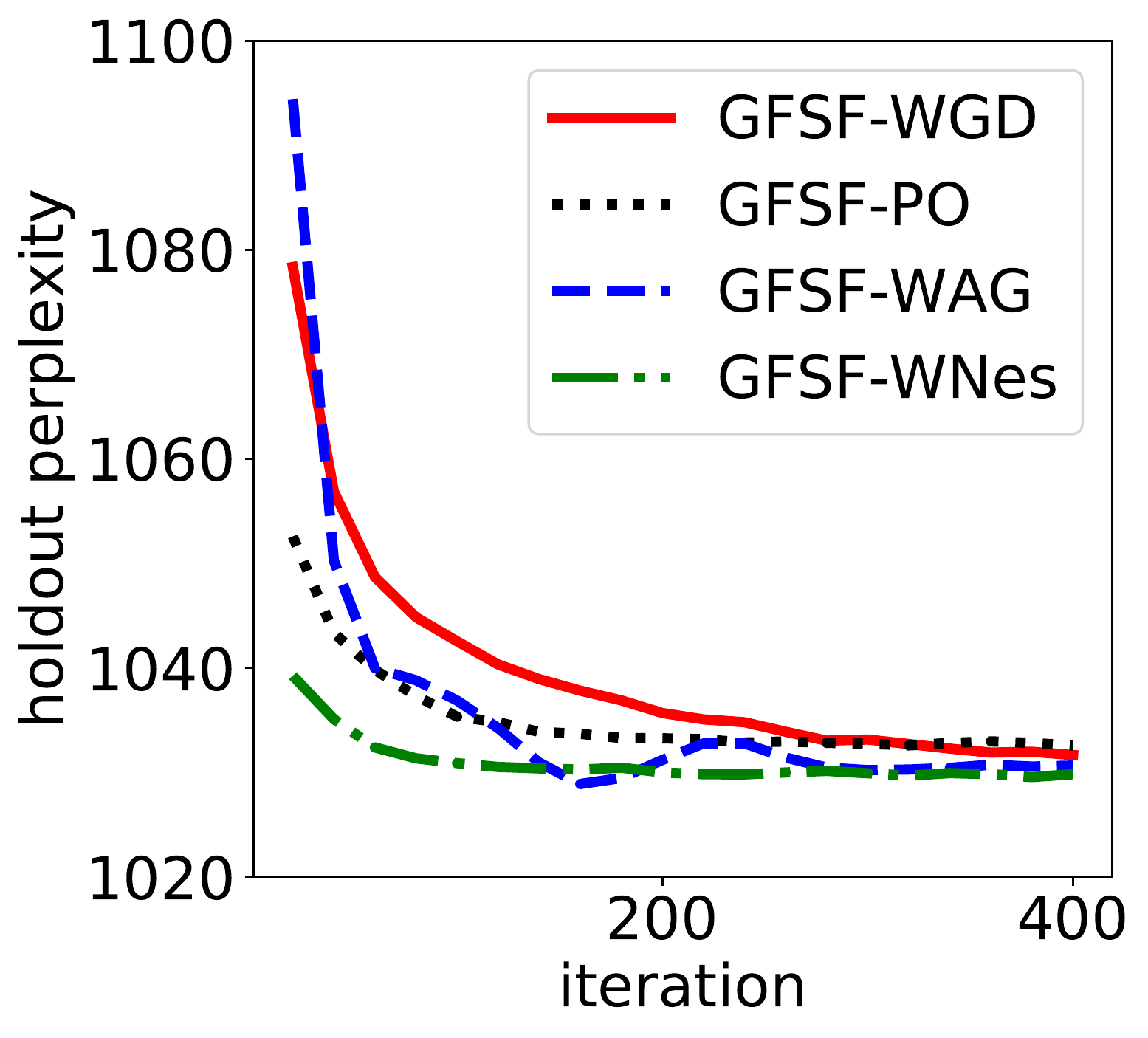}\vspace{-4pt}
    \label{fig:ldagfsf}
  }
  \vspace{-8pt}
  \caption{Acceleration effect of WAG and WNes on LDA. Curves are averaged over 10 runs.}
  \label{fig:lda}
  \vspace{-14pt}
\end{figure}

\begin{wrapfigure}{r}{.230\textwidth}
  \vspace{-14pt}
  \centering
  \includegraphics[width=.21\textwidth]{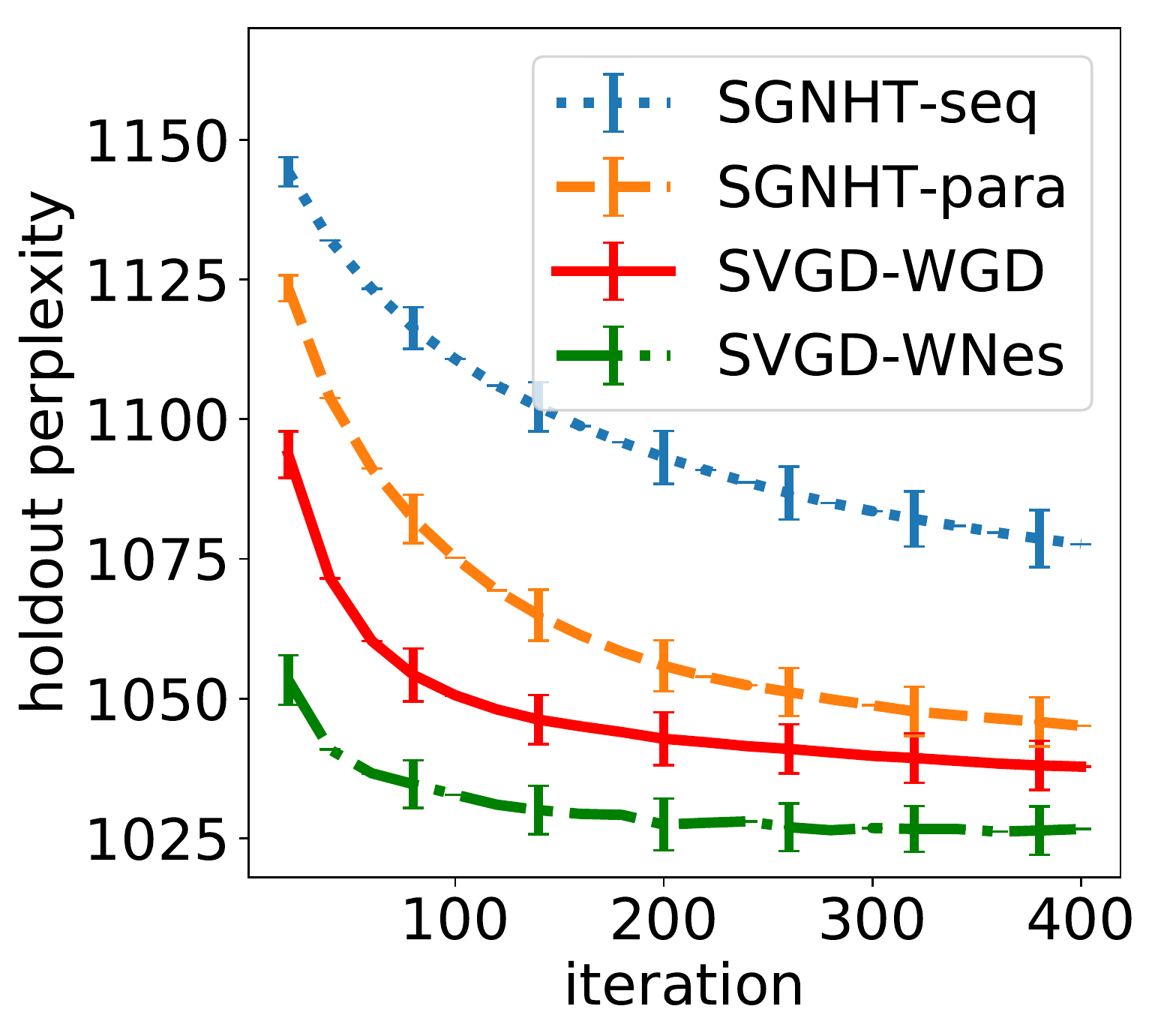}\vspace{-.4cm}
  \caption{Comparison of SVGD and SGNHT on LDA, as representatives of ParVIs and MCMCs. Average over 10 runs.}
  \label{fig:ldasgnht}
  \vspace{-5pt}
\end{wrapfigure}
We show the improved performance by the acceleration methods on an unsupervised task: posterior inference of LDA~\cite{blei2003latent}.
We follow the same settings as \citet{ding2014bayesian}, including the ICML dataset\footnote{\url{https://cse.buffalo.edu/~changyou/code/SGNHT.zip}} and the Expanded-Natural parameterization~\cite{patterson2013stochastic}. 
The particle size is fixed at 20. 
Inference results are evaluated by the conventional hold-out perplexity (the lower the better). 

The acceleration effect is shown in Fig.~\ref{fig:lda}.
We see again that WAG and WNes improve the convergence rate over WGD.
PO has a comparable empirical acceleration performance.
We note that WAG is sensitive to its parameter $\alpha$ and exhibits minor oscillation, while WNes is more stable.
We also compare ParVIs with stochastic gradient Nos\'{e}-Hoover thermostats (SGNHT)~\cite{ding2014bayesian}, an advanced MCMC method.
Its samples are taken as either the last 20 samples from one chain (-seq), or the very last sample from 20 parallel chains (-para).
From Figure~\ref{fig:ldasgnht}, we observe the faster convergence of ParVIs over the MCMC method.

\vspace{-9pt}
\section{Conclusions}
\vspace{-5pt}

By exploiting the $\clP_2(\clX)$ gradient flow perspective of ParVIs, we establish a unified theory on the finite-particle approximations of ParVIs, and propose an acceleration framework and a principled bandwidth-selection method to improve ParVIs.
The theory recognizes various approximations as a smoothing treatment, by either smoothing the density or smoothing functions.
The equivalence of the two smoothing forms connects existing ParVIs, and their necessity reveals the assumptions of ParVIs.
Algorithm acceleration is developed via a deep exploration on the geometry of $\clP_2(\clX)$ and the bandwidth method is based on a principle of smoothing.
Experiments show more representative particles by the principled bandwidth method, and the speed-up of ParVIs by the acceleration framework.


\section*{Acknowledgments}
This work was supported by the National Key Research and Development Program of China (No. 2017YFA0700904), NSFC Projects (Nos. 61620106010, 61621136008, 61571261), Beijing NSF Project (No. L172037), DITD Program JCKY2017204B064, Tiangong Institute for Intelligent Computing, Beijing Academy of Artificial Intelligence (BAAI), NVIDIA NVAIL Program, and the projects from Siemens and Intel.
This work was done when C. Liu was visiting Duke University, during which he was supported by the China Scholarship Council.

\bibliography{refs_thesis}

\begin{thebibliography}{59}
\providecommand{\natexlab}[1]{#1}
\providecommand{\url}[1]{\texttt{#1}}
\expandafter\ifx\csname urlstyle\endcsname\relax
  \providecommand{\doi}[1]{doi: #1}\else
  \providecommand{\doi}{doi: \begingroup \urlstyle{rm}\Url}\fi

\bibitem[Amari(2016)]{amari2016information}
Amari, S.-I.
\newblock \emph{Information geometry and its applications}.
\newblock Springer, 2016.

\bibitem[Ambrosio et~al.(2008)Ambrosio, Gigli, and
  Savar{\'e}]{ambrosio2008gradient}
Ambrosio, L., Gigli, N., and Savar{\'e}, G.
\newblock \emph{Gradient flows: in metric spaces and in the space of
  probability measures}.
\newblock Springer Science \& Business Media, 2008.

\bibitem[Benamou \& Brenier(2000)Benamou and Brenier]{benamou2000computational}
Benamou, J.-D. and Brenier, Y.
\newblock A computational fluid mechanics solution to the {M}onge-{K}antorovich
  mass transfer problem.
\newblock \emph{Numerische Mathematik}, 84\penalty0 (3):\penalty0 375--393,
  2000.

\bibitem[Blei et~al.(2003)Blei, Ng, and Jordan]{blei2003latent}
Blei, D.~M., Ng, A.~Y., and Jordan, M.~I.
\newblock Latent {D}irichlet allocation.
\newblock \emph{The Journal of Machine Learning Research}, 3:\penalty0
  993--1022, 2003.

\bibitem[Chen \& Zhang(2017)Chen and Zhang]{chen2017particle}
Chen, C. and Zhang, R.
\newblock Particle optimization in stochastic gradient {MCMC}.
\newblock \emph{arXiv preprint arXiv:1711.10927}, 2017.

\bibitem[Chen et~al.(2015)Chen, Ding, and Carin]{chen2015convergence}
Chen, C., Ding, N., and Carin, L.
\newblock On the convergence of stochastic gradient {MCMC} algorithms with
  high-order integrators.
\newblock In \emph{Advances in Neural Information Processing Systems}, pp.\
  2269--2277, Montréal, Canada, 2015. NIPS Foundation.

\bibitem[Chen et~al.(2018{\natexlab{a}})Chen, Zhang, Wang, Li, and
  Chen]{chen2018unified}
Chen, C., Zhang, R., Wang, W., Li, B., and Chen, L.
\newblock A unified particle-optimization framework for scalable {B}ayesian
  sampling.
\newblock In \emph{Proceedings of the Conference on Uncertainty in Artificial
  Intelligence (UAI 2018)}, Monterey, California USA, 2018{\natexlab{a}}.
  Association for Uncertainty in Artificial Intelligence.

\bibitem[Chen et~al.(2018{\natexlab{b}})Chen, Mackey, Gorham, Briol, and
  Oates]{chen2018stein}
Chen, W.~Y., Mackey, L., Gorham, J., Briol, F.-X., and Oates, C.~J.
\newblock {S}tein points.
\newblock \emph{arXiv preprint arXiv:1803.10161}, 2018{\natexlab{b}}.

\bibitem[Chen \& Li(2018)Chen and Li]{chen2018natural}
Chen, Y. and Li, W.
\newblock Natural gradient in {W}asserstein statistical manifold.
\newblock \emph{arXiv preprint arXiv:1805.08380}, 2018.

\bibitem[Cuturi(2013)]{cuturi2013sinkhorn}
Cuturi, M.
\newblock Sinkhorn distances: Lightspeed computation of optimal transport.
\newblock In \emph{Advances in Neural Information Processing Systems}, pp.\
  2292--2300, Lake Tahoe, Nevada USA, 2013. NIPS Foundation.

\bibitem[Detommaso et~al.(2018)Detommaso, Cui, Marzouk, Spantini, and
  Scheichl]{detommaso2018stein}
Detommaso, G., Cui, T., Marzouk, Y., Spantini, A., and Scheichl, R.
\newblock A {S}tein variational {N}ewton method.
\newblock In \emph{Advances in Neural Information Processing Systems}, pp.\
  9187--9197, Montréal, Canada, 2018. NIPS Foundation.

\bibitem[Ding et~al.(2014)Ding, Fang, Babbush, Chen, Skeel, and
  Neven]{ding2014bayesian}
Ding, N., Fang, Y., Babbush, R., Chen, C., Skeel, R.~D., and Neven, H.
\newblock {B}ayesian sampling using stochastic gradient thermostats.
\newblock In \emph{Advances in Neural Information Processing Systems}, pp.\
  3203--3211, Montréal, Canada, 2014. NIPS Foundation.

\bibitem[Do~Carmo(1992)]{do1992riemannian}
Do~Carmo, M.~P.
\newblock \emph{{R}iemannian Geometry}.
\newblock Birkh{\"a}user, 1992.

\bibitem[Dua \& Graff(2017)Dua and Graff]{asuncion2007uci}
Dua, D. and Graff, C.
\newblock {UCI} machine learning repository, 2017.
\newblock URL \url{http://archive.ics.uci.edu/ml}.

\bibitem[Ehlers et~al.(1972)Ehlers, Pirani, and Schild]{ehlers1972geometry}
Ehlers, J., Pirani, F., and Schild, A.
\newblock The geometry of free fall and light propagation, in the book
  “{G}eneral {R}elativity” (papers in honour of {JL} {S}ynge), 63--84,
  1972.

\bibitem[Erbar et~al.(2010)]{erbar2010heat}
Erbar, M. et~al.
\newblock The heat equation on manifolds as a gradient flow in the
  {W}asserstein space.
\newblock In \emph{Annales de l'Institut Henri {P}oincar{\'e}, Probabilit{\'e}s
  et Statistiques}, volume~46, pp.\  1--23. Institut Henri Poincar{\'e}, 2010.

\bibitem[Feng et~al.(2017)Feng, Wang, and Liu]{feng2017learning}
Feng, Y., Wang, D., and Liu, Q.
\newblock Learning to draw samples with amortized {S}tein variational gradient
  descent.
\newblock In \emph{Proceedings of the Conference on Uncertainty in Artificial
  Intelligence (UAI 2017)}, Sydney, Australia, 2017. Association for
  Uncertainty in Artificial Intelligence.

\bibitem[Futami et~al.(2018)Futami, Cui, Sato, and Sugiyama]{futami2018frank}
Futami, F., Cui, Z., Sato, I., and Sugiyama, M.
\newblock {F}rank-{W}olfe {S}tein sampling.
\newblock \emph{arXiv preprint arXiv:1805.07912}, 2018.

\bibitem[Gabay(1982)]{gabay1982minimizing}
Gabay, D.
\newblock Minimizing a differentiable function over a differential manifold.
\newblock \emph{Journal of Optimization Theory and Applications}, 37\penalty0
  (2):\penalty0 177--219, 1982.

\bibitem[Geman \& Geman(1987)Geman and Geman]{geman1987stochastic}
Geman, S. and Geman, D.
\newblock Stochastic relaxation, {G}ibbs distributions, and the {B}ayesian
  restoration of images.
\newblock In \emph{Readings in Computer Vision}, pp.\  564--584. Elsevier,
  1987.

\bibitem[Haarnoja et~al.(2017)Haarnoja, Tang, Abbeel, and
  Levine]{haarnoja2017reinforcement}
Haarnoja, T., Tang, H., Abbeel, P., and Levine, S.
\newblock Reinforcement learning with deep energy-based policies.
\newblock In \emph{Proceedings of the 34th International Conference on Machine
  Learning (ICML 2017)}, pp.\  1352--1361, Sydney, Australia, 2017. IMLS.

\bibitem[Hoffman et~al.(2013)Hoffman, Blei, Wang, and
  Paisley]{hoffman2013stochastic}
Hoffman, M.~D., Blei, D.~M., Wang, C., and Paisley, J.
\newblock Stochastic variational inference.
\newblock \emph{The Journal of Machine Learning Research}, 14\penalty0
  (1):\penalty0 1303--1347, 2013.

\bibitem[Jordan et~al.(1998)Jordan, Kinderlehrer, and
  Otto]{jordan1998variational}
Jordan, R., Kinderlehrer, D., and Otto, F.
\newblock The variational formulation of the {F}okker-{P}lanck equation.
\newblock \emph{SIAM journal on mathematical analysis}, 29\penalty0
  (1):\penalty0 1--17, 1998.

\bibitem[Kheyfets et~al.(2000)Kheyfets, Miller, and Newton]{kheyfets2000schild}
Kheyfets, A., Miller, W.~A., and Newton, G.~A.
\newblock Schild's ladder parallel transport procedure for an arbitrary
  connection.
\newblock \emph{International Journal of Theoretical Physics}, 39\penalty0
  (12):\penalty0 2891--2898, 2000.

\bibitem[Kov{\'a}{\v{c}}ik \& R{\'a}kosn{\'\i}k(1991)Kov{\'a}{\v{c}}ik and
  R{\'a}kosn{\'\i}k]{kovavcik1991spaces}
Kov{\'a}{\v{c}}ik, O. and R{\'a}kosn{\'\i}k, J.
\newblock On spaces {$L^p(x)$} and {$W^{k, p}(x)$}.
\newblock \emph{Czechoslovak Mathematical Journal}, 41\penalty0 (4):\penalty0
  592--618, 1991.

\bibitem[Li \& Turner(2018)Li and Turner]{li2018gradient}
Li, Y. and Turner, R.~E.
\newblock Gradient estimators for implicit models.
\newblock In \emph{Proceedings of the International Conference on Learning
  Representations (ICLR 2018)}, Vancouver, Canada, 2018. ICLR Committee.
\newblock URL \url{https://openreview.net/forum?id=SJi9WOeRb}.

\bibitem[Liu \& Zhu(2018)Liu and Zhu]{liu2018riemannian}
Liu, C. and Zhu, J.
\newblock {R}iemannian {S}tein variational gradient descent for {B}ayesian
  inference.
\newblock In \emph{The 32nd AAAI Conference on Artificial Intelligence}, pp.\
  3627--3634, New Orleans, Louisiana USA, 2018. AAAI press.
\newblock URL
  \url{https://aaai.org/ocs/index.php/AAAI/AAAI18/paper/view/17275}.

\bibitem[Liu(2017)]{liu2017steinflow}
Liu, Q.
\newblock {S}tein variational gradient descent as gradient flow.
\newblock In \emph{Advances in Neural Information Processing Systems}, pp.\
  3118--3126, Long Beach, California USA, 2017. NIPS Foundation.

\bibitem[Liu \& Wang(2016)Liu and Wang]{liu2016stein}
Liu, Q. and Wang, D.
\newblock {S}tein variational gradient descent: A general purpose {B}ayesian
  inference algorithm.
\newblock In \emph{Advances in Neural Information Processing Systems}, pp.\
  2370--2378, Barcelona, Spain, 2016. NIPS Foundation.

\bibitem[Liu et~al.(2017{\natexlab{a}})Liu, Ramachandran, Liu, and
  Peng]{liu2017steinpolicy}
Liu, Y., Ramachandran, P., Liu, Q., and Peng, J.
\newblock {S}tein variational policy gradient.
\newblock In \emph{Proceedings of the Conference on Uncertainty in Artificial
  Intelligence (UAI 2017)}, Sydney, Australia, 2017{\natexlab{a}}. Association
  for Uncertainty in Artificial Intelligence.

\bibitem[Liu et~al.(2017{\natexlab{b}})Liu, Shang, Cheng, Cheng, and
  Jiao]{liu2017accelerated}
Liu, Y., Shang, F., Cheng, J., Cheng, H., and Jiao, L.
\newblock Accelerated first-order methods for geodesically convex optimization
  on {R}iemannian manifolds.
\newblock In \emph{Advances in Neural Information Processing Systems}, pp.\
  4875--4884, Long Beach, California USA, 2017{\natexlab{b}}. NIPS Foundation.

\bibitem[Lott(2008)]{lott2008some}
Lott, J.
\newblock Some geometric calculations on {W}asserstein space.
\newblock \emph{Communications in Mathematical Physics}, 277\penalty0
  (2):\penalty0 423--437, 2008.

\bibitem[Lott(2017)]{lott2017intrinsic}
Lott, J.
\newblock An intrinsic parallel transport in {W}asserstein space.
\newblock \emph{Proceedings of the American Mathematical Society}, 145\penalty0
  (12):\penalty0 5329--5340, 2017.

\bibitem[Neal(2011)]{neal2011mcmc}
Neal, R.~M.
\newblock {MCMC} using {H}amiltonian dynamics.
\newblock \emph{Handbook of {M}arkov Chain {M}onte {C}arlo}, 2, 2011.

\bibitem[Nesterov(1983)]{nesterov1983method}
Nesterov, Y.
\newblock A method of solving a convex programming problem with convergence
  rate {$O(1/k^2)$}.
\newblock In \emph{Soviet Mathematics Doklady}, volume~27, pp.\  372--376,
  Moscow, 1983. Russian Academy of Sciences.

\bibitem[Otto(2001)]{otto2001geometry}
Otto, F.
\newblock The geometry of dissipative evolution equations: the porous medium
  equation.
\newblock 2001.

\bibitem[Patterson \& Teh(2013)Patterson and Teh]{patterson2013stochastic}
Patterson, S. and Teh, Y.~W.
\newblock Stochastic gradient {R}iemannian {L}angevin dynamics on the
  probability simplex.
\newblock In \emph{Advances in Neural Information Processing Systems}, pp.\
  3102--3110, Lake Tahoe, Nevada USA, 2013. NIPS Foundation.

\bibitem[Pele \& Werman(2009)Pele and Werman]{pele2009fast}
Pele, O. and Werman, M.
\newblock Fast and robust earth mover's distances.
\newblock In \emph{Proceedings of the 12th International Conference on Computer
  Vision (ICCV-09)}, volume~9, pp.\  460--467, Kyoto, Japan, 2009. IEEE.

\bibitem[Polyak(1964)]{polyak1964some}
Polyak, B.~T.
\newblock Some methods of speeding up the convergence of iteration methods.
\newblock \emph{USSR Computational Mathematics and Mathematical Physics},
  4\penalty0 (5):\penalty0 1--17, 1964.

\bibitem[Pu et~al.(2017)Pu, Gan, Henao, Li, Han, and Carin]{pu2017vae}
Pu, Y., Gan, Z., Henao, R., Li, C., Han, S., and Carin, L.
\newblock {VAE} learning via {S}tein variational gradient descent.
\newblock In \emph{Advances in Neural Information Processing Systems}, pp.\
  4239--4248, Long Beach, California USA, 2017. NIPS Foundation.

\bibitem[Qi et~al.(2010)Qi, Gallivan, and Absil]{qi2010riemannian}
Qi, C., Gallivan, K.~A., and Absil, P.-A.
\newblock {R}iemannian {BFGS} algorithm with applications.
\newblock In \emph{Recent advances in optimization and its applications in
  engineering}, pp.\  183--192. Springer, 2010.

\bibitem[Rezende \& Mohamed(2015)Rezende and Mohamed]{rezende2015variational}
Rezende, D. and Mohamed, S.
\newblock Variational inference with normalizing flows.
\newblock In \emph{Proceedings of The 32nd International Conference on Machine
  Learning (ICML 2015)}, pp.\  1530--1538, Lille, France, 2015. IMLS.

\bibitem[Roberts \& Stramer(2002)Roberts and Stramer]{roberts2002langevin}
Roberts, G.~O. and Stramer, O.
\newblock {L}angevin diffusions and {M}etropolis-{H}astings algorithms.
\newblock \emph{Methodology and computing in applied probability}, 4\penalty0
  (4):\penalty0 337--357, 2002.

\bibitem[Shi et~al.(2018)Shi, Sun, and Zhu]{shi2018spectral}
Shi, J., Sun, S., and Zhu, J.
\newblock A spectral approach to gradient estimation for implicit
  distributions.
\newblock In \emph{Proceedings of the 35th International Conference on Machine
  Learning (ICML 2018)}, pp.\  4651--4660, Stockholm, Sweden, 2018. IMLS.

\bibitem[Steinwart \& Christmann(2008)Steinwart and
  Christmann]{steinwart2008support}
Steinwart, I. and Christmann, A.
\newblock \emph{Support vector machines}.
\newblock Springer Science \& Business Media, New York, 2008.

\bibitem[Sutskever et~al.(2013)Sutskever, Martens, Dahl, and
  Hinton]{sutskever2013importance}
Sutskever, I., Martens, J., Dahl, G., and Hinton, G.
\newblock On the importance of initialization and momentum in deep learning.
\newblock In \emph{Proceedings of the 30th International Conference on Machine
  Learning (ICML 2013)}, pp.\  1139--1147, Atlanta, Georgia USA, 2013. IMLS.

\bibitem[Taghvaei \& Mehta(2019)Taghvaei and Mehta]{taghvaei2018accelerated}
Taghvaei, A. and Mehta, P.~G.
\newblock Accelerated gradient flow for probability distributions.
\newblock In \emph{Proceedings of the 36th International Conference on Machine
  Learning (ICML 2019)}, Long Beach, California USA, 2019. IMLS.

\bibitem[Villani(2008)]{villani2008optimal}
Villani, C.
\newblock \emph{Optimal transport: old and new}, volume 338.
\newblock Springer Science \& Business Media, 2008.

\bibitem[Wainwright et~al.(2008)Wainwright, Jordan,
  et~al.]{wainwright2008graphical}
Wainwright, M.~J., Jordan, M.~I., et~al.
\newblock Graphical models, exponential families, and variational inference.
\newblock \emph{Foundations and Trends{\textregistered} in Machine Learning},
  1\penalty0 (1--2):\penalty0 1--305, 2008.

\bibitem[Welling \& Teh(2011)Welling and Teh]{welling2011bayesian}
Welling, M. and Teh, Y.~W.
\newblock {B}ayesian learning via stochastic gradient {L}angevin dynamics.
\newblock In \emph{Proceedings of the 28th International Conference on Machine
  Learning (ICML 2011)}, pp.\  681--688, Bellevue, Washington USA, 2011. IMLS.

\bibitem[Wibisono et~al.(2016)Wibisono, Wilson, and
  Jordan]{wibisono2016variational}
Wibisono, A., Wilson, A.~C., and Jordan, M.~I.
\newblock A variational perspective on accelerated methods in optimization.
\newblock \emph{Proceedings of the National Academy of Sciences}, 113\penalty0
  (47):\penalty0 E7351--E7358, 2016.

\bibitem[Xie et~al.(2018)Xie, Wang, Wang, and Zha]{xie2018fast}
Xie, Y., Wang, X., Wang, R., and Zha, H.
\newblock A fast proximal point method for computing {W}asserstein distance.
\newblock \emph{arXiv preprint arXiv:1802.04307}, 2018.

\bibitem[Yuan et~al.(2016)Yuan, Huang, Absil, and Gallivan]{yuan2016riemannian}
Yuan, X., Huang, W., Absil, P.-A., and Gallivan, K.~A.
\newblock A {R}iemannian limited-memory {BFGS} algorithm for computing the
  matrix geometric mean.
\newblock \emph{Procedia Computer Science}, 80:\penalty0 2147--2157, 2016.

\bibitem[Zhang \& Sra(2018)Zhang and Sra]{zhang2018estimate}
Zhang, H. and Sra, S.
\newblock An estimate sequence for geodesically convex optimization.
\newblock In \emph{Proceedings of the 31st Annual Conference on Learning Theory
  (COLT 2018)}, pp.\  1703--1723, Stockholm, Sweden, 2018. IMLS.

\bibitem[Zhang et~al.(2016)Zhang, Reddi, and Sra]{zhang2016riemannian}
Zhang, H., Reddi, S.~J., and Sra, S.
\newblock {R}iemannian {SVRG}: Fast stochastic optimization on {R}iemannian
  manifolds.
\newblock In \emph{Advances in Neural Information Processing Systems}, pp.\
  4592--4600, Barcelona, Spain, 2016. NIPS Foundation.

\bibitem[Zhang et~al.(2018)Zhang, Chen, Li, and Duke]{zhang2018policy}
Zhang, R., Chen, C., Li, C., and Duke, L.~C.
\newblock Policy optimization as {W}asserstein gradient flows.
\newblock In \emph{Proceedings of the 35th International Conference on Machine
  Learning (ICML 2018)}, pp.\  5741--5750, Stockholm, Sweden, 2018. IMLS.

\bibitem[Zhang et~al.(2019)Zhang, Wen, Chen, and Carin]{zhang2019scalable}
Zhang, R., Wen, Z., Chen, C., and Carin, L.
\newblock Scalable {T}hompson sampling via optimal transport.
\newblock In \emph{Proceedings of the 22nd International Conference on
  Artificial Intelligence and Statistics (AISTATS-19)}, Naha, Okinawa Japan,
  2019. AISTATS Committee.

\bibitem[Zhou(2008)]{zhou2008derivative}
Zhou, D.-X.
\newblock Derivative reproducing properties for kernel methods in learning
  theory.
\newblock \emph{Journal of computational and Applied Mathematics}, 220\penalty0
  (1):\penalty0 456--463, 2008.

\bibitem[Zhuo et~al.(2018)Zhuo, Liu, Shi, Zhu, Chen, and
  Zhang]{zhuo2018message}
Zhuo, J., Liu, C., Shi, J., Zhu, J., Chen, N., and Zhang, B.
\newblock Message passing {S}tein variational gradient descent.
\newblock In Dy, J. and Krause, A. (eds.), \emph{Proceedings of the 35th
  International Conference on Machine Learning}, volume~80 of \emph{Proceedings
  of Machine Learning Research}, pp.\  6018--6027, Stockholmsmässan, Stockholm
  Sweden, 10--15 Jul 2018. IMLS, PMLR.
\newblock URL \url{http://proceedings.mlr.press/v80/zhuo18a.html}.

\end{thebibliography}
\bibliographystyle{icml2019}

\newpage

\section*{Appendix}

\subsection*{A. Proofs}
\subsubsection*{A.1: Proof of Theorem~\ref{thm:svgd}}
For $v\in\clH^D$, the objective can be expressed as:
\begin{align}
	& \lrangle{ \vgf, v }_{\clL^2_q} \\
  = & \bbE_q[ (\nabla\log p - \nabla\log q) \cdot v ] \\
  = & \bbE_q[ \nabla\log p \cdot v] - \int_{\clX} \nabla q \cdot v \dd x \\
  \stackrel{(*)}{=}
    & \bbE_q[ \nabla\log p \cdot v] + \int_{\clX} q \nabla \cdot v \dd x \\
  = & \bbE_{q(x)}\bigg[ \sum_{\alpha=1}^D \Big( \partial_{\alpha}\log p(x) v_{\alpha}(x) + \partial_{\alpha} v_{\alpha}(x) \Big) \bigg] \\
  \stackrel{(\#)}{=}
    & \bbE_{q(x)}\bigg[ \sum_{\alpha=1}^D \Big( \partial_{\alpha}\log p(x) \lrangle{ K(x,\cdot), v_{\alpha}(\cdot) }_{\clH} \\
    & {} + \lrangle{ \partial_{\alpha} K(x,\cdot), v_{\alpha}(\cdot) }_{\clH} \Big) \bigg] \\
  = & \bbE_{q(x)}\left[ \lrangle{ K(x,\cdot) \nabla\log p(x), v(\cdot) }_{\clH^D} + \lrangle{ \nabla K(x,\cdot), v(\cdot) }_{\clH^D} \right] \\
  = & \bbE_{q(x)}\left[ \lrangle{ K(x,\cdot) \nabla\log p(x) + \nabla K(x,\cdot), v(\cdot) }_{\clH^D} \right] \\
  = & \lrangle{ \bbE_{q(x)}\left[ K(x,\cdot) \nabla\log p(x) + \nabla K(x,\cdot) \right], v(\cdot) }_{\clH^D} \\
  = & \lrangle{ \vsvgd, v }_{\clH^D},
\end{align}
where the equality $(*)$ holds due to the definition of weak derivative of distributions, and equality $(\#)$ holds due to the reproducing property for any function $f$ in the reproducing kernel Hilbert space $\clH$ of kernel $K$: $\lrangle{ K(x,\cdot), f(\cdot) }_{\clH} = f(x)$~(\citet{steinwart2008support}, Chapter~4), and $\lrangle{ \partial_{x_{\alpha}}K(x,\cdot), f(\cdot) }_{\clH} = \partial_{x_{\alpha}} f(x)$~\cite{zhou2008derivative}.

\subsubsection*{A.2: Proof of Theorem~\ref{thm:smfun}}
When $q$ is absolutely continuous with respect to the Lebesgue measure of $\clX = \bbR^D$, $\clL^2_q$ has the same topological properties as $\clL^2$, so conclusions we cite below can be adapted from $\clL^2$ to $\clL^2_q$.
Note that the map $\phi \mapsto \phi*K, \clL^2 \to \clL^2$ is continuous, so $\clG := \overline{\{\phi*K: \phi\in\clCC\}}^{\clL^2_q} = \{\phi*K: \phi \in \overline{\clCC}^{\clL^2} \} = \{\phi*K: \phi \in \clL^2 \} = \{\phi*K: \phi \in L^2 \}^D$, where the second last equality holds due to \eg, Theorem~2.11 of \cite{kovavcik1991spaces}.
On the other hand, due to Proposition~4.46 and Theorem~4.47 of \cite{steinwart2008support}, the map $\phi \mapsto \phi*K$ is an isometric isomorphism between $\{\phi*K: \phi \in L^2 \}$ and $\clH$, the reproducing kernel Hilbert space of $K$.
This indicates that $\clG$ is isometrically isomorphic to $\clH^D$.

\subsubsection*{A.3: Proof of Theorem~\ref{thm:rawsvgd}}

We will redefine some notations in this proof.
According to the deduction in Appendix~A.1, the objective of the optimization problem \eqref{eqn:svgdopt} $\lrangle{ \vgf, v }_{\clL^2_q}$ can be cast as $\bbE_q[\nabla\log p \cdot v + \nabla \cdot v]$.
With $q = \hqq$ and $v \in \clL^2_p$, we write the optimization problem as:
\begin{align}
  \sup_{v \in \clL^2_p, \lrVert{v}=1} \sum_{i=1}^N \Big( \nabla \log p(x^{(i)}) \cdot v (x^{(i)}) + \nabla \cdot v(x^{(i)}) \Big),
  \label{eqn:appxobj}
\end{align}
We will find a sequence of functions $\{v_n\}$ satisfying conditions in \eqref{eqn:appxobj} while the objective goes to infinity.

We assume that there exists $r_0>0$ such that $p(x)>0$ for any $\lrVert{ x-x^{(i)} }_{\infty} < r_0$, $i = 1,2,\cdots,N$, which is reasonable because it is almost impossible to sample $x^{(i)}$ with $p(x)$ vanishes in every neighborhood of $x^{(i)}$.

Denoting $v(x)=(v_1(x), \cdots ,v_D(x))\trs$ for any $D$-dimensional vector function $v$ and $\nabla f(x) =(\partial_1 f(x), \cdots ,\partial_D f(x))\trs$ for any real-valued function $f$, the objective can be written as:
\begin{align}
  \clL_{v} =& \sum_{i=1}^N \Big(\nabla \log p(x^{(i)}) \cdot v (x^{(i)}) + \nabla \cdot v(x^{(i)}) \Big) \\
  =& \sum_{i=1}^N \Big( \sum_{\alpha=1}^D \partial_{\alpha}[ \log p(x^{(i)}) ]v_{\alpha}(x^{(i)}) + \sum_{\alpha=1}^D \partial_{\alpha}[ v_{\alpha}(x^{(i)}) ] \Big) \\
  =& \sum_{\alpha=1}^D \sum_{i=1}^N \Big(\partial_{\alpha}[ \log p(x^{(i)}) ]v_{\alpha}(x^{(i)}) + \partial_{\alpha}[ v_{\alpha}(x^{(i)}) ] \Big).
  \label{eqn:appxred}
\end{align}
For every $v \in \clL^2_p, \lrVert{v}=1$, we can define a function $\phi=(\phi_1,\cdots,\phi_D)\trs \in \clL^2$ correspondingly, such that 
$\phi(x) = p(x)^{\frac{1}{2}}v(x)$, which means $\phi_\alpha (x) = p(x)^{\frac{1}{2}} v_{\alpha}(x)$, and
\begin{align}
  \lrVert{\phi}_2^2 =& \int_{\bbR^D} \phi^2 \dd x  = \int_{\bbR^D} \sum_{\alpha=1}^D (\phi_{\alpha}(x))^2 \dd x \\
  =& \int_{\bbR^D} \sum_{\alpha=1}^D (v_{\alpha}(x))^2 p(x) \dd x = \lrVert{ v }^2 = 1.
\end{align}
Rewrite \eqref{eqn:appxred} in term of $\phi$, we have:
\begin{align}
  \label{eqn:appxred2}
  \hspace{-12pt}
  \clL_\phi =& \sum_{\alpha=1}^D \sum_{i=1}^N \Big(\partial_{\alpha}[ \log p(x^{(i)}) ] v_{\alpha}(x^{(i)}) + \partial_{\alpha}[ v_{\alpha}(x^{(i)}) ] \Big) \\
  =& \sum_{\alpha=1}^D \sum_{i=1}^N \Big(\partial_{\alpha}[ \log p(x^{(i)}) ]\phi_{\alpha}(x^{(i)}){p(x^{(i)})}^{-\frac{1}{2}} \\ 
   & {} + \partial_{\alpha}[ \phi_{\alpha}(x^{(i)}){p(x^{(i)})}^{-\frac{1}{2}} ] \Big) \displaybreak \\
  =& \sum_{\alpha=1}^D \sum_{i=1}^N \Big(\frac{1}{2} p(x^{(i)})^{-\frac{3}{2}}\partial_{\alpha}[p(x^{(i)})] \phi_{\alpha}(x^{(i)}) \\ 
   & {} +p(x^{(i)})^{-\frac{1}{2}} \partial_{\alpha}[\phi_{\alpha}(x^{(i)})] \Big) \\
  = & \sum_{\alpha=1}^D \sum_{i=1}^N \Big( A_{\alpha}^{(i)}\phi_{\alpha}(x^{(i)}) + B^{(i)} \partial_{\alpha}[ \phi_{\alpha}(x^{(i)}) ] \Big),
\end{align}
where $A_{\alpha}^{(i)} = \frac{1}{2} p(x^{(i)})^{-\frac{3}{2}}\partial_{\alpha}[p(x^{(i)})] $ and $B^{(i)} = p(x^{(i)})^{-\frac{1}{2}} > 0$. We will now construct a sequence $\{\phi_n\}$ to show the following problem:
\begin{align}
  \label{eqn:appxred_obj}
  \inf_{\phi \in \clL^2, \lrVert{\phi} = 1}\sum_{\alpha=1}^D \sum_{i=1}^N \Big( A_{\alpha}^{(i)}\phi_{\alpha}(x^{(i)}) + B^{(i)} \partial_{\alpha}[ \phi_{\alpha}(x^{(i)}) ] \Big)
\end{align}
has no solution, then induce a sequence $\{ v_n \}$ by $\{ \phi_n \}$ for problem \eqref{eqn:appxobj}.

Define a sequence of functions
\begin{align}
  \chi_n(x) = \begin{dcases}
	{I_n^{-1/2}}{(1-x^2)^{n/2}}, & \text{for } x \in [-1,1], \\ 
	0, & \text{otherwise.} 
  \end{dcases}
\end{align}
We have $\int_{\bbR} \chi_n(x)^2 \dd x = 1$ with $I_n = \int_{-1}^{1} (1-x^2)^n \dd x = \sqrt{\pi} \frac{\Gamma(n+1)}{\Gamma(n+3/2)}$, where $\Gamma(\cdot)$ is the Gamma function.
Note that when $x = -1/\sqrt{n}$,
\begin{align}
  \chi_n'(x) =& {-n}{I_n^{-\frac{1}{2}}}x(1-x^2)^{\frac{n-2}{2}} \\
  =& \pi^{-\frac{1}{4}} \sqrt{\frac{\Gamma(n+\frac{3}{2})}{\Gamma(n+1)}} \sqrt{n} (1-\frac{1}{n})^{\frac{n-2}{2}} \;\;\;\;\;\; \text{($x=-\frac{1}{\sqrt{n}}$)} \\
  >& \pi^{-\frac{1}{4}} \sqrt{n}(1-\frac{1}{n})^{\frac{n-2}{2}}, \;\;\;\;\;\; \text{(${\Gamma(n+\frac{3}{2})}>{\Gamma(n+1)}$)}
  \label{eqn:appxd0}
\end{align}
therefore,
\begin{align}
  & \lim_{n \to \infty} \chi'_n(-\frac{1}{\sqrt{n}}) \\
  >& \lim_{n \to \infty} \pi^{-\frac{1}{4}} \sqrt{n}(1-\frac{1}{n})^{\frac{n-2}{2}} = \pi^{-\frac{1}{4}} e^{-\frac{1}{2}} \lim_{n \to \infty} \sqrt{n} = +\infty.
\end{align}

Denote $x^{(i)} = (x^{(i)}_1, x^{(i)}_2, \cdots , x^{(i)}_D)\trs \in \bbR^D, i=1,\cdots, N$ and
\begin{align}
  r_1 = \frac{1}{3} \min_{i \neq j} \lrVert{ x^{(i)} -x^{(j)} }_\infty =\frac{1}{3} \min_{\substack{\alpha \in \{1,\cdots,D \}, i\neq j}} \lrvert{ x_{\alpha}^{(i)} - x_{\alpha}^{(j)} }.
\end{align}
We extend $\chi_n$ to $\bbR^D$ as $\xi_n$ with support $\supp(\xi_n) = [-r,r]^D$,
\begin{align}
  \xi_n(x_1,x_2,\cdots ,x_D) = r^{-D/2} \prod_{\alpha=1}^D \chi_n(\frac{x_i}{r}),
\end{align}
where $r=\min\{r_0,r_1\}$.
It is easy to show that 
$\int_{\bbR^D} \xi_n(x)^2 \dd x = 1$, and 
\begin{align}
  \lim_{n \to \infty} \partial_{\alpha} \xi_n(-\epsilon_n) = +\infty, \; \alpha = 1,2,\cdots,D,
\end{align}
with $\epsilon_n = \frac{r}{\sqrt{n}} (1,1,\cdots,1)\trs$.

We choose $\phi_{\alpha}(x)=\frac{1}{ND} \sum_{i=1}^N \psi_{\alpha}^{(i)}$, where $\psi_{\alpha}^{(i)}$ is defined by:
\begin{align}
  \psi_{\alpha}^{(i)}(x) = \begin{dcases}
	\xi_n(x - x^{(i)} - \epsilon_n), & \text{if } A_{\alpha}^{(i)} >= 0, \\
	-\xi_n(x - x^{(i)} + \epsilon_n), & \text{if } A_{\alpha}^{(i)} < 0.
  \end{dcases}
\end{align}
With $\int_{\bbR^D} \psi^{(i)}_{\alpha}(x)  \psi^{(j)}_{\alpha}(x) \dd x = 0, \forall i\neq j$, we know $\phi_n$ satisfies conditions in \eqref{eqn:appxred_obj}.
Noting that $\forall i,j$, $A_{\alpha}^{(i)} \psi_{\alpha}^{(j)}(x^{(i)}) \geq 0$, and 
\begin{align}
  \partial_{\alpha} \psi_{\alpha}^{(j)}(x^{(i)}) = \begin{dcases}
	+\infty, & \text{when } n \to \infty, \text{if } i=j, \\
	0, & \text{if } i \neq j,
  \end{dcases}
\end{align}
we see $\clL_{\phi_n} \to +\infty$ in \eqref{eqn:appxred2} when $n \to \infty$ .

Since $\supp(\phi_n) \subset \supp(p)$, we can induce a sequence of $\{v_n\}$ from $\{\phi_n\}$ as $v_n = \phi_n/\sqrt{p(x)}$, which satisfies restrictions in \eqref{eqn:appxobj} and the objective $\clL_{v_n}$ will go to infinity when $n \to \infty$.
Note that any element in $\clL^2_p$, as a function, cannot take infinite value.
So the infinite supremum of the objective in \eqref{eqn:appxobj} cannot be obtained by any element in $\clL^2_p$, thus no optimal solution for the optimization problem.

\subsubsection*{A.4: Deduction of Proposition~\ref{prop:invexp}}

We first derive the exact inverse exponential map on the Wasserstein space $\clP_2(\clX)$, then develop finite-particle estimation for it.
Given $q, r \in \clP_2(\clX)$, $\Exp^{-1}_q (r)$ is defined as the tangent vector at $q$ of the geodesic curve $q_{t\in[0,1]}$ from $q$ to $r$.
When $q$ is absolutely continuous, the optimal transport map $\clT_q^r$ from $q$ to $r$ exists (\citet{villani2008optimal}, Thm.~10.38).
This condition fits the case of ParVIs since as our theory indicates, ParVIs have to do a smoothing treatment in any way, which is equivalent to assume an absolutely continuous distribution $q$.
Under this case, the geodesic is given by $q_t = \big( (1-t) \id + t \clT_q^r \big)_{\#} q$ (\citet{ambrosio2008gradient}, Thm.~7.2.2), and its tangent vector at $q$ (\ie, $t=0$) can be characterized by $\Exp^{-1}_q (r) = \lim_{t\to 0} \frac{1}{t}(\clT_q^{q_t} - \id)$ (\citet{ambrosio2008gradient}, Prop.~8.4.6).
Due to the uniqueness of optimal transport map, we have $\clT_q^{q_t} = (1-t) \id + t \clT_q^r$, so we finally get $\Exp^{-1}_q (r) = \clT_q^r - \id$.

To estimate it with a finite set of particles, we approximate the optimal transport map with the discrete one from particles $\{x^{(i)}\}_{i=1}^N$ of $q$ to particles $\{y^{(i)}\}_{i=1}^N$ of $r$.
As mentioned in the main context, it is a costly task, and the Sinkhorn approximations (for both the original version~\cite{cuturi2013sinkhorn} and an improved version~\cite{xie2018fast}) suffers from an unstable behavior in our experiments.
We now utilize the pairwise-close condition and develop a light and stable approximation.
The pairwise-close condition $d(x^{(i)}, y^{(i)}) \ll \min \! \big\{ \! \min_{j \neq i} d(x^{(i)}, x^{(j)}), \min_{j \neq i} d(y^{(i)}, y^{(j)}) \! \big\}$ indicates that $\frac{d(x^{(i)}, x^{(j)})}{d(x^{(j)}, y^{(j)})} \gg 1$, for any $i \neq j$.
On the other hand, due to triangle inequality, we have $d(x^{(i)}, y^{(j)}) \ge \lrvert{ d(x^{(i)}, x^{(j)}) - d(x^{(j)}, y^{(j)}) }$, or equivalently $\frac{d(x^{(i)}, y^{(j)})}{d(x^{(j)}, y^{(j)})} \ge \lrvert{ \frac{d(x^{(i)}, x^{(j)})}{d(x^{(j)}, y^{(j)})} - 1 }$.
Due to the above knowledge $\frac{d(x^{(i)}, x^{(j)})}{d(x^{(j)}, y^{(j)})} \gg 1$ by the pairwise-close condition, we have $\frac{d(x^{(i)}, y^{(j)})}{d(x^{(j)}, y^{(j)})} \gg 1$, or equivalently (by switching $i$ and $j$) $d(x^{(i)}, y^{(i)}) \ll \min_{j \neq i} d(x^{(i)}, y^{(j)})$.
This means that when transporting $\{x^{(i)}\}_i$ to $\{y^{(i)}\}_i$, the map $x^{(i)} \mapsto y^{(i)}$ for any $i$, has presumably the least cost.
More formally, consider any amount of transportation from $x^{(i)}$ to $y^{(j)}$ other than $y^{(i)}$.
It will introduce a change in the transportation cost that is proportional to $d(x^{(i)}, y^{(j)}) - d(x^{(i)}, y^{(i)}) + d(x^{(j)}, y^{(i)}) - d(x^{(j)}, y^{(j)})$, which is always positive due to our above recognition.
Thus we can reasonably approximate the optimal transport map $\clT_q^r$ by the discrete one $\clT_q^r(x^{(i)}) \approx y^{(i)}$.
With this approximation, we have $\big( \Exp^{-1}_q (r) \big) (x^{(i)}) = \clT_q^r(x^{(i)}) - x^{(i)} \approx y^{(i)} - x^{(i)}$.

\subsubsection*{A.5: Deduction of Proposition~\ref{prop:para}}

\begin{figure}
  \centering
  \includegraphics[width=.35\textwidth]{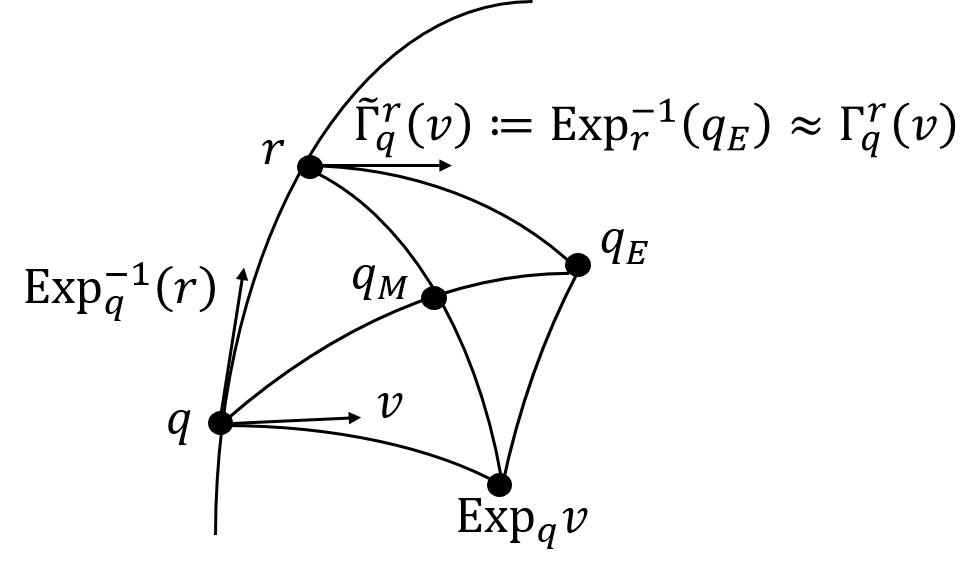}
  \vspace{-3pt}
  \caption{Illustration of the Schild's ladder method. Figure inspired by~\cite{kheyfets2000schild}.}
  \vspace{-5pt}
  \label{fig:schild}
\end{figure}

We derive the finite-particle estimation of the parallel transport on the Wasserstein space $\clP_2(\clX)$.
We follow the Schild's ladder method \cite{ehlers1972geometry, kheyfets2000schild} to parallel transport a tangent vector at $q$, $v \in T_q\clP_2(\clX)$, to the tangent space at $r$, $T_r\clP_2(\clX)$.
As shown in Fig.~\ref{fig:schild}, given $q$, $r$ and $v\in T_{q}\clP_2(\clX)$, the procedure to approximate $\Gamma_{q}^{r}(v)$ is
\begin{enumerate}
  \item find the point $\Exp_{q}(v)$;
  \item find the midpoint of the geodesic from $\Exp_{q}(v)$ to $r$: $q_M := \Exp_{\Exp_{q}(v)} ( \frac{1}{2} \Exp^{-1}_{\Exp_{q}(v)}(r) )$;
  \item extrapolate the geodesic from $q$ to $q_M$ by doubling the length to find $q_E := \Exp_{q}( 2 \Exp^{-1}_{q}(q_M) )$;
  \item take the approximator $\tGamma_{q}^{r}(v) := \Exp^{-1}_{r}(q_E) \approx \Gamma_{q}^{r}(v)$.
\end{enumerate}
Overall, the approximator $\tGamma_{q}^{r}(v)$ can be expressed as:
\begin{align}
  \Exp^{-1}_{r} \!\! \bigg( \!\! \Exp_{q} \! \bigg( \! 2 \Exp^{-1}_{q} \!\! \Big( \! \Exp_{\Exp_{q} \! (v)} \! \big( \frac{1}{2} \Exp^{-1}_{\Exp_{q} \! (v)}(r) \! \big) \! \Big) \!\! \bigg) \! \bigg).
  \label{eqn:schild-overall}
\end{align}
Note that the Schild's ladder method only requires the exponential map and its inverse.
It provides a tractable first-order approximation $\tGamma_{q}^{r}$ of the parallel transport $\Gamma_{q}^{r}$ under Levi-Civita connection, as needed.

Assume $q$ and $r$ are close in the sense of the Wasserstein distance, so that the Schild's ladder finds a good first-order approximation.
In the following we consider transporting $\varepsilon v$ for small $\varepsilon>0$ for the sake of the pairwise-close condition, and the result can be recovered by noting the linearity of the parallel transport: $\Gamma_{q}^{r}(\varepsilon v) = \varepsilon\Gamma_{q}^{r}(v)$.
Let $\{x^{(i)}\}_{i=1}^N$ and $\{y^{(i)}\}_{i=1}^N$ be the sets of samples of $q$ and $r$, respectively, and assume that they are pairwise close.

Now we follow the procedure.
\begin{enumerate}
  \item The measure $\Exp_{q} (\varepsilon v)$ can be identified as $(\id + \varepsilon v)_{\#}q$ due to the knowledge on the exponential map on $\clP_2(\clX)$ explained in Section~\ref{sec:wnag}, thus $\{x^{(i)} + \varepsilon v(x^{(i)})\}_{i=1}^N$ is a set of samples of $\Exp_{q} (\varepsilon v)$, and still pairwise close to $\{y^{(i)}\}_i$.
  \item The optimal map $\clT$ from $\Exp_{q} (\varepsilon v)$ to $r$ can be approximated by $\clT( x^{(i)} + \varepsilon v(x^{(i)}) ) \approx y^{(i)}$ since the two sets of samples are pairwise close.
	According to Theorem~7.2.2 of \cite{ambrosio2008gradient}, the geodesic from $\Exp_{q} (\varepsilon v)$ to $r$ is $t \mapsto \big( (1 - t) \id + t \clT \big)_{\#} (\Exp_{q} (\varepsilon v) )$, so due to the construction, we have $q_M = \big( \frac{1}{2} \id + \frac{1}{2} \clT \big)_{\#} (\Exp_{q} (\varepsilon v) )$.
	With the set of samples $\{x^{(i)} + \varepsilon v(x^{(i)})\}_{i=1}^N$ of $\Exp_{q} (\varepsilon v)$, we have a set of samples of $q_M$ by applying the transformation: $\big\{ \big( \frac{1}{2} \id + \frac{1}{2} \clT \big) ( x^{(i)} + \varepsilon v(x^{(i)}) ) \big\}_i \approx \left\{ \frac{1}{2} \big( x^{(i)} + \varepsilon v(x^{(i)}) + y^{(i)} \big) \right\}_i$.
  \item Similarly, a set of samples of $q_E$ is found as $\left\{ (1-t) x^{(i)} + \frac{1}{2} t \big( x^{(i)} + \varepsilon v(x^{(i)}) + y^{(i)} \big) \right\}_i \big|_{t=2} = \left\{ y^{(i)} + \varepsilon v(x^{(i)}) \right\}_i$ and is pairwise close to $\{y^{(i)}\}_i$.
  \item The approximated transported tangent vector $\Exp^{-1}_{r}(q_E)$ satisfies $\big( \Exp^{-1}_{r}(q_E) \big) (y^{(i)}) = \varepsilon v(x^{(i)})$.
\end{enumerate}
Finally, we get the approximation $\big( \Gamma_{q}^{r}(v) \big) (y^{(i)}) \approx \big( \tGamma_{q}^{r}(v) \big) (y^{(i)}) = v(x^{(i)})$.

\subsection*{B. Derivations of GFSF Vector Field $\hugfsf$}

\subsubsection*{B.1: Derivation with Vector-Valued Functions}
The vector field $\hugfsf$ is identified by the optimization problem~\origeqref{eqn:gfsf}:
\begin{align}
  \min_{u\in\clL^2} \max_{\substack{\phi\in\clH^D, \\ \lrVert{\phi}_{\clH^D}=1}} \Big(\sum_{i=1}^N \big(\phi(x^{(i)})\cdot u^{(i)} - \nabla\cdot\phi(x^{(i)})\big) \Big)^2,
\end{align}
where $u^{(i)} := u(x^{(i)})$.
For $\phi$ in $\clH^D$, by using the reproducing property $\lrangle{ \phi_{\alpha}(\cdot), K(x, \cdot) }_{\clH} = \phi_{\alpha}(x)$ and $\lrangle{ \phi_{\alpha}(\cdot), \partial_{x_{\beta}} K(x, \cdot) }_{\clH} = \partial_{x_{\beta}} \phi_{\alpha}(x)$~\cite{zhou2008derivative}, we can write the objective function as:
\begin{align}
  &\Big( \sum_{\alpha} \sum_j \big( u^{(j)}_{\alpha} \phi_{\alpha}(x^{(j)}) - \partial_{x^{(j)}_{\alpha}} \phi_{\alpha}(x^{(j)}) \big) \Big)^2 \\
  =& \left( \sum_{\alpha} \lrangle{ \sum_j \big( u^{(j)}_{\alpha} K(x^{(j)}, \cdot) - \partial_{x^{(j)}_{\alpha}} K(x^{(j)}, \cdot) \big), \phi_{\alpha}(\cdot) }_{\!\! \clH} \right)^{\!\! 2} \\
  =& \lrangle{ \sum_j \big( u^{(j)} K(x^{(j)}, \cdot) - \nabla_{x^{(j)}} K(x^{(j)}, \cdot) \big), \phi(\cdot) }_{\clH^D}^2.
\end{align}
We denote $\zeta := \sum_j \big( u^{(j)} K(x^{(j)}, \cdot) - \nabla_{x^{(j)}} K(x^{(j)}, \cdot) \big) \in \clH^D$.
Then the optimal value of the objective after maximizing out $\phi$ is $\lrVert{\zeta}_{\clH^D}^2 = \sum_{i,j}\big( u^{(i)}u^{(j)}K(x^{(i)}, x^{(j)}) - 2 u^{(i)} \nabla_{x^{(j)}} K(x^{(j)}, x^{(i)}) + \nabla_{x^{(i)}}\nabla_{x^{(j)}} K(x^{(i)}, x^{(j)}) \big) = \tr(\huu \hK \huu\trs) - 2 \tr(\hKp \huu\trs) + \const$, where $\huu_{:,i} := u^{(i)}$, and $\hK$, $\hKp$ are defined in the main text.
To minimize this quadratic function with respect to $\huu$, we further differentiate it with respect to $\huu$ and solve for the stationary point.
This finally gives the result $\hugfsf = \hKp \hK^{-1}$.

\subsubsection*{B.2: Derivation with Scalar-Valued Functions}
We denote $\varphi$ as scalar-valued functions on $\clX$.
For the equality $u(x) = -\nabla \log q(x)$, or $u(x)q(x) + \nabla q(x) = 0$, to hold in the weak sense with scalar-valued test function, we mean:
\begin{align}
  \bbE_{q(x)} [\varphi(x) u(x) - \nabla\varphi(x)] = 0, \forall \varphi\in\CC.
\end{align}
Let $\{x^{(j)}\}_j$ be a set of samples of $q(x)$.
Then the above requirement on $u(x)$ is:
\begin{align}
  \sum_j \left( \varphi(x^{(j)}) u^{(j)} - \nabla \varphi(x^{(j)}) \right) = 0, \forall \varphi\in\CC,
  \label{eqn:distrderiv_disc}
\end{align}
where $u^{(j)} = u(x^{(j)})$.
As analyzed above, for a valid vector field, we have to smooth the function $\varphi$.

For the above considerations, we restrict $\varphi$ in \eqref{eqn:distrderiv_disc} to be in the Reproducing Kernel Hilbert Space (RKHS) $\clH$ of some kernel $K(\cdot,\cdot)$, and convert the equation as the following optimization problem:
\begin{gather}
  \min_{\huu\in\bbR^{D\times N}} \max_{\substack{\varphi\in\clH, \\ \lrVert{\varphi}_{\clH} = 1}} J(\huu, \varphi), \\
  J(\huu, \varphi) := \sum_{j,\alpha} \left( \varphi(x^{(j)}) \huu_{\alpha j} - \partial_{x^{(j)}_{\alpha}}\varphi(x^{(j)}) \right)^2,
\end{gather}
where $\huu_{\alpha j} := u_{\alpha}(x^{(j)})$.
By using the reproducing properties of RKHS, we can write $J(\huu, \varphi)$ as:
\begin{align}
  J(\huu, \varphi) =& \sum_{\alpha} \lrangle{ \varphi(\cdot), \zeta_{\alpha}(\cdot) }^2_{\clH}, \\
  \zeta_{\alpha}(\cdot) :=& \sum_j \left( \huu_{\alpha j} K(x^{(j)}, \cdot) - \partial_{x^{(j)}_{\alpha}} K(x^{(j)}, \cdot) \right).
\end{align}
By linear algebra operations, we have:
\begin{align}
  \max_{\varphi\in\clH, \lrVert{\varphi}_{\clH}=1} J(\huu, \varphi) = \lambda_1(A(\huu)),
\end{align}
where $\lambda_1(A(\huu))$ is the largest eigenvalue of matrix $A$, and $A(\huu)_{\alpha \beta} = \lrangle{ \zeta_{\alpha}(\cdot), \zeta_{\beta}(\cdot) }_{\clH}$, or:
\begin{align}
  A(\huu) = \huu \hK \huu{}\trs - (\hKp \huu{}\trs + \huu \hKp{}\trs) + \hKpp,
\end{align}
with $\hKpp_{\alpha \beta} := \sum_{i,j} \partial_{x^{(i)}_{\alpha}}\partial_{x^{(j)}_{\beta}} K(x^{(i)}, x^{(j)})$.
For distinct samples $\hK$ is positive-definite, so we can conduct Cholesky decomposition: $\hK = G G\trs$ with $G$ non-singular.
Note that $A(\huu) = (\huu G - \hKp G^{-1}{}\trs) (\huu G - \hKp G^{-1}{}\trs)\trs + (\hKpp - \hKp \hK^{-1} \hKp{}\trs)$.
So whenever $\huu G \neq \hKp G^{-1}{}\trs$, the first term will be positive semidefinite with positive largest eigenvalue, which makes $\lambda_1(A(u)) > \lambda_1(\hKpp- \hKp \hK^{-1} \hKp{}\trs)$, a constant with respect to $\huu$.
So to minimize $\lambda_1(A(\huu))$, we require $\huu G = \hKp G^{-1}{}\trs$, \ie, ${\huu} = \hKp (G G\trs)^{-1} = \hKp \hK^{-1}$.
This result coincides with the one for vector-valued functions $\phi\in\clH^D$.

In practice, for numerical stability, we add a small diagonal matrix to $\hK$ before conducting inversion.
This is a common practice.
Particularly, it is adopted by \citet{li2018gradient} for the same estimate.

\subsection*{C. Details on Accelerated First-Order Methods on the Wasserstein Space $\clP_2(\clX)$}

\subsubsection*{C.1: Simplification of Riemannian Accelerated Gradient (RAG) with Approximations}

We consider the general version of RAG (Alg.~2 of \citet{liu2017accelerated}).
It updates the target variable $q_k$ as:
\begin{align}
  q_{k} = \Exp_{r_{k-1}} (\varepsilon v_{k-1}),
\end{align}
where $v_{k-1} := -\grad \KL(r_{k-1})$.
The update rule for the auxiliary variable $r_k$ is given by the solution of the following non-linear equation (see Alg.~2 and Eq.~(5) of \citet{liu2017accelerated}):
\begin{align}
  & \Gamma_{r_k}^{r_{k-1}} \!\left(\! \frac{k}{\alpha-1} \Exp^{-1}_{r_k}\!(q_k) + \frac{D v_k}{\lrVert{v_k}_{r_k}} \!\right) \\
  \!=\! & \frac{k-1}{\alpha-1} \Exp^{-1}_{r_{k-1}}\!(q_{k-1}) - \frac{k+\alpha-2}{\alpha-1} \varepsilon v_{k-1} + \frac{D v_{k-1}}{\lrVert{v_{k-1}}_{r_{k-1}}}.
\end{align}
Here we focus on simplifying this complicated update rule for $r_k$ with moderate approximations.
We note that the original work of RAG \cite{liu2017accelerated} actually adopted these approximations in experiments, but the simplification of the general algorithm is not given in the work.

Applying $\left( \Gamma_{r_k}^{r_{k-1}} \right)^{-1}$ to both sides of the equation and noticing that $\left( \Gamma_{r_k}^{r_{k-1}} \right)^{-1} = \Gamma_{r_{k-1}}^{r_k}$, the above equation can be reformulated as:
\begin{align}
  & \frac{k}{\alpha-1} \Exp^{-1}_{r_k}\!(q_k) + \frac{D v_k}{\lrVert{v_k}_{r_k}} \\
  = & \Gamma_{r_{k-1}}^{r_k} \left( \frac{k-1}{\alpha-1} \Exp^{-1}_{r_{k-1}}\!(q_{k-1}) - \frac{k+\alpha-2}{\alpha-1} \varepsilon v_{k-1} \right) \\
  & + \frac{D \Gamma_{r_{k-1}}^{r_k} (v_{k-1})}{\lrVert{v_{k-1}}_{r_{k-1}}}.
\end{align}
Approximating $v_k$ on the left hand side of the equation by $\Gamma_{r_{k-1}}^{r_{k}}(v_{k-1})$ and noting that $\lrVert{\Gamma_{r_{k-1}}^{r_{k}}(v_{k-1})}_{r_{k}} = \lrVert{v_{k-1}}_{r_{k-1}}$, we have:
\begin{align}
  & \frac{k}{\alpha-1} \Exp^{-1}_{r_k}\!(q_k) \\
  = & \Gamma_{r_{k-1}}^{r_k} \left( \frac{k-1}{\alpha-1} \Exp^{-1}_{r_{k-1}}\!(q_{k-1}) - \frac{k+\alpha-2}{\alpha-1} \varepsilon v_{k-1} \right).
\end{align}
Using the fact that $\Exp^{-1}_{r_k}(q_k) = -\Gamma_{q_k}^{r_k}(\Exp^{-1}_{q_k}(r_k))$ and applying $\left( \Gamma_{q_k}^{r_k} \right)^{-1} = \Gamma_{r_k}^{q_k}$ to both sides of the equation, we have:
\begin{align}
  & \Exp^{-1}_{q_k}(r_k) \\
  = & - \Gamma_{r_k}^{q_k} \Gamma_{r_{k-1}}^{r_k} \left( \frac{k-1}{k} \Exp^{-1}_{r_{k-1}}\!(q_{k-1}) - \frac{k+\alpha-2}{k} \varepsilon v_{k-1} \right).
\end{align}
Approximating $\Gamma_{r_k}^{q_k}\Gamma_{r_{k-1}}^{r_k}$ by $\Gamma_{r_{k-1}}^{q_k}$, we finally have $r_k =$
\begin{align}
  \Exp_{q_k} \!\! \left[\!  - \Gamma_{r_{k-1}}^{q_k}\!\!  \left( \frac{k-1}{k} \Exp^{-1}_{r_{k-1}}\!(q_{k-1}) - \frac{k\!+\!\alpha\!-\!2}{k} \varepsilon v_{k-1} \right) \right].
\end{align}

\subsubsection*{C.2: Reformulation of Riemannian Nesterov's Method (RNes)}

We consider the constant step version of RNes (Alg.~2 of~\cite{zhang2018estimate}).
It introduces an additional auxiliary variable $s_k \in \clP_2(\clX)$, and update the variables in iteration $k$ as:
\begin{subequations}
\begin{align}
  & r_{k-1} = \Exp_{q_{k-1}} \left( c_1 \Exp^{-1}_{q_{k-1}} (s_{k-1}) \right), \label{eqn:rnes-orig-1} \\
  & q_{k} = \Exp_{r_{k-1}} \left( \varepsilon v_{k-1} \right), \label{eqn:rnes-orig-2} \\
  & s_{k} = \Exp_{r_{k-1}} \left( \frac{1 - \alpha}{1 + \beta} \Exp_{r_{k-1}}^{-1} (s_{k-1}) + \frac{\alpha}{(1 + \beta) \gamma} v_{k-1} \right), \label{eqn:rnes-orig-3}
\end{align}
\end{subequations}
where $v_{k-1} := -\grad \KL(r_{k-1})$, and the coefficients $\alpha, \gamma, c_1$ are set by a step size $\varepsilon > 0$, a shrinkage parameter $\beta > 0$, and a parameter $\mu > 0$ upper bounding the Lipschitz coefficient of the gradient of the objective, in the following way:
\begin{align}
  \label{eqn:rnes-params}
  \begin{split}
	& \alpha = \frac{\sqrt{\beta^2 + 4(1 + \beta) \mu \varepsilon} - \beta}{2}, \\
	& \gamma = \frac{\sqrt{\beta^2 + 4(1 + \beta) \mu \varepsilon} - \beta}{\sqrt{\beta^2 + 4(1 + \beta) \mu \varepsilon} + \beta} \mu, \\
	& c_1 = \frac{\alpha \gamma}{\gamma + \alpha \mu}.
  \end{split}
\end{align}

Now we simplify the update rule by collapsing the variable $s$.
Referring to \eqref{eqn:rnes-orig-1}, the variable $s_{k-1}$ can be expressed by:
\begin{align}
  s_{k-1} = \Exp_{q_{k-1}} \left( \frac{1}{c_1} \Exp_{q_{k-1}}^{-1} (r_{k-1}) \right).
\end{align}
This result indicates that $s_{k-1}$ lies on the $1 / c_1$ portion of the geodesic from $q_{k-1}$ to $r_{k-1}$, which is the $(1 - 1 / c_1)$ portion of the geodesic from $r_{k-1}$ to $q_{k-1}$.
According to this knowledge, we have:
\begin{align}
  \Exp_{r_{k-1}}^{-1} (s_{k-1}) = \left( 1 - \frac{1}{c_1} \right) \Exp_{r_{k-1}}^{-1} (q_{k-1}).
\end{align}
Substitute this result into \eqref{eqn:rnes-orig-3}, we have:
\begin{align}
  s_{k} = \Exp_{r_{k-1}} \bigg( & \frac{1 - \alpha}{1 + \beta} \left( 1 - \frac{1}{c_1} \right) \Exp_{r_{k-1}}^{-1} (q_{k-1}) \\
  & + \frac{\alpha}{(1 + \beta) \gamma \varepsilon} \Exp_{r_{k-1}}^{-1} (q_k) \bigg),
\end{align}
where we have also substituted $v_{k-1}$ with $\frac{1}{\varepsilon} \Exp_{r_{k-1}}^{-1} (q_k)$ according to \eqref{eqn:rnes-orig-2}.
Leveraging \eqref{eqn:rnes-params} to simplify the coefficients in the above equation, we get:
\begin{align}
  s_{k} = \Exp_{r_{k-1}} \! \left( \! \left( \! 1 \! - \! c_2 \! \right) \Exp_{r_{k-1}}^{-1} \! (q_{k-1}) + c_2 \Exp_{r_{k-1}}^{-1} \! (q_k) \! \right),
\end{align}
where the coefficient $c_2 := 1 / \alpha$.
Replacing $k \to k + 1$ in \eqref{eqn:rnes-orig-1} and substitute with the above result, we have the update rule for $r_k$:
\begin{align}
  r_{k} = \Exp_{q_{k}} \! \Big\{ \! c_1 \Exp^{-1}_{q_{k}} \! \Big[ \Exp_{r_{k-1}} \! \Big( \! (1 - c_2) \Exp_{r_{k-1}}^{-1} (q_{k-1}) \\
  + c_2 \Exp_{r_{k-1}}^{-1} (q_k) \Big) \Big] \Big\},
\end{align}
which builds the update rule of RNes together with \eqref{eqn:rnes-orig-1}.

In our implementation, the parameters are tackled with $\varepsilon, \beta, \mu$ instead of setting $c_1, c_2$ directly.
The shrinkage parameter $\beta$ is set in the scale of $\sqrt{\mu \varepsilon}$.
In our Alg.~\ref{alg:wacc}, the coefficient $c_1 (c_2 - 1)$ can be expressed as:
\begin{align}
  1 + \beta - \frac{2(1+\beta)(2+\beta) \mu \varepsilon}{\sqrt{\beta^2 + 4(1+\beta)\mu\varepsilon} - \beta + 2(1+\beta)\mu\varepsilon}.
\end{align}

\subsubsection*{C.3: Deduction of Wasserstein Accelerated Gradient (WAG) and Wasserstein Nesterov's Method (WNes) (Alg.~\ref{alg:wacc})}

First consider developing WAG based on RAG.
We denote the vector field $\zeta_{k-1} := \frac{k-1}{k} \Exp^{-1}_{r_{k-1}}(q_{k-1}) - \frac{k+\alpha-2}{k} \varepsilon v_{k-1}$ for simplicity, so $r_{k} = \Exp_{q_{k}}\left[ -\Gamma_{r_{k-1}}^{q_{k}} (\zeta_{k-1}) \right]$, due to the update rule of RAG.
We assume that $\{x^{(i)}_{k-1}\}_{i=1}^N$ of $q_{k-1}$ and $\{y^{(i)}_{k-1}\}_{i=1}^N$ of $r_{k-1}$ are pairwise close, so from Section~\ref{sec:wnag} we know that $\Exp^{-1}_{r_{k-1}}(q_{k-1})(y^{(i)}_{k-1}) = x^{(i)}_{k-1} - y^{(i)}_{k-1}$, thus $\zeta_{k-1}(y^{(i)}_{k-1}) = \frac{k-1}{k}(x^{(i)}_{k-1} - y^{(i)}_{k-1}) - \frac{k+\alpha-2}{k} \varepsilon v_{k-1}^{(i)}$.
Due to the update rule for $q_k$ that we already discovered: $x^{(i)}_k = y^{(i)}_{k-1} + \varepsilon v^{(i)}_{k-1}$, we know that $\{x^{(i)}_k\}_{i=1}^N$ of $q_k$ and $\{y^{(i)}_{k-1}\}_{i=1}^N$ of $r_{k-1}$ are pairwise close, for small enough step size $\varepsilon$.
Using the parallel transport estimate developed above with Schild's ladder method, $\big( \Gamma_{r_{k-1}}^{q_k}(\zeta_{k-1}) \big) (x^{(i)}_{k}) \approx \zeta_{k-1}(y^{(i)}_{k-1})$.
So finally, we assign $y^{(i)}_k = x^{(i)}_k - \big( \Gamma_{r_{k-1}}^{q_k} (\zeta_{k-1}) \big) (x^{(i)}_k) \approx x^{(i)}_k - \zeta_{k-1}(y^{(i)}_{k-1}) = x^{(i)}_k - \frac{k-1}{k}(x^{(i)}_{k-1} - y^{(i)}_{k-1}) + \frac{k+\alpha-2}{k} \varepsilon v_{k-1}^{(i)}$ as a sample of $r_k$.

We note that initially $x^{(i)}_0 = y^{(i)}_0$.
Assume $\{x^{(i)}_{k-1}\}_{i=1}^N$ and $\{y^{(i)}_{k-1}\}_{i=1}^N$ are pairwise close, so for sufficiently small $\varepsilon$, $\zeta_{k-1}(y^{(i)}_{k-1})$ is an infinitesimal vector for all $i$.
This, in turn, indicates that $\{x^{(i)}_k\}_{i=1}^N$ of $q_k$ and $\{y^{(i)}_k\}_{i=1}^N$ of $r_k$ are pairwise close, which provides the assumption for the next iteration.
The derivation of WNes based on RNes can be developed similarly, and we omit verbosing the procedure.

\subsection*{D. Details on the HE Method for Bandwidth Selection}
We first note that the bandwidth selection problem cannot be solved using theories of heat kernels, which aims to find the evolving density under the Brownian motion with known initial distribution, while in our case the density is unknown and we want to find an update on samples to approximate the effect of Brownian motion.

According to the derivation in the main context, we write the dimensionless final objective explicitly:
\begin{align}
  & \frac{1}{h^{D+2}} \sum_k \lambda(x^{(k)})^2 \\
  =& \frac{1}{h^{D+2}} \sum_k \Big[ \Delta\tqq(x^{(k)}; \{x^{(i)}\}_i) \\
  &\hspace{10pt} {} + \sum_j \nabla_{\! x^{(j)}} \tqq(x^{(k)}; \{x^{(i)}\!\}_i) \cdot \nabla\!\log\tqq(x^{(j)}; \{x^{(i)}\!\}_i) \Big]^2.
\end{align}
For $\tqq(x; \{x^{(j)}\}_j) = (1/Z)\sum_j c(\lrVert{x-x^{(j)}}^2/(2h))$, the above objective becomes: 
\begin{align}
  \sum_k & \bigg( \sum_j \bigg[ c''_j(x) \lrVert{x-x^{(j)}}^2 + Dh c'_j(x) \\
  & {} + \frac{\big(\sum_i c'_{ij} x^{(i)}\big) - \big(\sum_i c'_{ij}\big) x^{(j)}}{\big(\sum_i c_{ij}\big)} \cdot (x - x^{(j)}) c'_j(x) \bigg] \bigg)^2,
\end{align}
where $c'_j(x) = c'(\lrVert{x-x^{(j)}}^2/(2h))$, $c'_{ij} = c'_j(x^{(i)})$, $c_{ij} = c(\lrVert{x^{(i)}-x^{(j)}}^2/(2h))$.
For Gaussian kernel $c(r) = (2\pi h)^{-\frac{D}{2}} e^{-r}$, denoting $g_k^2(h)$ as the summand for $k$ of the l.h.s. of the above equation, we have:
\begin{align}
  & (2\pi)^{\frac{D}{2}} g_k(h) \\
 =& \hphantom{ {} - {} } \big( \sum_j e_{kj} \lrVert{d_{kj}}^2 \big) - hD\big( \sum_j e_{kj} \big) \\
  & {} - \sum_j \big(\sum_i e_{ij}\big)^{-1} e_{jk} d_{jk} \cdot \big( \sum_i e_{ij} d_{ij} \big),\\
  & (2\pi)^{\frac{D}{2}} g'_k(h) \\
  =& \hphantom{ {} + {} } \frac{1}{2h^2} \big( \sum_j e_{jk} \lrVert{d_{jk}}^4 \big) - \frac{D}{h} \big( \sum_j e_{jk} \lrVert{d_{jk}}^2 \big) \\
  & {} + \big( \frac{D^2}{2} - D \big) \big( \sum_j e_{jk} \big) \\
  & {} - \frac{1}{2h^2} \sum_j \big(\sum_i e_{ij}\big)^{-1} e_{jk} d_{jk} \cdot \big( \sum_i e_{ij} \lrVert{d_{ij}}^2 d_{ij} \big) \\
  & {} - \frac{1}{2h^2} \sum_j \big(\sum_i e_{ij}\big)^{-1} e_{jk} \lrVert{d_{jk}}^2 d_{jk} \cdot \big( \sum_i e_{ij} d_{ij} \big) \\
  & {} + \frac{1}{2h^2} \sum_j \big(\sum_i e_{ij}\big)^{-2} \big(\sum_i e_{ij} \lrVert{d_{ij}}^2\big) e_{jk} d_{jk} \cdot \big( \sum_i e_{ij} d_{ij} \big) \\
  & {} + \frac{D}{2h} \sum_j \big(\sum_i e_{ij}\big)^{-1} e_{jk} d_{jk} \cdot \big(\sum_i e_{ij} d_{ij} \big),
\end{align} 
where $d_{ij} = x^{(i)} - x^{(j)}$, $e_{ij} = e^{-\lrVert{d_{ij}}^2/(2h) -(D/2)\log h}$.

Although the evaluation of $g_k(h)$ may induce some computation cost, the optimization is with respect to a scalar.
In each particle update iteration, before estimating the vector field $v$, we first update the previous bandwidth by one-step exploration with quadratic interpolation, which only requires one derivative evaluation and two value evaluations.


\subsection*{E. Detailed Settings and Parameters of Experiments}

\subsubsection*{E.1: Detailed Settings and Parameters of the Synthetic Experiment}
The bimodal toy target distribution is inspired by the one of \cite{rezende2015variational}.
The logarithm of the target density $p(z)$ for $z = (z_1, z_2) \in \bbR^2$ is given by:
\begin{align}
  \log p(z) =& -2 ( \lrVert{z}_2^2 - 3)^2 + \log ( e^{-2 (z_1 - 3)^2} + e^{-2 (z_1 + 3)^2} ) \\
  & {} + \const.
\end{align}
The region shown in each figure is $[-4,4] \times [-4,4]$.
The number of particles is 200, and all particles are initialized with standard Gaussian $\clN(0, 1)$.

All methods are run for 400 iterations, and all follows the plain WGD method.
SVGD uses fixed step size $0.3$, while other methods (Blob, GFSD, GFSF) share the fixed step size $0.01$.
This is because that the updating direction of SVGD is a kernel smoothed one, so it may have a different scale from other methods.
Note that the AdaGrad with momentum method in the original SVGD paper \cite{liu2016stein} is not used.
For GFSF, a small diagonal matrix $0.01 I$ is added to $\hK$ before conducting inversion, as discussed at the end of Appendix~B.2.

\subsubsection*{E.2: Detailed Settings and Parameters of the BLR Experiment}

We adopt the same settings as \cite{liu2016stein}, which is also adopted by \cite{chen2018unified}.
The Covertype dataset contains 581,012 items with each 54 features.
Each run uses a random 80\%-20\% split of the dataset.

For the model, parameters of the Gamma prior on the precision of the Gaussian prior of the weight are $a_0 = 1.0$, $b_0 = 100$ ($b_0$ is the scale parameter, not the rate parameter).
All methods use 100 particles, randomly initialized by the prior.
Batch size for all methods is 50.

Detailed parameters of various methods are provided in Table~\ref{tab:blr-params}.
The WGD column provides the step size.
The format of the PO column is ``PO parameters(decaying exponent, remember rate, injected noise variance), step size''.
Both methods use a fixed step size, while the WAG and WNes methods use a decaying step size.
The format of WAG column is ``WAG parameter $\alpha$, (step size decaying exponent, step size)'' (see Alg.~\ref{alg:wacc}),
and the format of WNes column is ``Wnes parameters ($\mu$, $\beta$), (step size decaying exponent, step size)'' (see Appendix~C.2).
One exception is that SVGD-WGD uses the AdaGrad with momentum method to reproduce the results of \cite{liu2016stein}, which uses remember rate $0.9$ and step size $0.03$.
For GFSF, the small diagonal matrix is (1e-5)$I$.

\begin{table}[t]
  \centering
  \caption{Parameters of various methods in the BLR experiment}
  \label{tab:blr-params}
  \begin{tabular}{l|cc}
	\toprule
	& WGD & PO \\
	\midrule
	SVGD & 3e-2 & (1.0, 0.7, 1e-7), 3e-6 \\
	Blob & 1e-6 & (1.0, 0.7, 1e-7), 3e-7 \\
	GFSD & 1e-6 & (1.0, 0.7, 1e-7), 3e-7 \\
	GFSF & 1e-6 & (1.0, 0.7, 1e-7), 3e-7 \\
	\bottomrule
	& WAG & WNes \\
	\midrule
	SVGD & 3.9, (0.9, 1e-6) & ( 300, 0.2), (0.8, 3e-4) \\
	Blob & 3.9, (0.9, 1e-6) & (1000, 0.2), (0.9, 1e-5) \\
	GFSD & 3.9, (0.9, 1e-6) & (1000, 0.2), (0.9, 1e-5) \\
	GFSF & 3.9, (0.9, 1e-6) & (1000, 0.2), (0.9, 1e-5) \\
	\bottomrule
  \end{tabular}
\end{table}


\subsubsection*{E.3: Detailed Settings and Parameters of the BNN Experiment}

We follow the same settings as \citet{liu2016stein}.
For each run, a random 90\%-10\% train-test split is conducted.
The BNN model contains one hidden layer with 50 hidden nodes, and sigmoid activation is used.
The parameters of the Gamma prior on the precision parameter of the Gaussian prior of the weights are $a_0 = 1.0$, $b_0 = 0.1$.
Batch size is set to 100.
Number of particles is fixed as 20 for all methods.
Results are collected after 8,000 iterations for every method.

Detailed parameters of various methods are provided in Table~\ref{tab:bnn-params}.
The format of each column is the same as illustrated in Appendix~E.2, except that all SVGD methods uses the AdaGrad with momentum method with remember rate $0.9$, so we only provide the step size.
WGD and PO methods also adopt the decaying step size, so we provide the decaying exponent.
For GFSF, the small diagonal matrix is (1e-2)$I$.

\begin{table}[t]
  \centering
  \caption{Parameters of various methods in the BNN experiment}
  \label{tab:bnn-params}
  \begin{tabular}{l|cc}
	\toprule
	& WGD & PO \\
	\midrule
	SVGD & 1e-3        & (1.0, 0.6, 1e-7), 1e-4        \\
	Blob & (0.5, 3e-5) & (1.0, 0.8, 1e-7), (0.5, 3e-5) \\
	GFSD & (0.5, 3e-5) & (1.0, 0.8, 1e-7), (0.5, 3e-5) \\
	GFSF & (0.5, 3e-5) & (1.0, 0.8, 1e-7), (0.5, 3e-5) \\
	\bottomrule
	& WAG & WNes \\
	\midrule
	SVGD & 3.6, 1e-6        & (1000, 0.2), 1e-4 \\
	Blob & 3.5, (0.5, 1e-5) & (3000, 0.2), (0.6, 1e-4) \\
	GFSD & 3.5, (0.5, 1e-5) & (3000, 0.2), (0.6, 1e-4) \\
	GFSF & 3.5, (0.5, 1e-5) & (3000, 0.2), (0.6, 1e-4) \\
	\bottomrule
  \end{tabular}
\end{table}


\subsubsection*{E.4: Detailed Settings and Parameters of the LDA Experiment}

We follow the same settings as \citet{ding2014bayesian}.
The dataset is the ICML dataset\footnote{\url{https://cse.buffalo.edu/~changyou/code/SGNHT.zip}} that contains 765 documents with vocabulary size 1,918.
We also adopt the Expanded-Natural parameterization \cite{patterson2013stochastic}, and collapsed Gibbs sampling for stochastic gradient estimation.
In each document, 90\% of words are used for training the topic proportion of the document, and the left 10\% words are used for evaluation.
For each run, a random 80\%-20\% train-test split of the dataset is done.

For the LDA model, we fix the parameter of the Dirichlet prior on topics as 0.1, and mean and standard deviation of the Gaussian prior on the topic proportion as 0.1 and 1.0, respectively.
The number of topics is fixed as 30, and batch size is fixed as 100 for all methods.
Collapsed Gibbs sampling is run for 50 iterations for each stochastic gradient estimation.
Particle size is fixed as 20 for all methods.

\begin{table}[t]
  \centering
  \caption{Parameters of various methods in the LDA experiment}
  \label{tab:lda-params}
  \begin{tabular}{l|cc}
	\toprule
	& WGD & PO \\
	\midrule
	SVGD & 3.0 & (0.7, 0.7, 1e-4), 10.0 \\
	Blob & 0.3 & (0.7, 0.7, 1e-4), 0.30 \\
	GFSD & 0.3 & (0.7, 0.7, 1e-4), 0.30 \\
	GFSF & 0.3 & (0.7, 0.7, 1e-4), 0.30 \\
	\bottomrule
	& WAG & WNes \\
	\midrule
	SVGD & 2.5, 3.0  & (3.0, 0.2), 10.0 \\
	Blob & 2.1, 3e-2 & (0.3, 0.2), 0.30  \\
	GFSD & 2.1, 3e-2 & (0.3, 0.2), 0.30  \\
	GFSF & 2.1, 3e-2 & (0.3, 0.2), 0.30  \\
	\bottomrule
  \end{tabular}
\end{table}

\begin{figure}[t]\vspace{-5pt}
  \centering
  \hspace{-5pt}
  \subfigure[SVGD]{
	\includegraphics[scale=.23]{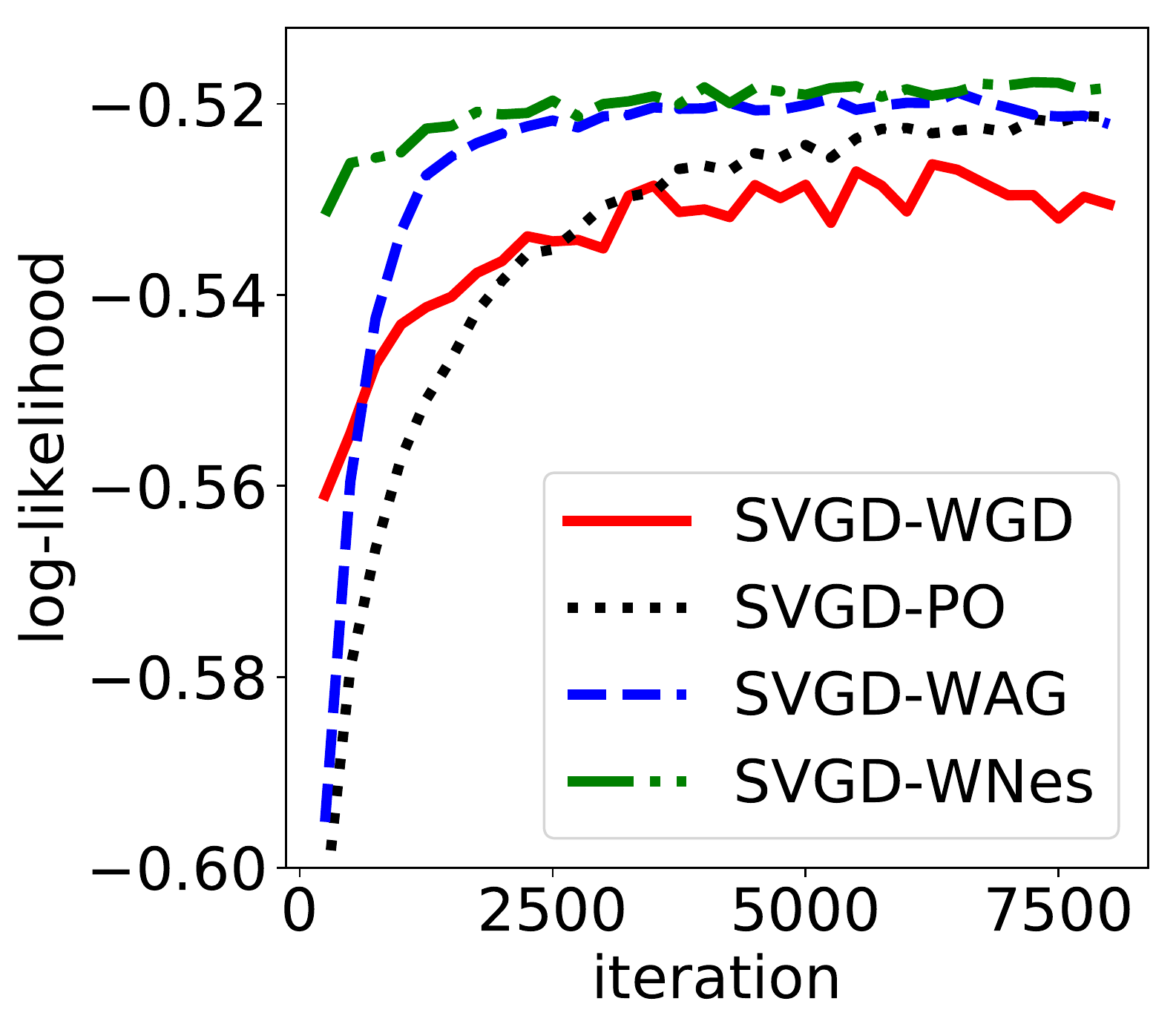}\vspace{-4pt}
  }
  \hspace{-5pt}
  \subfigure[Blob]{
	\includegraphics[scale=.23]{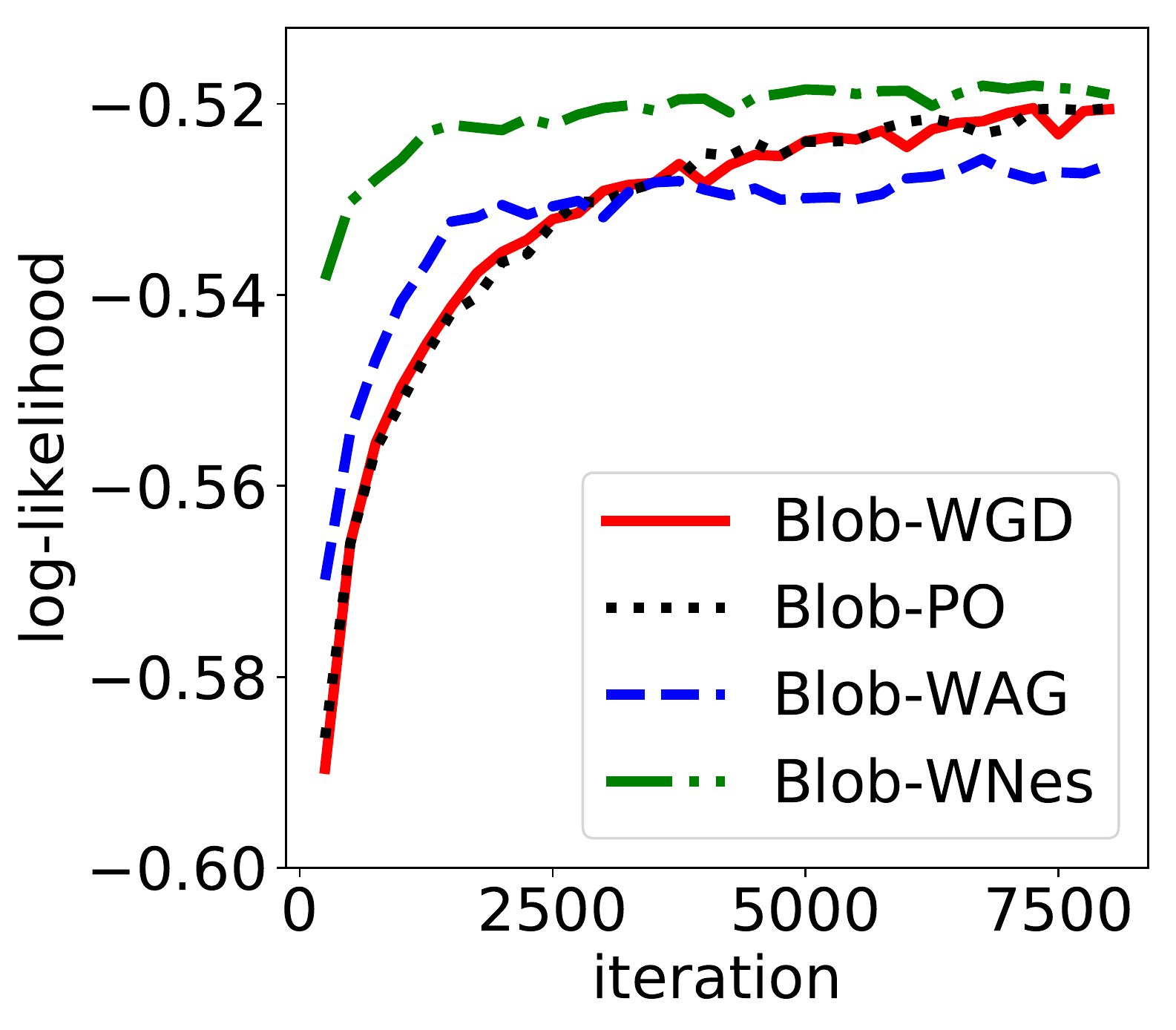}\vspace{-4pt}
  }
  \\
  \hspace{-5pt}
  \subfigure[GFSD]{
	\includegraphics[scale=.23]{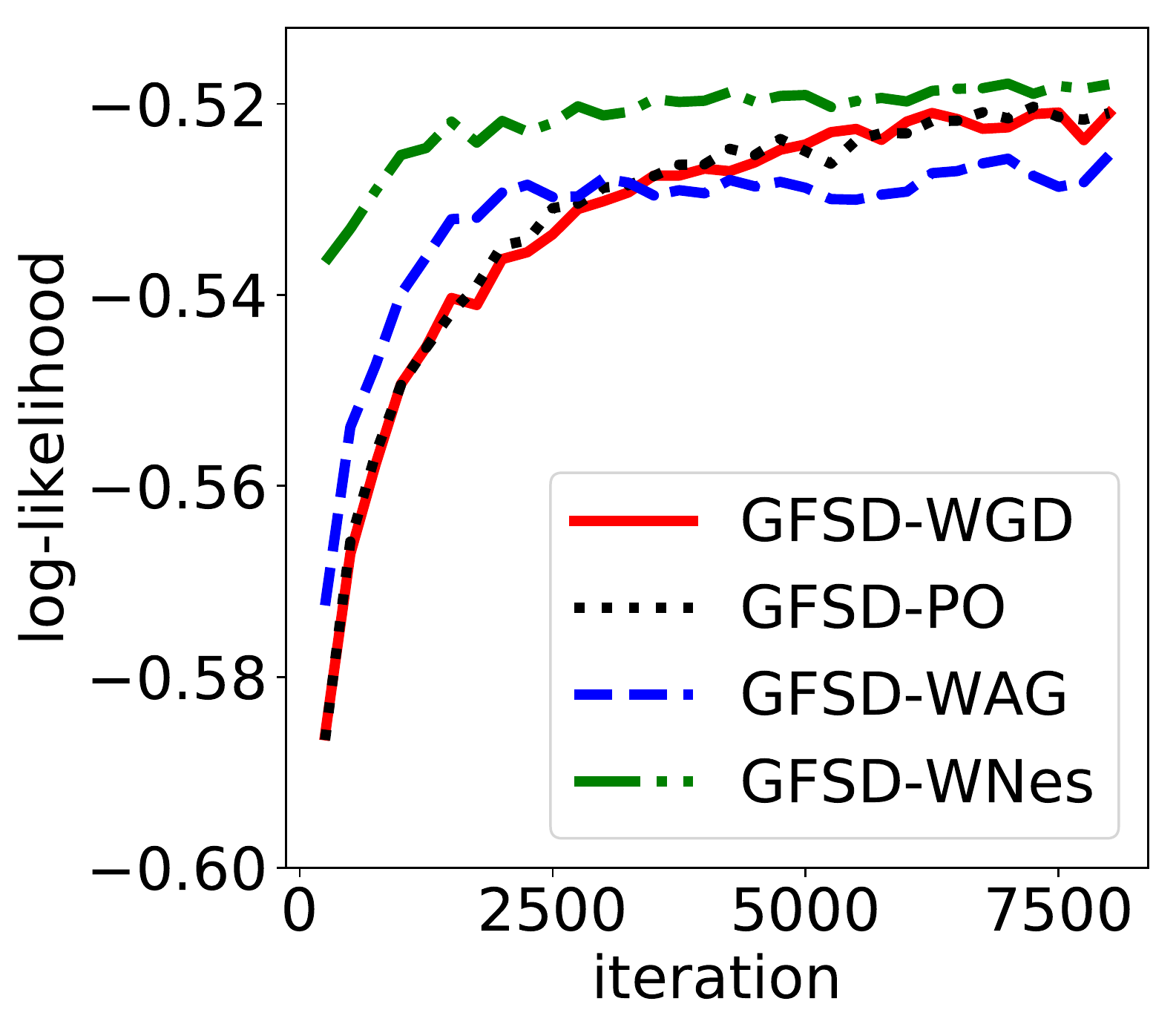}\vspace{-4pt}
  }
  \hspace{-5pt}
  \subfigure[GFSF]{
	\includegraphics[scale=.23]{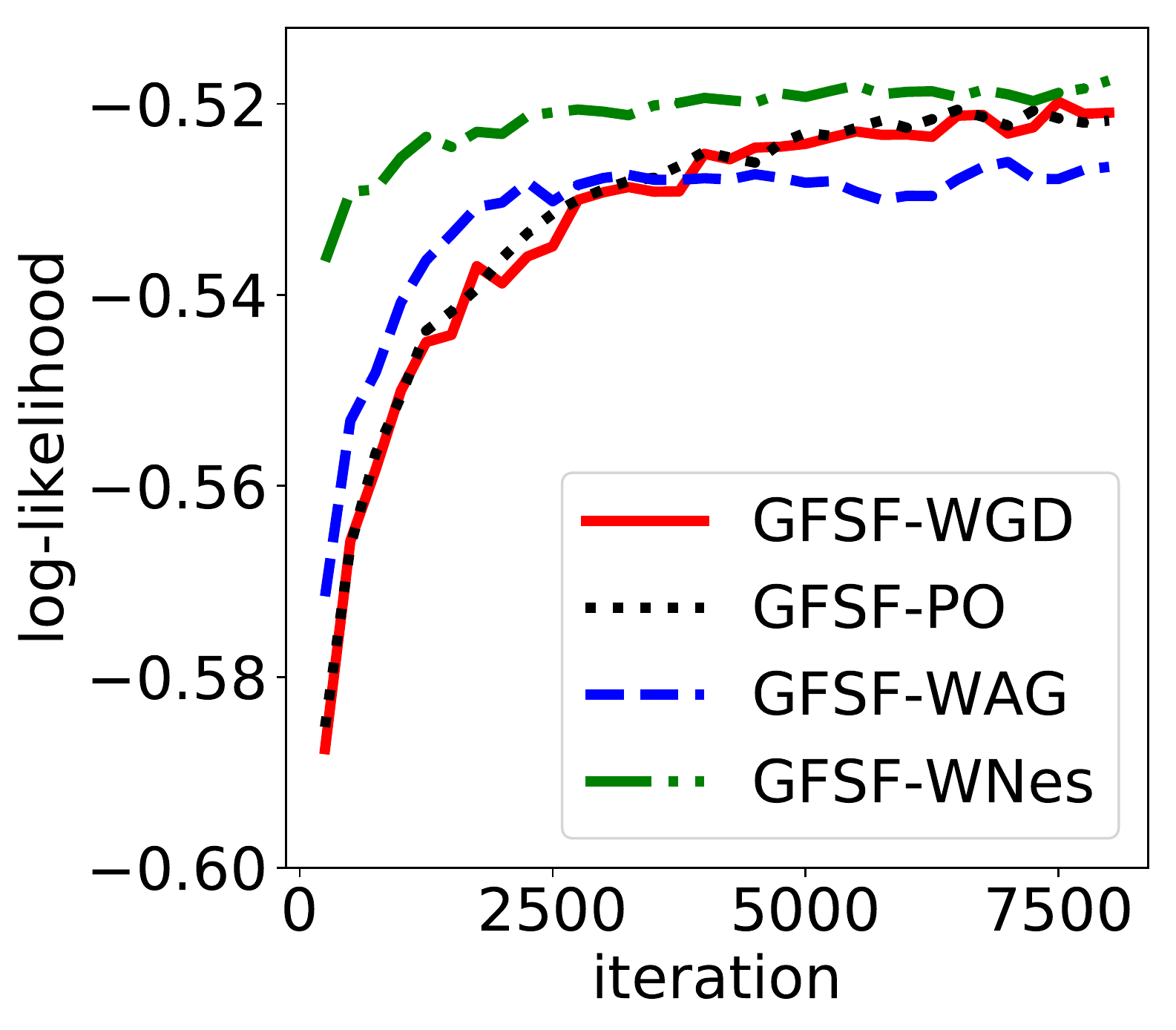}\vspace{-4pt}
  }
  \vspace{-3pt}
  \caption{Acceleration effect of WAG and WNes on BLR on the Covertype dataset, measured in log-likelihood.
	See Appendix~E.2 for detailed experiment settings and parameters.
  }
  \label{fig:blr-llh}
  \vspace{-10pt}
\end{figure}

Detailed parameters of all methods are provided in Table~\ref{tab:lda-params}.
The format of each column is the same as illustrated in Appendix~E.2, except that all methods uses a decaying step size with decaying exponent 0.55 and initial steps 1,000, so we only provide the step size for all methods.
SVGD methods do not use the AdaGrad with momentum method.
For GFSF, the small diagonal matrix is (1e-5)$I$.

For SGNHT, both its sequential and parallel simulations use the fixed step size of 0.03, and its mass and diffusion parameters are set to 1.0 and 22.4.

\subsection*{F. More Experimental Results}

\subsubsection*{F.1: More Results on the BLR Experiment}

The results measured in log-likelihood on test dataset corresponding to the results measured in test accuracy in Fig.~\ref{fig:blr-acc} is provided in Fig.~\ref{fig:blr-llh}.
Acceleration effect of our WAG and WNes methods is again clearly demonstrated, making a consistent support.
Particularly, the WNes method is more stable than the WAG method.

\end{document}